\documentclass[twoside]{article}
\usepackage{color}
\usepackage[preprint]{aistats2027}
\usepackage{booktabs}
\usepackage{graphicx}
\usepackage{subcaption}
\usepackage{multirow}
\usepackage{url}
\usepackage{cite}
\usepackage{amsthm}  
\usepackage{float}
\newtheorem{theorem}{\bf Theorem}
\newtheorem{assumption}{\bf Assumption}
\newtheorem{lemma}{\bf Lemma}
\usepackage{amssymb}
\newcommand{\Prob}{\mathbb{P}}

\begin{document}

%

%

\twocolumn[

\runningtitle{Sequential Testing on Complex Data Streams via Unsupervised Model Errors}
\aistatstitle{ Sequential Testing on Complex and High Dimensional Data Streams Leveraging Unsupervised Model Errors}

\aistatsauthor{George Stamtelis \And Iordanis Koutsopoulos}

\aistatsaddress{ Athens University of Economics and Business} ]

\begin{abstract}
We propose two general, distribution-free test statistics for binary Sequential Hypothesis Testing (SHT) and Sequential Change Point Detection (SCPD) respectively, in streaming environments where the data (e.g., network traffic metadata) are too complicated for the exact probabilistic modeling required by conventional likelihood ratio-driven tests. Our framework also addresses the challenge of information asymmetry, where data is available only from the null/pre-change regime and not from the alternative/post-change one  such as zero-day attacks or emerging disease related regimes. The proposed statistics  monitor the  errors of a predictive or reconstructive machine learning model trained solely on null/pre-change data, relying on imperfect generalization (the model's performance will degrade when deployed on data from the alternative/post-change class). Under mild  assumptions on the model's errors, we establish rigorous analytical guarantees: strict non-asymptotic error control for SHT, and explicit bounds on the Average Run Length (ARL) and the Expected Detection Delay (EDD) for SCPD.  Extensive evaluations in target tracking, chaotic dynamical systems, and real-world network security datasets, demonstrate that our framework consistently satisfies prescribed  constraints where existing algorithms succumb to inflated missed detections and/or false alarms. Finally, because our methods operate on scalar errors without requiring online retraining or kernel evaluations, they are highly scalable.
\end{abstract}

\section{Introduction}
We consider a Decision Maker (DM) processing a sequence of observations generated by a  high-dimensional stochastic process $\mathcal{D}_N$ or $\mathcal{D}_A$ online to make inference decisions.  The two fundamental  problems considered are Sequential Hypothesis Testing (SHT), where the data-generating process remains constant and the objective is rapid classification, and Sequential Change Point Detection (SCPD), 
where the generative process suddenly shifts from $\mathcal{D}_N$ to $\mathcal{D}_A$ and the goal is to detect the change quickly without raising false alarms. 
  A core operational constraint in both settings is  information asymmetry: while abundant historical data  exists to characterize  $\mathcal{D}_N$, data from the alternative/post-change regime $\mathcal{D}_A$ is  unavailable. 

Although the theoretical foundations of sequential testing can be traced  decades back \cite{shewhart_chart,Wald1947,CUSUM_PAGE}, they remain vital in many fields, including cybersecurity~\cite{SPRTPortscan,SPRTSybil,SPRTMobileReplica,TartakovskyCPD_Network_Intrusion},  communications~\cite{QuickestDetectionCognitiveRadio}, and clinical trials~\cite{SequentialMedicalTrials}. However, classical formulations predominantly rely on independent and identically distributed (i.i.d.) observations and necessitate explicit likelihood specifications for both candidate regimes—an assumption that rarely holds in modern, high-dimensional applications. While recent research has sought to overcome these limitations via  data-driven frameworks e.g., ~\cite{NormalizingFlowSPRT,CUMAD,ahmed2025deepllrcusum,ScanB,kNNCPD}, existing state-of-the-art approaches still present significant shortcomings across both SHT and  SCPD. Many operate strictly under fully supervised paradigms (requiring labeled data from both $\mathcal{D}_N, \text{and } \mathcal{D}_A$), precluding their utility in open-world settings where alternative or post-change regimes are undefined a priori (e.g., novel zero-day intrusions or emergent pathologies). Others rely on rigid  hyperparameters that cannot be calibrated without supervision~\cite{CUMAD,B-Cave}, impose restrictive  assumptions~\cite{PCA_KDD_15,msSA}, or suffer from severe computational bottlenecks~\cite{NormalizingFlowSPRT,DAVT,kawahara2009change}.

To  address these challenges, we introduce a new class of distribution-free tests suitable for high-dimensional environments,  where data from  $\mathcal{D}_A$ is unavailable. More specifically, we propose the Clipped Unsupervised Residual Empirical Bernstein Sequential Hypothesis Test (CUREB-SHT) and the CUR-CUMulative SUM (CUR-CUSUM) procedure. Both tests operate by monitoring the clipped scalar errors of a base machine learning model trained exclusively on data from $\mathcal{D}_N$. Hence, they can be seamlessly deployed on high-dimensional  and temporally correlated streams without requiring any  labeled instances of $\mathcal{D}_A$.  Our approach leverages the principle of imperfect generalization: the model's expected error or error variance will reliably inflate when exposed to out-of-distribution data.
Beyond achieving strong experimental performance, we also derive theoretical guarantees. CUREB-SHT allows strict, non-asymptotic control over detection errors, and CUR-CUSUM provides  bounds on the Expected Detection Delay (EDD) and Average Run Length (ARL). 
Finally, our methods constitute lightweight wrappers for pretrained models, offering  improved complexity  over methods that compute kernel distances or retrain models online.

\section{Literature Review}
\paragraph{Sequential Hypothesis Testing}
\label{subsec:prelim_sht}
In the canonical SHT formulation, the DM seeks to determine whether the sequence of $d$-dimensional observations $y_1,y_2,\cdots$ is generated by a known distribution $P$ under the null hypothesis $H_0$ or $Q$ under the alternative $H_1$. 
The DM aims to reach an inference decision with as few samples as possible while ensuring that the False Positive (FP) probability (type I error) is below a constant $\alpha$ and the False Negative (FN) is below $\beta$. 
The optimal algorithm for SHT is 
 Wald's Sequential Probability Ratio Test (SPRT), using the Log-Likelihood Ratio (LLR):
\begin{equation}
    \Lambda_t = \sum_{\tau=1}^t \ln \frac{Q[y_\tau]}{P[y_\tau]}.
\end{equation}
The DM stops and accepts $H_0$ if $\Lambda_t \le A$, accepts $H_1$ if $\Lambda_t \ge B$, or continues sampling if $A < \Lambda_t < B$, where the thresholds $A \approx \ln \frac{\beta}{1-\alpha}$ and $B \approx \ln \frac{1-\beta}{\alpha}$ provably bound the error probabilities by $\alpha$ and $\beta$~\cite{Wald1947}.

Classical extensions designed for composite distributions  \cite{GSLRT_main,huber1965robust}   fundamentally depend on specifying parametric families (e.g., exponential family), which can be unrealistic in complex high-dimensional streams. Real-world data (e.g., network metadata or chaotic  systems) exhibit heavy-tailed distributions, complex  geometries, and  temporal dependencies. Forcing these data into rigid, mathematically tractable families leads to severe errors.

Recent  extensions attempt to bypass parametric modeling by approximating log-likelihoods via Neural Autoregressive Flows (NAF) \cite{NormalizingFlowSPRT}. However, this approach is supervised  and processes the expanding observation history through deep encoders at each step, incurring immense computational cost that precludes high-throughput streaming. 
Alternatively, Deep Anytime-Valid Hypothesis Testing (DAVT)~\cite{DAVT} provides non-parametric error control via game-theoretic betting with neural networks. Yet, DAVT requires pre-specified transformation operators to define null and alternative behavior. Moreover, it  relies on retraining neural networks at each time step, incurring prohibitive computational overhead in streaming environments.
Conversely, the unsupervised CUMmulative Anomaly Detection (CUMAD) framework \cite{CUMAD} restores $\mathcal{O}(1)$ updates by training an AutoEncoder (AE)  offline on nominal data, but binarizes the reconstruction errors into a Bernoulli sequence against a heuristic threshold  and treats the anomaly probabilities as rigid hyperparameters (fixed to $\theta_0= \Prob(y=1|H_0)=0.2, \theta_1= \Prob(y=1|H_1)=0.8$). This rigid binarization discards error magnitudes entirely, while relying on manually guessed $\theta_0, \theta_1$  hinders the detector's robustness and strips away error guarantees.
\paragraph{Sequential Change Point Detection}
\label{subsec:prelim_scpd}
In the SCPD framework, observations $y_t$ are generated by a nominal baseline distribution $P$ up until an unknown time $\tau^*$, after which they are generated by an alternative $Q$. The goal is to detect $\tau^*$ as quickly as possible (minimizing  EDD) while maintaining a large expected time between false alarms (ARL). 
The foundational algorithm for online CPD is the CUmulative Sum (CUSUM) procedure \cite{CUSUM_PAGE}, which recursively accumulates an increment score $s_t$:
\begin{equation}
    W_0 = 0, \quad W_t = \max\left(0, W_{t-1} + s_t \right).
\end{equation}
An alarm is triggered at stopping time $\hat{\tau} = \inf \{ t > 0 : W_t > h \}$. The optimal increment $s_t$ is the LLR; however, computing it requires perfect knowledge of $P$ and $Q$. While recent  adaptations approximate the LLR via deep networks \cite{ahmed2025deepllrcusum}, they remain strictly supervised.

Nonparametric methods evaluate  discrepancies over sliding windows. The Scan-B method \cite{ScanB} partitions data into reference and test blocks of size $B_0$, computing the Maximum Mean Discrepancy (MMD)~\cite{gretton2012MMD} via  kernel evaluations, improving older MMD-based change detectors, e.g., \cite{zaremba2013_Btest}.
Similarly, graph-based approaches  \cite{kNNCPD} track the proportion of $k$-Nearest Neighbor (k-NN) edges that span across a candidate split time within a sliding window of length $L$. However, these methods introduce severe operational hurdles for high-velocity streams. First, they incur prohibitive computational costs for pairwise distance and kernel evaluations in high-dimensional feature spaces. Furthermore, the success of MMD-based statistics hinges  on the selection of an appropriate kernel function, which is non-trivial for complex data manifolds and adds further computational overhead.

Beyond that, window-based estimators suffer from strict structural vulnerabilities. Their performance is  sensitive to the chosen block or window sizes ($B_0, L$): selecting a value that is too small renders the test statistic highly noisy, while a large window inherently forces an unacceptable detection delay. Moreover, the k-NN framework explicitly requires an initial historical baseline of $N_0$ clean observations to construct the reference neighbor graph, imposing the rigid assumption that the unknown change time occurs strictly after this initialization period ($\tau^* > N_0$). This restriction renders the detector completely blind to early-onset anomalies that occur shortly after deployment.

 An alternative  direction tracks deviations from nominal spatial or temporal geometry. The Principal Component Analysis (PCA) framework of \cite{PCA_KDD_15} monitors  divergences along top principal components using Page-Hinkley (PH) tests. To capture temporal dependencies, multivariate Singular Spectrum Analysis (mSSA) \cite{msSA} embeds lagged trajectory matrices into a pre-change linear subspace, updating a CUSUM statistic based on the orthogonal projection distance. 
Such  approaches share two fundamental weaknesses. First, they are strictly constrained by linearity assumptions, suffering severe performance degradation when applied to  complex nonlinear systems. Second, they depend heavily on rigid heuristic stopping rules. 
\paragraph{Connection to Anomaly Detection} The considered formulations are closely related to the problem of Anomaly Detection (AD). Specifically, SHT addresses settings where the anomaly is a persistent, sequence-wide phenomenon, whereas SCPD targets the detection of emergent/abrupt anomalies. Classic AD models~\cite{IsoFO,OC_SVM} process a fixed set of features to detect anomalies. Recent research has extended these approaches to sequential settings by tracking the errors of sequential models (e.g., recurrent networks) over sequences~\cite{LSTM_AD_1,LSTM_AD_2,VAE_LSTM_ICASSP,VAE_LSTM_ROBOT,T_CyB_RecAD}. However, in this direction, the  horizon is  fixed and there is no option for early termination when the status has been reliably inferred. Furthermore, these works  focus on empirical design without  mathematical guarantees, using heuristic decision rules. Alternatively, information-theoretic studies on sequential AD with rigorous guarantees are available e.g., \cite{Theory_AD_1,Theory_AD_2}, where the DM monitors a stream of observations generated from a known normal distribution $P$ or an unknown abnormal distribution. However, they impose restrictive parametric and structural assumptions on the anomaly, limiting their practicality.
\section{Proposed Methods}
\paragraph{SHT Problem}
We consider a binary hypothesis setting where
under $H_0$ the  observations are generated by  $\mathcal{D}_N$  and under $H_1$ by $\mathcal{D}_A$, with the structure of both data-generating processes unknown. Since in practice we do not know if or to what degree past observations influence the current one, we write the most general form:
\begin{equation}
    y_t\sim \begin{cases}
        \mathcal{D}_N(\cdot| y_{1:t-1}), \quad \text{$H_0$},\\
        \mathcal{D}_A(\cdot|y_{1:t-1}), \quad \text{$H_1$,}
    \end{cases}
\end{equation}
We  assume access to a large dataset $\mathcal{D}$ consisting exclusively of  trajectories generated by $\mathcal{D}_N$.\footnote{Specifically, we assume $\mathcal{D}$ is sufficiently large to effectively train the base model  and  compute accurate statistics for the null errors, without imposing  other requirements.} Using $\mathcal{D}$, we can train a model $\boldsymbol{\theta}$ such as an LSTM~\cite{LSTM} or a Transformer \textit{offline} to predict the most recent observation $y_t$ using historical data.  Alternatively, $\boldsymbol{\theta} $  can be a traditional  tracker like a Kalman Filter (KF)~\cite{KFOriginal}. For (approximately) i.i.d. data it can be an AutoEncoder (AE) whose reconstruction error we track~\cite{CUMAD}.  Let $e_t$ denote the   error of the trained model at time $t$:
\begin{equation}
e_t = || y_t - \mathrm{FP}(\mathcal{I}_t;\boldsymbol{\theta}) ||_2^2,
\end{equation}
where $\mathrm{FP}(\cdot ; \boldsymbol{\theta})$ denotes the forward-pass operation of the model on the relevant input $\mathcal{I}_t$. If a predictive architecture is chosen for  $\boldsymbol{\theta}$, then the input is historical data, e.g., $\mathcal{I}_t=y_{1:t}$. If a reconstructive model is employed, then $\mathcal{I}_t=y_t$.
Even though the raw observations $y_t$ may exhibit  temporal correlations, a highly representative, well-trained  $\boldsymbol{\theta}$ successfully learns these correlations. Consequently, the resulting tracking errors $e_t$ predominantly correspond to unstructured observation noise, rendering them approximately i.i.d.. To establish our primary mathematical framework, we formalize this ideal condition:
\begin{assumption}
    The error terms $\{e_t\}_{t \ge 1}$ are independent and identically distributed (i.i.d.).
\end{assumption}
This assumption streamlines our exposition, and we will explicitly detail how our core guarantees extend beyond it  at the conclusion of this section.   

To ensure robustness against potential isolated error spikes caused by transient noise, or numerical issues, we track the error $e_t$ clipped to the interval $[-C,C]$~\cite{huber_book_robust_stats}.
For notational simplicity, we henceforth abuse notation slightly: all subsequent references to the error $e_t$, refer exclusively to this clipped error.

The mean and variance of $e_t$ under the null and alternative hypothesis are denoted as $\mu_N,\sigma_N^2$ and $\mu_A,\sigma_A^2$ respectively. Because we assume access to a large training dataset $\mathcal{D}$ of null trajectories, $\mu_N$ and $\sigma_N^2$ can be estimated with negligible error during the model's training phase (developing tests that are strictly robust to these finite-sample estimation errors is left for future work), whereas the alternative hypothesis' parameters remain  unknown. 
CUREB-SHT computes the instantaneous zero-centered squared error:
\begin{equation}
    \phi_\tau = (e_\tau - \mu_N)^2 - \sigma_N^2.
\end{equation}
Then it evaluates the running additive test statistic:
\begin{equation}
    \label{eq:test_stat_SHT}
    f_t = \frac{1}{t} \sum_{\tau=1}^t \phi_\tau.
\end{equation}
This specific $\phi_t$ is strategically designed to translate raw tracking errors into a zero-mean process under $\mathcal{D}_N$. By computing the observed squared deviation $(e_\tau - \mu_N)^2$, the statistic inherently captures both mean shift and variance inflation in a single term. Under  $\mathcal{D}_N$ it holds that $\mathbb{E}_{\mathcal{D}_N}[\phi_\tau]=\mathbb{E}_{\mathcal{D}_N} [f_t]=0$. 
Conversely under $H_1$,   $\boldsymbol{\theta}$'s predictions inevitably degrade \footnote{We rely on \textit{imperfect generalization}. An unseen regime will worsen $\boldsymbol{\theta}$'s performance,  increasing  the error's mean or variance.  If it causes the  mean error to decrease or the variance to decrease under the same mean,  the proposed one-sided statistics will remain blind to the shift.}; this increases the error mean or variance, inducing a strictly positive expected drift.   
\begin{align}
\label{eq:alternate_shift}
    \Delta_A &\triangleq \mathbb{E}_{\mathcal{D}_A}[(e_\tau -\mu_N)^2]-\sigma^2_N \nonumber \\
    &=(\sigma_A^2-\sigma_N^2) +(\mu_A-\mu_N)^2 >0.
\end{align}
Using Hoeffding's inequality ~\cite{hoeffding1963probability} we can show  that  $f_t$ separates the two hypotheses with an exponential convergence rate, concentrating on  values below $\Delta_A/2$ under  $\mathcal{D}_N$, and  values above $\Delta_A/2$ under $\mathcal{D}_A$.
\begin{theorem}
    Under Assumption 1, it holds: 
    \begin{align}
        \mathbb{P}[f_t \leq \Delta_A / 2 \mid y_t \sim \mathcal{D}_A] &\leq e^{-\frac{\Delta_A^2}{32C^4}t}, \\
        \mathbb{P}[f_t > \Delta_A / 2 \mid y_t \sim \mathcal{D}_N] &\leq e^{-\frac{\Delta_A^2}{32C^4}t}.
    \end{align}
\end{theorem}
All proofs are contained in  Appendix \ref{ap:all_proofs}.

Because our setting is strictly unsupervised, $\Delta_A$ is unknown . To bridge this gap, we  use a  parameter,  the \textbf{Target Alternative Margin}, $\Delta_{\text{target}} > 0$. This parameter defines the ''minimum  detectable alternative profile'' CUREB-SHT is calibrated to notice: any perturbation inducing a shift $\Delta_A \ge \Delta_{\text{target}}$ is provably detected within the strict error tolerances $\alpha$ and $\beta$.
\paragraph{Two-Sided Sequential Decision Rules}
\label{subsec:sht_rules}
The DM evaluates the running statistic $f_t$ against two dynamic, time-varying boundaries: an upper threshold $\gamma_t$ and a lower threshold $\xi_t$ (with $\xi_t < \gamma_t$). If $f_t \ge \gamma_t$, the test terminates and rejects $H_0$ (declaring $H_1$); if $f_t \le \xi_t$, the test terminates and accepts $H_0$. Otherwise, while $\xi_t < f_t < \gamma_t$, the system advances to $t+1$.
Under this framework, the two operational errors are:
\begin{align}
    \text{FP:} \quad &P_{\mathrm{FP}}(t) \triangleq \mathbb{P}\left(f_t \ge \gamma_t \;\middle|\; H_0\right), \\
    \text{FN:} \quad &P_{\mathrm{FN}}(t) \triangleq \mathbb{P}\left(f_t \le \xi_t \;\middle|\; H_1\right).
\end{align}
While Hoeffding bounds provide valid finite-sample guarantees, they are fundamentally variance-blind. Their reliance on the extreme worst-case range ($4C^2$) yields overly conservative intervals that unnecessarily inflate the sample complexity required to accept $H_0$. To construct tighter, data-adaptive boundaries, we instead leverage the Empirical Bernstein bound~\cite{EmpiricalBernsteinBound_Initial,EmpiricalBernStein}. Let $\hat{\sigma}_t^2$ denote the empirical sample variance of $f_t$:
\begin{align}
    \hat{\sigma}_t^2 &= \frac{1}{t}\sum_{\tau=1}^t (\phi_\tau - f_t)^2 \nonumber \\
    &= \frac{1}{t}\sum_{\tau=1}^t \phi_\tau^2 - 2 f_t^2 + f_t^2 = \frac{1}{t}\sum_{\tau=1}^t \phi_\tau^2 - f_t^2.
\end{align}
By factoring in $\hat{\sigma}_t^2$, the Empirical Bernstein bound~\cite{EmpiricalBernStein} shrinks dynamically when the realized variance is substantially smaller than the theoretical worst-case bound $4C^4$ leading to quicker inference.  In the sequel, we state two theorems in which we prove error guarantees for CUREB-SHT using this bound.
\begin{theorem}
\label{thm:bernstein_FP}
Let the observations satisfy Assumption~1 with $\mathbb{E}_{\mathcal{D}_N}[\phi_\tau] = 0$. For any target FP level $\alpha \in (0, 1)$, the probability of an erroneous alarm at time $t$ satisfies $P_{FP}(t) \le \alpha$ when the dynamic upper threshold is set to:
\begin{equation}
    \gamma_t = \sqrt{\frac{2 \hat{\sigma}_t^2 \ln(3/\alpha)}{t}} + \frac{12 C^2 \ln(3/\alpha)}{t}.
\end{equation}
\end{theorem}

\begin{theorem}
\label{thm:bernstein_FN}
Let $\Delta_{\text{target}} > 0$ denote a user-specified minimum detectable margin. For any target FN level $\beta \in (0, 1)$ and any alternative hypothesis satisfying $\Delta_A \ge \Delta_{\text{target}}$, setting the dynamic lower threshold to:
\begin{equation}
    \xi_t = \Delta_{\text{target}} - \left( \sqrt{\frac{2 \hat{\sigma}_t^2 \ln(3/\beta)}{t}} + \frac{12 C^2 \ln(3/\beta)}{t} \right)
\end{equation}
guarantees that the Type~II error probability satisfies $P_{FN}(t) \le \beta$ at time $t$.
\end{theorem}
Crucially, Theorems \ref{thm:bernstein_FP} and \ref{thm:bernstein_FN} establish that the proposed detector provides strict, non-asymptotic error guarantees: the probability of an FP or an FN is bounded by $\alpha$ and $\beta$ at every time step $t$. These dynamic boundaries ($\gamma_t, \xi_t$) depend explicitly on the stream's empirical variance $\hat{\sigma}_t^2$, the   clipping range $C$, and the  error tolerances. Because the penalty term containing the worst-case bound $C^2$ decays at a rapid $\mathcal{O}(1/t)$ rate, it quickly diminishes compared to the $\mathcal{O}(1/\sqrt{t})$ empirical variance term. This ensures the thresholds are tightly governed by the actual dispersion of the data rather than the absolute clipping envelope. Consequently, in low-variance nominal streams ($\hat{\sigma}_t^2 \ll C^4$), rapid detection is achieved.

To guarantee  error control across the entire  monitoring horizon—rather than merely at fixed points—the proposed point bounds are extended to be anytime-valid. We adopt the time-decaying error allocation strategy  of \cite{EmpiricalBernStein},  substituting the static error budgets $\alpha$ and $\beta$ with time-dependent allocations  $\alpha_t = 6\alpha / (\pi^2 t^2)$ (and similarly for $\beta$). The union bound over all evaluation steps strictly guarantees the overall sequential error criteria.  Because the infinite sum $\sum_{t=1}^\infty 6/(\pi^2 t^2)$ converges exactly to 1, the cumulative FP probability over infinite time remains strictly bounded by $\alpha$ (i.e., $P(\exists t : f_t \ge \gamma_t \mid H_0) \le \alpha$). This simple correction ensures continuous, time-uniform error control.
\begin{table*}[!tb]
\centering
\caption{Performance Comparison of SHT Methods in 2D and 3D Nonlinear Tracking Environments. }
\label{tab:tracking_results}
\begin{tabular}{@{}lcccc@{}}
\toprule
\textbf{Method} & \textbf{FN (\%)} & \textbf{Delay ($H_1$)} & \textbf{FP (\%)} & \textbf{Delay ($H_0$)} \\ \midrule
\multicolumn{5}{c}{\textbf{2D Nonlinear Tracking Environment}} \\ \midrule
Data-Driven SPRT (Gaussian) & $95.7 \pm 0.8$ & $6.1 \pm 0.2$ & $7.8 \pm 0.8$ & $6.1 \pm 0.0$ \\
Oracle KF SPRT & $1.0 \pm 0.5$ & $9.2 \pm 0.1$ & $2.7 \pm 1.8$ & $10.2 \pm 0.1$ \\
\textbf{Proposed CUREB-SHT } & $0.0 \pm 0.0$ & $17.7 \pm 0.0$ & $0.6 \pm 0.2$ & $21.6 \pm 0.2$ \\ \midrule
\multicolumn{5}{c}{\textbf{3D Nonlinear Tracking Environment}} \\ \midrule
Data-Driven SPRT (Gaussian) & $95.1 \pm 0.7$ & $6.5 \pm 0.3$ & $3.5 \pm 0.8$ & $6.6 \pm 0.1$ \\
Oracle EKF SPRT & $0.2 \pm 0.2$ & $11.2 \pm 0.1$ & $0.0 \pm 0.0$ & $30.5 \pm 0.2$ \\
\textbf{Proposed CUREB-SHT} & $2.1 \pm 0.3$ & $21.4 \pm 0.4$ & $0.7 \pm 0.5$ & $22.0 \pm 0.1$ \\ \bottomrule
\end{tabular}
\end{table*}\\
\textbf{One-Sided Bootstrap Stabilization} 
During the initial monitoring steps (small $t$), the test statistic is highly noisy, creating a significant risk of prematurely accepting $H_0$. To mitigate this FN risk, we enforce a bootstrap period $T_{\text{boot}}$ only on null acceptance ($f_t \leq \xi_t$) that gives the statistic time to stabilize. If the sequence is sampled from $\mathcal{D}_A$, this delay ensures the statistic can establish its positive drift and eventually converge to a value safely above $\xi_t$. Immediate evaluation of the upper bound $\gamma_t$ starting at $t=1$ poses no risk of FPs. Because $\gamma_t$ naturally scales with the large initial variance estimates, the threshold remains exceedingly high for small $t$. Consequently, only trajectories that produce massive model errors possess the momentum to cross it early—a behavior that is inherently characteristic of the alternative data.
\paragraph{CPD Extension}
To detect abrupt structural changes from $\mathcal{D}_N$ to an unknown alternative process $\mathcal{D}_A$ online, we incorporate our error monitoring framework into a CUSUM procedure:
\begin{equation}
    W_0 = 0, \quad W_t = \max\left(0, W_{t-1} + \phi_t - c\right)
\end{equation}
where $c = \Delta_{\text{target}} / 2$ is the drift slack parameter determined by a user-defined target alternative margin $\Delta_{\text{target}}$. An alarm is triggered at the stopping time:
\begin{equation}
    \hat{\tau} = \inf\{t > 0 : W_t \ge h\}
\end{equation}
where $h$ is the decision threshold.
Because the bounded increments are strictly contained within an interval of total width $B = 4C^2$, we establish rigorous analytical guarantees relating the decision threshold $h$ to the ARL and the EDD.

\begin{theorem}
For a given target ARL $\text{ARL}_{\text{target}}$ under $\mathcal{D}_N$ (i.e. no change to $\mathcal{D}_A$), choosing the decision threshold as:
\begin{equation}
    h = \frac{B^2 \ln(\text{ARL}_{\text{target}})}{8c} = \frac{2C^4 \ln(\text{ARL}_{\text{target}})}{c}
\end{equation}
guarantees that the expected time between false alarms satisfies $\mathbb{E}_{\mathcal{D}_N}[\hat{\tau}] \ge \text{ARL}_{\text{target}}$.
\end{theorem}

\begin{theorem}
If the true expected post-change drift satisfies $\nu_A = \mathbb{E}_{\mathcal{D}_A}[\phi_t] - c > 0$, the EDD is bounded by:
\begin{equation}
    \text{EDD} \le \frac{h}{\nu_A} + O(1)
\end{equation}
\end{theorem}
While Assumption 1 significantly eases the mathematical exposition, it is not merely a theoretical over-idealization. Because a well-tuned base estimator actively learns the predictable temporal dynamics from the data, the remaining tracking errors are often  near-i.i.d. in practice—a hypothesis strongly supported by very strong experimental performance.    However, even when strict independence does not perfectly hold, the rigorous error guarantees of our proposed detectors remain intact. In Appendix \ref{app:empirical_bernstein_extensions}, we formally extend our analysis under  strictly weaker Martingale assumptions. The SCPD theorems hold exactly. Furthermore, while the SHT bounds exhibit minor algebraic differences when escaping the i.i.d. regime, they preserve the exact same asymptotic convergence rates and experimental behavior. 
\section{Numerical Experiments}\label{sec:main_Results}
We evaluate our algorithms across a diverse set of  environments (target kinematics, network traffic metadata, and chaotic time-series), pairing them with different base estimator architectures $\boldsymbol{\theta}$ tailored to each domain's structure. To show that our detector avoids tedious case-by-case hyperparameter tuning, an identical calibration strategy is maintained across all experiments: the bootstrap period for CUREB-SHT is fixed to $T_{\rm boot}=10$, the clipping threshold $C$ is consistently fixed to the 95th percentile of the errors observed during validation, and the target  margin is uniformly configured as $\Delta_{\text{target}} = 0.8C$. Furthermore, we set $\alpha = \beta = 0.05$, demanding at least 95\%  accuracy for both hypotheses.  All results are averaged over $10$ different initialization seeds, and detailed reproducibility information is available in Appendix \ref{ap:reproducibility}.



\paragraph{Target Tracking} 

We evaluate the proposed framework on tracking maneuvering targets in 2D and 3D space, where the objective is to rapidly decide whether a target follows its nominal flight pattern ($H_0$) or has initiated an  anomalous maneuver ($H_1$). The latent state $x_t$ tracks positions and velocities, while the surveillance system receives noisy position measurements $y_t$ under Additive White Gaussian Noise (AWGN).  
Because our framework is unsupervised, the base predictor $\boldsymbol{\theta}$ models strictly the  $H_0$ kinematics. 
In the \textbf{2D Setting}, the nominal trajectory  follows a linear Constant Velocity (CV) model, enabling exact state tracking via a standard linear KF \cite{KFOriginal}. The alternative  executes a nonlinear Coordinated Turn (CT) maneuver with dynamic angular acceleration. In the 
     \textbf{3D Setting},  both regimes are nonlinear: $H_0$ follows a steady 3D helical climb, while $H_1$ corresponds to an accelerating spiral. Here, $\boldsymbol{\theta}$ is an Extended KF(EKF)~\cite{EKF-ref1} configured for the nominal helical model.
Complete details are provided in supplementary Appendix \ref{app:tracking_kinematics}.

\textbf{Evaluated Benchmarks:} We compare CUREB-SHT against three baselines: (1) \textbf{Oracle SPRT}, an ideal baseline with knowledge of the true equations of motion for both $H_0$ and $H_1$, which computes exact Gaussian log-LLRs via two parallel filters; (2) \textbf{Data-Driven SPRT (Supervised)}, which fits empirical Gaussian density models over the errors using labeled trajectories from both regimes; and (3) \textbf{CUMAD}~\cite{CUMAD}, where the AE processes a sliding window of the $5$ most recent observations at each $t$.

The comparative results across 2D and 3D settings are summarized in Table~\ref{tab:tracking_results} \footnote{All SHT metrics are computed by running $1000$ Null and $1000$ alternative trajectories for each seed.}. While the supervised Data-Driven SPRT suffers from extreme missed detections  due to density miscalibration (e.g., empirical errors exhibiting heavier tails than the fitted Gaussian densities),  and CUMAD consistently violates the error constraints regardless of parameter selection (detailed results for different CUMAD configurations are provided in Appendix \ref{ap:cumad_res_tracking}), CUREB-SHT strictly respects the prescribed thresholds $\alpha = \beta = 0.05$. 
\paragraph{Neural Base Models}
\begin{table}[t]
\centering
\caption{SHT Results with Neural Base Models}
\label{tab:SHT_neural_base_models}
\resizebox{\columnwidth}{!}{%
\begin{tabular}{lcc}
\toprule
\textbf{Metric} & \textbf{Proposed} & \textbf{Data-Driven SPRT} \\ 
\midrule
\multicolumn{3}{c}{\textit{Network Traffic Tracking: CIC-IDC2018}} \\ 
\midrule
\textbf{FN (\%)} & $0.40 \pm 0.20$ & $31.00 \pm 3.60$ \\
\textbf{FP (\%)} & $0.80 \pm 1.10$ & $1.00 \pm 2.00$ \\
\textbf{Delay ($H_1$)} & $6.00 \pm 0.00$ & $4.73 \pm 0.89$ \\
\textbf{Delay ($H_0$)} & $11.35 \pm 0.22$ & $2.27 \pm 1.11$ \\ 
\midrule
\multicolumn{3}{c}{\textit{Network Traffic Tracking: CESNET24}} \\ 
\midrule
\textbf{FN (\%)} & $1.00 \pm 0.35$ & $14.80 \pm 3.20$ \\
\textbf{FP (\%)} & $1.00 \pm 0.21$ & $5.52 \pm 3.52$ \\
\textbf{Delay ($H_1$)} & $13.54 \pm 1.33$ & $1.30 \pm 0.00$ \\
\textbf{Delay ($H_0$)} & $12.86 \pm 0.39$ & $1.21 \pm 0.55$ \\ 
\midrule
\multicolumn{3}{c}{\textit{Chaotic Systems: H{\'e}non Map}} \\ 
\midrule
\textbf{FN (\%)} & $4.60 \pm 3.30$ & $59.50 \pm 10.20$ \\
\textbf{FP (\%)} & $0.20 \pm 0.01$ & $3.67 \pm 0.52$ \\
\textbf{Delay ($H_1$)} & $9.91 \pm 1.35$ & $1.41 \pm 0.15$ \\
\textbf{Delay ($H_0$)} & $16.65 \pm 0.05$ & $1.12 \pm 0.30$ \\ 
\midrule
\multicolumn{3}{c}{\textit{Chaotic Systems: Lorenz-Chen System}} \\ 
\midrule
\textbf{FN (\%)} & $0.00 \pm 0.00$ & $12.50 \pm 0.22$ \\
\textbf{FP (\%)} & $3.22 \pm 0.87$ & $2.65 \pm 0.41$ \\
\textbf{Delay ($H_1$)} & $5.42 \pm 0.55$ & $1.10 \pm 2.13$ \\
\textbf{Delay ($H_0$)} & $10.00 \pm 0.00$ & $1.10 \pm 0.00$ \\ 
\bottomrule
\end{tabular}%
}
\end{table}
\begin{figure*}[!t]
    \centering
    \begin{subfigure}[b]{0.24\textwidth}
        \centering
        \includegraphics[width=\linewidth]{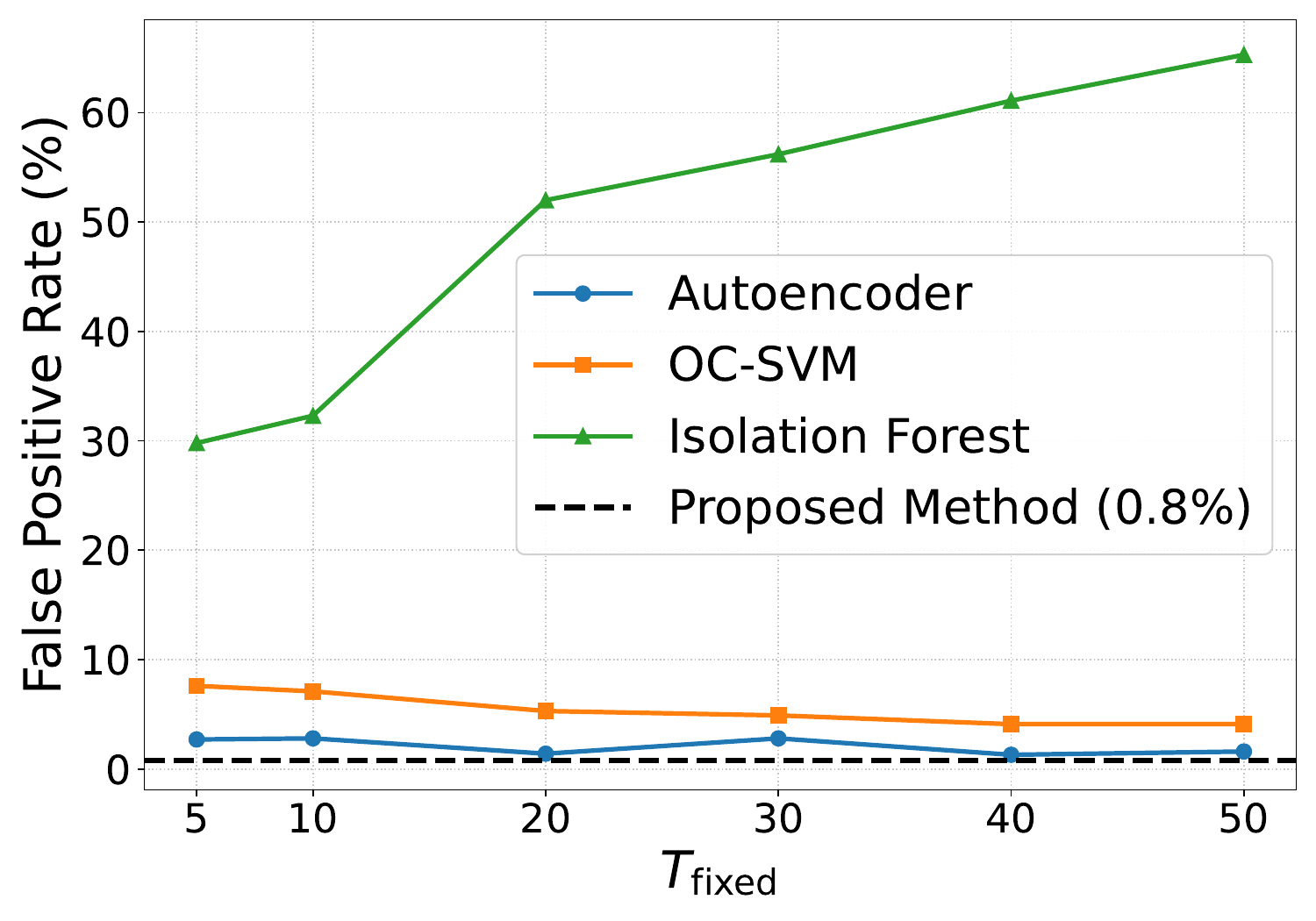}
        \caption{CIC-IDS-2018}
        \label{fig:fp_network}
    \end{subfigure}\hfill
    \begin{subfigure}[b]{0.24\textwidth}
        \centering
        \includegraphics[width=\linewidth]{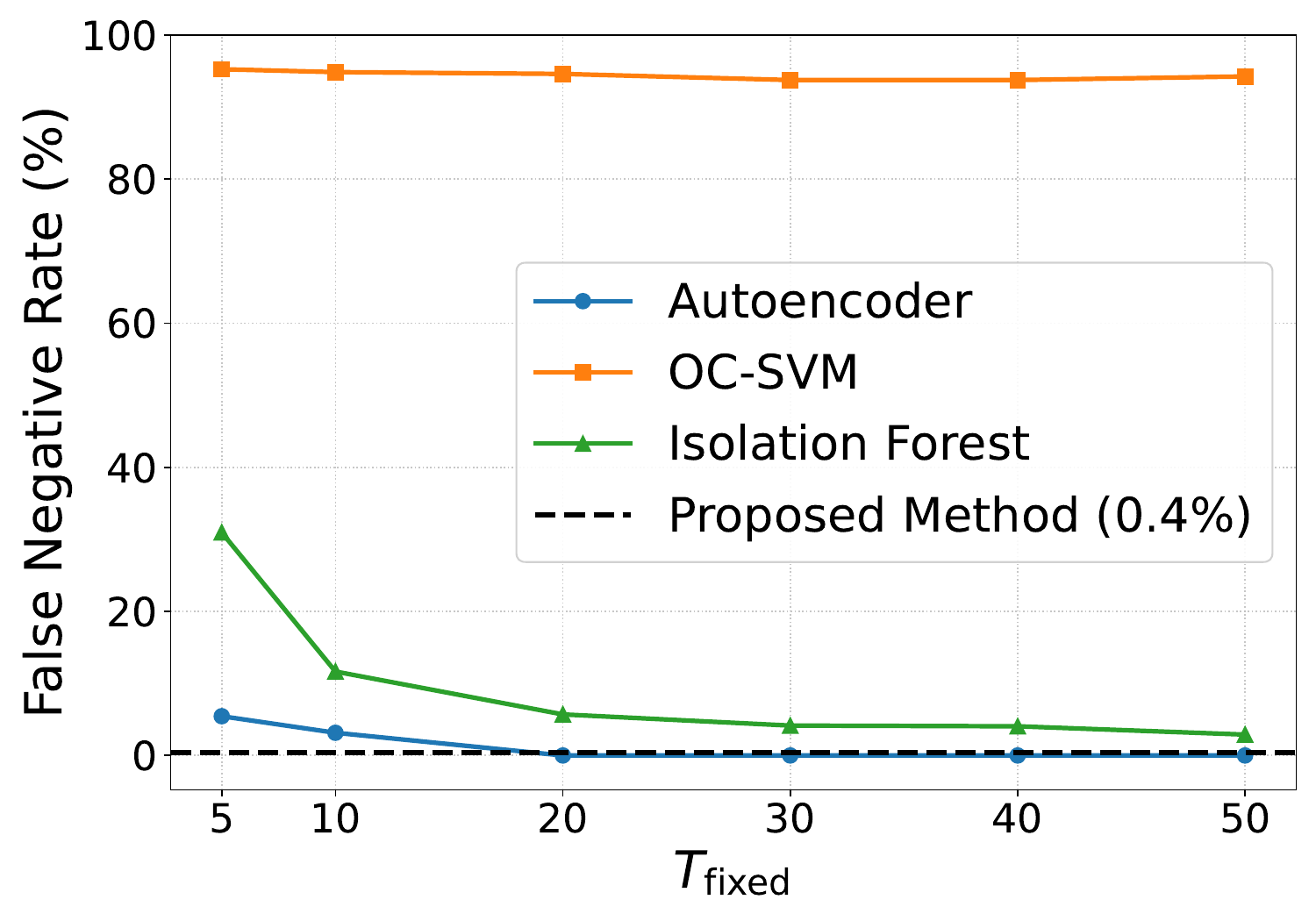}
        \caption{CIC-IDS-2018}
        \label{fig:fn_network}
    \end{subfigure}\hfill
    \begin{subfigure}[b]{0.24\textwidth}
        \centering
        \includegraphics[width=\linewidth]{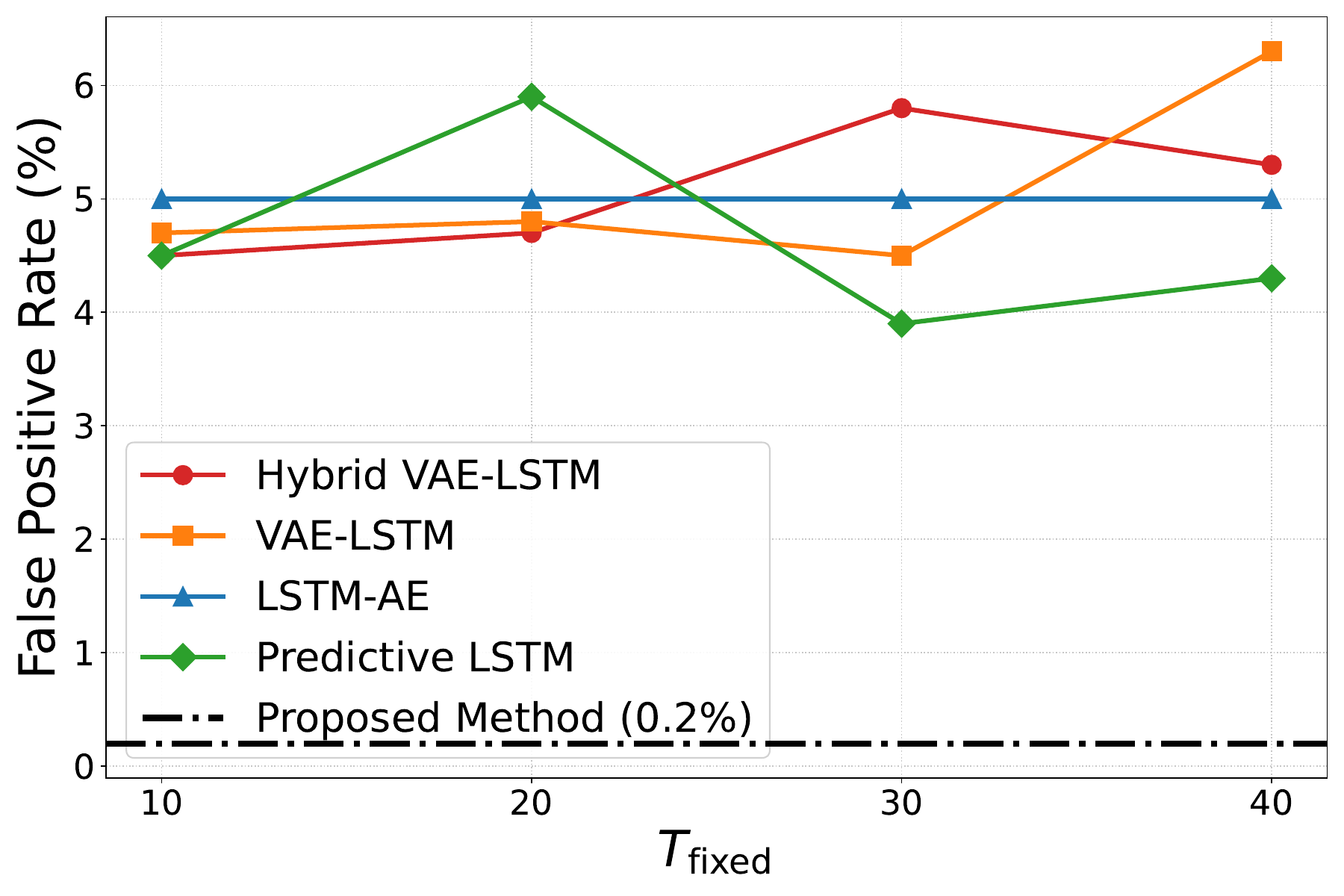}
        \caption{H{\'e}non}
        \label{fig:fp_henon}
    \end{subfigure}\hfill
    \begin{subfigure}[b]{0.24\textwidth}
        \centering
        \includegraphics[width=\linewidth]{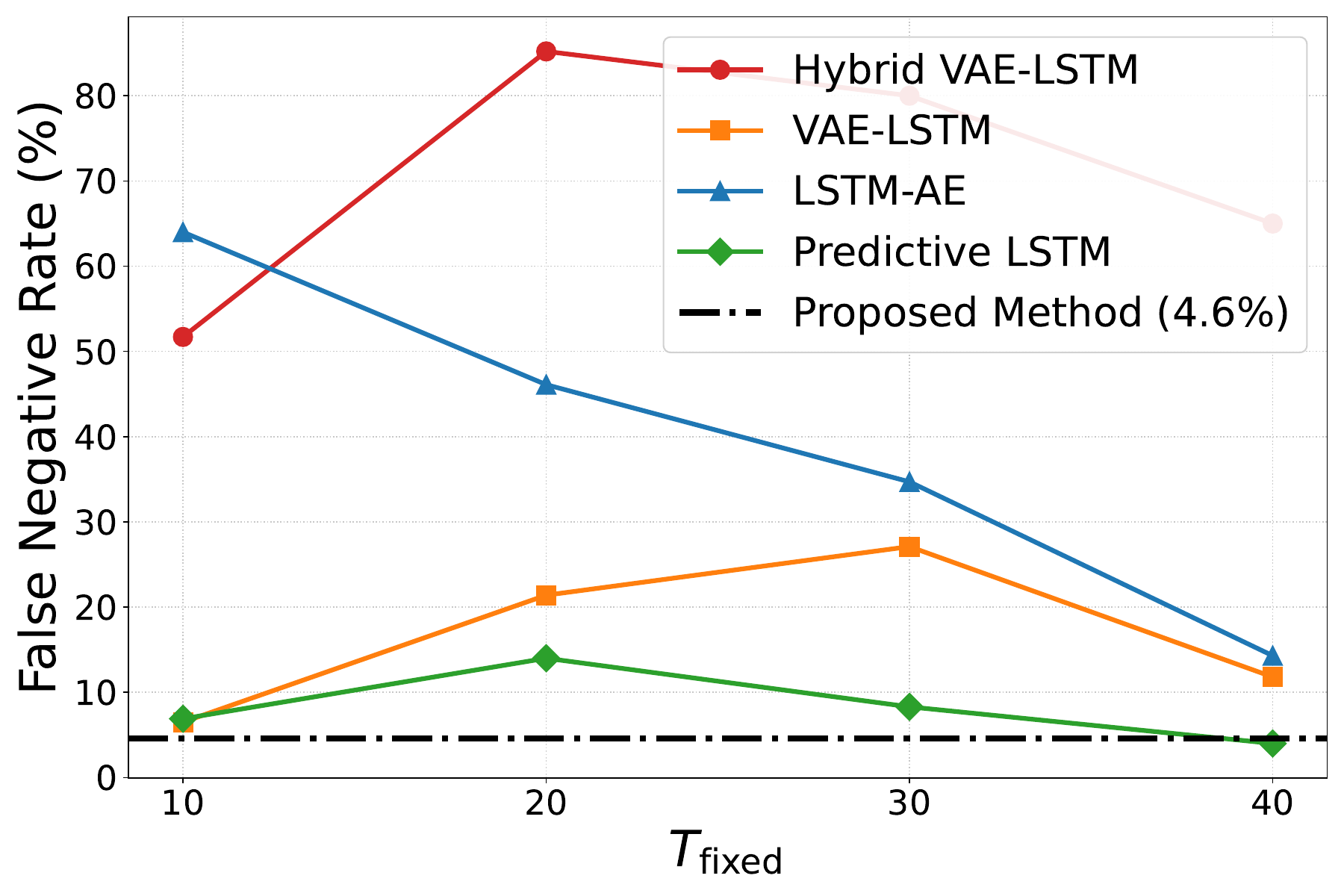}
        \caption{H{\'e}non}
        \label{fig:fn_henon}
    \end{subfigure}
    \caption{Error rates vs.\ fixed-horizon length $T_{\text{fixed}}$ for AD benchmarks.}
    \label{fig:ad_baselines_comparison}
\end{figure*}
Having successfully established the effectiveness of the proposed SHT when operating with traditional estimators as base models, we will now evaluate it using two different neural  network architectures as base models. 
First, to evaluate the framework on real-world  data, we apply it to network intrusion detection using two  traffic metadata datasets. We utilize CIC-IDS-2018~\cite{aws_csecicids2018}, which tracks high-dimensional flow statistics, e.g., packet lengths and inter-arrival times at the network perimeter, alongside CESNET-24~\cite{cesnettimeseries24}, which monitors lower-dimensional host-level behavior. For the first dataset, evaluation episodes under the null hypothesis are formed by sampling contiguous sequences of benign network traffic, whereas under the alternative, they consist of contiguous sequences of Denial-of-Service (DoS) attacks. In the CESNET-24 experiment, null episodes are generated by sampling contiguous sequences corresponding to the behavior of an authorized host IP (device ID 100610), while alternative trajectories are drawn from another  "impostor" device (device ID 10125). For both experiments, the maximum sequence length is set to $T_{\max}=300$, with undeclared episodes counted as errors.

Furthermore, we consider time-series data with a forecasting LSTM as  base model. We construct two distinct experiments based on  chaotic systems. In the discrete-time setting, the null hypothesis corresponds to a 2D Hénon map with parameters $(a=1.35, b=0.2)$, while the alternative to parameters $(a=1.35, b=0.3)$. In the continuous-time setting, the  null hypothesis corresponds to  3D Lorenz attractor, and $H_1$ to a  Chen attractor, where the sequences are generated via numerical integration of their respective polynomial differential equations. Full   data generation procedures are deferred to Appendix \ref{ap:Chaotic_Generation}. Such systems serve as a rigorous benchmark in   machine learning literature because their complexity can be controlled to be arbitrarily difficult, whereas real-world time-series are often heavily preprocessed to ease estimation \cite{lorenzNeurIPS}.

The performance of our method vs the data-driven SPRT is provided in Table~\ref{tab:SHT_neural_base_models}. CUMAD\footnote{For the time series case, the CUMAD logic is applied on the LSTM errors, using the same rigid thresholding rule, and $\theta_0,\theta_1$ parameterization of likelihoods as \cite{CUMAD}.} again performed poorly and could not satisfy the error constraints, with the results for all considered $\theta_0,\theta_1$ being delegated to Appendices~\ref{ap:CUMAD_NetworkTraffic}--\ref{ap:CUMAD_res_Lorenz_Chen}. Across all benchmarks, the proposed method strictly abides by the prescribed error bounds, reducing error rates to negligible levels ($0.00\%$--$4.60\%$) compared to the data-driven SPRT (FN rates in $12.5\%-59.5\%$). The latter exhibits quicker termination, but this is attributed solely to its extremely inflated FNs.


\begin{table*}[!t]
    \centering
    \caption{ CPD Performance Comparison on Target Tracking ( 1000 Monte Carlo Simulations).}
\label{tab:tracking_results_cpd}
    \begin{tabular}{@{}llccc@{}}
        \toprule
        \textbf{Detector Framework} & \textbf{Parameters} & \textbf{Early Terminations} & \textbf{Delay} & \textbf{Missed (\%)} \\
        \midrule
        
        \multirow{4}{*}{\shortstack[l]{$k$-NN \\ (Raw Observations)}} 
        & $L=10, k=3$ & 0.5\% $\pm 1.86\%$  & $42.50 \pm 5.52$ & $48.5\% \pm 3.21\%$ \\
        & $L=20, k=3$ & 0.0\% $\pm 2.01\%$ & $30.79 \pm 2.62$& $14.0\% \pm 2.52\%$ \\
        & $L=10, k=5$ & 2.4\% $\pm 5.52\%$ & $41.25 \pm 4.21$ & 87.4\% $\pm 0.30\%$ \\
        & $L=20, k=5$ & 1.4\% $\pm 2.13\% $ & $29.59 \pm 1.06$  & 16.5\% $\pm 1.98\%$ \\
        \midrule
        
        \multirow{4}{*}{\shortstack[l]{$k$-NN \\ (KF Residuals)}}
        & $L=10, k=3$ & 2.5\% $\pm 0.87\%$& 27.93 $\pm 6.62$ & 76.5\% $\pm 0.01\%$\\
        & $L=20, k=3$ & 0.5\% $\pm 0.0\%$ & 20.93  $\pm 4.81$ & 25.0\%  $\pm 0.25\%$\\
        & $L=10, k=5$ & 0.5\%  $\pm 0.0\%$ & 26.04 $\pm 2.81$& 87.0\% $\pm 0.00\%$\\
        & $L=20, k=5$ & 3.5\%  $\pm 0.63\%$& 29.72 $\pm 0.02$  & 13.0\% $\pm 0.00\%$ \\
        \midrule
        
        \multirow{4}{*}{\shortstack[l]{Scan-B \\ (KF Residuals)}}
        & $B_0=5$  & 0.0\%  $\pm 0.0\%$ & $44.53 \pm 11.95$ & $3.3 \pm 2.47\%$ \\
        & $B_0=10$ & 0.0\% $\pm 0.0\%$ & $20.62 \pm 3.52\phantom{0}$  & $0.5 \pm 0.50\%$ \\
        & $B_0=20$ & 0.0\% $\pm 0.0\%$ & $20.77 \pm 0.37\phantom{0}$  & $0.0 \pm 0.12\%$ \\
        & $B_0=30$ & 0.0\% $\pm 0.0\%$ & $25.95 \pm 0.36\phantom{0}$  & $0.0 \pm 0.00\%$ \\
        \midrule
        
        \textbf{Proposed Framework} & \textbf{ARL = 5K} & \textbf{0.0\%}  $\mathbf{\pm 0.0\%}$& $\mathbf{20.51 \pm 0.03\phantom{0}}$ & \textbf{0.0} $\pm$ \textbf{0.00\%} \\
        \bottomrule
    \end{tabular}
\end{table*}
\textbf{Anomaly Detection Methods}\footnote{Implementation details  are provided in Appendix \ref{ap:reproducibility}.}
This experiment compares our dynamic approach to standard AD pipelines constrained to a fixed window $T_{\rm fixed}$. For  network traffic data, we will consider an AE     trained  on normal network traffic to minimize reconstruction error. The validation data is segmented into rigid windows of  fixed length $T_{\rm fixed}$. Every instance within each window is processed by the model, and the resulting instance-level reconstruction errors are averaged to yield a single sequence-level score. A decision threshold is calibrated at a target percentile (95th).
During deployment, if the  score of a  sequence exceeds this threshold, it is classified as an anomaly.
Besides the AE we also considered  a  linear One-Class SVM~\cite{OC_SVM} and an Isolation Forest~\cite{IsoFO} in the same set-up. Instead of reconstruction error, instance-level anomaly scores for these models are derived from their  decision functions. 
The CIC-IDS-2018-based setup was considered and as figures  \ref{fig:fp_network} and \ref{fig:fn_network}
 reveal, the proposed approach yields  lower errors, without the need to configure an exact stopping horizon. Even though the  fixed-length AE can match  CUREB-SHT's accuracy, it happens only for $T_{\rm fixed} \geq 20$ which is substantially larger than our method's termination times under both hypotheses ($6$ and $13.35$ respectively).

We then repeated the  same study using  LSTM-based fixed length  anomaly detectors on the H\'enon map experiment, again configured on a percentile of scores from $H_0$ trajectories.  We considered four models: (1) a  Many-to-One Predictive LSTM that deterministically forecasts the final step of a sequence, (2) a deterministic Sequence-to-Sequence LSTM Autoencoder (LSTM-AE) \cite{LSTM_AD_2}, (3) a probabilistic LSTM-VAE that maps the temporal window to a latent distribution prior to reconstruction \cite{VAE_LSTM_ROBOT}, and (4) a hybrid VAE-LSTM model that sequentially applies an LSTM over locally extracted VAE embeddings \cite{VAE_LSTM_ICASSP}.  The results are reported in Figs. \ref{fig:fp_henon} and \ref{fig:fn_henon}  where it is evident that CUREB-SHT is more accurate. All considered AD models yield substantially higher FP rates.

\paragraph{Change Point Detection}\label{sec:CPD_main_res}
Revisiting the 3D target tracking experiment, trajectories now begin with the nominal helical motion and abruptly transition to the accelerated spiral at $\tau^*=50$. We enforce a maximum  horizon of $200$ steps—after which undetected changes are classified as misses—and wherever mathematically feasible, calibrate the sequential tests to a target ARL of $5000$. 
For the k-NN baseline~\cite{kNNCPD}, we evaluate sliding windows $L \in \{10,20\}$ with $k \in \{3,5\}$, granting the detector oracle knowledge that $\tau^* > 20$ to initialize its reference graph. For the Scan-B test, we use block sizes $B_0 \in \{5, \dots, 30\}$ with a Radial Basis Function (RBF) kernel. To isolate the advantage of our \textit{unified}  formulation, versus simply tracking the model's errors with existing methods, we evaluate both k-NN and Scan-B twice: directly on the raw kinematic observations and on the EKF error residuals. 

Results are reported in Table~\ref{tab:tracking_results_cpd}. Raw Scan-B and PCA-based detectors \cite{PCA_KDD_15,msSA} are excluded due to systemic detection failures\footnote{A visualization of their performance on this task is provided in Appendix \ref{ap:raw_scanB_PCA_tracking}.}.
CUR-CUSUM consistently detects the maneuver without false alarms and an average delay of approximately $20.5$ steps, substantially outperforming all k-NN configurations. Although Scan-B applied to residuals achieves barely larger delays for optimally tuned block sizes ($B_0 \in \{10, 20\}$), it still suffers from a small number of misses. We must note that in practice tuning its block size to minimize delay and misses requires using post-change data.

To confirm the effectiveness of CUR-CUSUM  across diverse datasets, we conduct additional SCPD evaluations on network traffic and chaotic systems. As detailed in Appendix \ref{ap:extra_cpd_experiments},  CUR-CUSUM reliably detects changes in both cases, whereas the considered benchmarks  suffer from extreme variance inflation, rendering threshold calibration practically impossible.

\textbf{Discussion}
Our evaluations reveal three key insights: 
First,
 \textbf{parametric assumptions fail:} Analytical densities (Data-Driven SPRT) cannot handle highly non-Gaussian streaming errors, causing severe misses.
Secondly, \textbf{mathematical thresholding is vital:}  Heuristic stopping rules (e.g., \cite{CUMAD,PCA_KDD_15}) systematically violate error constraints, whereas our  methods maintain strict control. 
Finally, we believe that \textbf{well-calibrated models yield near-i.i.d. errors:} By extracting the learnable temporal dynamics, quality estimators leave primarily noise. The  empirical accuracy of our algorithms leads us to hypothesize that these residuals practically satisfy Assumption 1, or at least the Martingale assumptions of Appendix \ref{app:empirical_bernstein_extensions}.
\section{Conclusion}
We introduced a novel distribution-free framework for SHT and SCPD in complex streaming environments. Our framework operates in an entirely alternative-agnostic manner monitoring the errors of an unsupervised model. We provide mathematical results for error guarantees, and empirically verify the effectiveness of the two statistics in multiple applications. 

This work opens several exciting directions for future research. First, while CUREB-SHT and CUR-CUSUM provide strict error guarantees, they are by no means optimal in terms of detection delay. As research in  learning theory progresses~\cite{estimatingMeansByBetting,timeUniformAnalStats,EmpiricalBernsteinSharp}, methods that enable faster termination may be uncovered. 
Furthermore, more advanced statistics that track high-order error moments or rely on other assumptions  could be explored.  Finally, because our  analysis assumes that the null or pre-change error statistics are estimated  with negligible error, a critical next step is to rigorously quantify the impact of finite-sample estimation errors on the provided theoretical guarantees.
\section*{AI Use Statement}
In this work, we used generative AI tools to provide feedback on experiments, clean up result presentation/figures, suggest experimental parameters,  identify relevant literature, and detect  writing issues in order to improve presentation.  We have not used generative AI tools for other tasks required for disclosure, such as formulating mathematical claims, writing code, proofs,  or drafting entire segments.  We take responsibility for the final content of this work, including text, claims, or artifacts
produced with the aid of generative AI.
\bibliographystyle{ieeetr}
\bibliography{references}
\section*{Checklist}


\textbf{In your paper, please delete this instructions block and only keep the Checklist section heading above along with the questions/answers below.}

\begin{enumerate}

  \item For all models and algorithms presented, check if you include:
  \begin{enumerate}
    \item A clear description of the mathematical setting, assumptions, algorithm, and/or model. [Yes]\\ Assumptions 1, 2 and the surrounding text regarding imperfect generalization. The implementation details for all machine learning models are provided in the Supplementary Appendix \ref{ap:reproducibility}.
    \item An analysis of the properties and complexity (time, space, sample size) of any algorithm. [Yes] \\ The complexity is straight-forward, as the algorithms just track the scalar errors of existing unsupervised models according to simple equations. 
    \item (Optional) Anonymized source code, with specification of all dependencies, including external libraries. [No]\\ We have a  Supplementary Appendix with all implementation details. Software can be released upon acceptance.
  \end{enumerate}

  \item For any theoretical claim, check if you include:
  \begin{enumerate}
    \item Statements of the full set of assumptions of all theoretical results. [Yes]
    \item Complete proofs of all theoretical results. [Yes]\\
    All proofs are rigorously detailed in the Supplementary Appendix.
    \item Clear explanations of any assumptions. [Yes]
  \end{enumerate}

  \item For all figures and tables that present empirical results, check if you include:
  \begin{enumerate}
    \item The code, data, and instructions needed to reproduce the main experimental results (either in the supplemental material or as a URL). [Yes]
    \item All the training details (e.g., data splits, hyperparameters, how they were chosen). [Yes]
    \item A clear definition of the specific measure or statistics and error bars (e.g., with respect to the random seed after running experiments multiple times). [Yes]
    \item A description of the computing infrastructure used. (e.g., type of GPUs, internal cluster, or cloud provider). [Yes] \\
    Please see the beginning of Supplementary Appendix \ref{ap:reproducibility}.
  \end{enumerate}

  \item If you are using existing assets (e.g., code, data, models) or curating/releasing new assets, check if you include:
  \begin{enumerate}
    \item Citations of the creator If your work uses existing assets. [Yes] \\
    We use two popular public  datasets with citation.
    \item The license information of the assets, if applicable. [Not Applicable]
    \item New assets either in the supplemental material or as a URL, if applicable. [Not Applicable]
    \item Information about consent from data providers/curators. [Not Applicable]
    \item Discussion of sensible content if applicable, e.g., personally identifiable information or offensive content. [Not Applicable]
  \end{enumerate}

  \item If you used crowdsourcing or conducted research with human subjects, check if you include:
  \begin{enumerate}
    \item The full text of instructions given to participants and screenshots. [Not Applicable]
    \item Descriptions of potential participant risks, with links to Institutional Review Board (IRB) approvals if applicable. [Not Applicable]
    \item The estimated hourly wage paid to participants and the total amount spent on participant compensation. [Not Applicable]
  \end{enumerate}

\end{enumerate}

\clearpage
\appendix
\thispagestyle{empty}

\onecolumn
\aistatstitle{Supplementary Material}




\section{Proofs}\label{ap:all_proofs}
\subsection{Mathematical Prerequisites}
\label{app:prerequisites}

We first state the foundational concentration inequalities and martingale theorems utilized throughout the subsequent proofs.

\begin{lemma}[Hoeffding's Inequality~\cite{hoeffding1963probability}]
Let $Z_1, Z_2, \dots, Z_t$ be independent random variables such that $Z_\tau \in [a_\tau, b_\tau]$ almost surely for all $\tau \in \{1, \dots, t\}$. Let $\bar{Z}_t = \frac{1}{t} \sum_{\tau=1}^t Z_\tau$ denote their empirical average. Then, for any $\epsilon > 0$:
\begin{align}
    \mathbb{P}\left(\bar{Z}_t - \mathbb{E}[\bar{Z}_t] \ge \epsilon\right) &\le \exp\left(-\frac{2 t^2 \epsilon^2}{\sum_{\tau=1}^t (b_\tau - a_\tau)^2}\right), \\
    \mathbb{P}\left(\bar{Z}_t - \mathbb{E}[\bar{Z}_t] \le -\epsilon\right) &\le \exp\left(-\frac{2 t^2 \epsilon^2}{\sum_{\tau=1}^t (b_\tau - a_\tau)^2}\right).
\end{align}
\end{lemma}

\begin{lemma}[Empirical Bernstein Bound]
\label{lem:empirical_bernstein}
Let $Z_1, Z_2, \dots, Z_t$ be i.i.d. random variables taking values in a bounded interval with range $R$ almost surely. Let $\mu = \mathbb{E}[Z_i]$, $\bar{Z}_t = \frac{1}{t}\sum_{\tau=1}^t Z_\tau$, and let $\hat{\sigma}_t^2 = \frac{1}{t}\sum_{\tau=1}^t (Z_\tau - \bar{Z}_t)^2$ denote the empirical sample variance. Then, for any $\delta \in (0, 1)$, the one-sided tail deviations satisfy:
\begin{align}
    \label{eq:eb_upper_tail}
    \mathbb{P}\left(\bar{Z}_t - \mu \ge \hat{\sigma}_t \sqrt{\frac{2 \ln(3/\delta)}{t}} + \frac{3 R \ln(3/\delta)}{t}\right) &\le \delta, \\
    \label{eq:eb_lower_tail}
    \mathbb{P}\left(\mu - \bar{Z}_t \ge \hat{\sigma}_t \sqrt{\frac{2 \ln(3/\delta)}{t}} + \frac{3 R \ln(3/\delta)}{t}\right) &\le \delta.
\end{align}
\end{lemma}

\begin{lemma}[Freedman's Inequality, \cite{freedman1975tail}]
\label{lem:freedman}
Let $\{Y_\tau\}$ be a Martingale Difference Sequence (MDS) bounded by $R$. Let $M_t = \sum_{\tau=1}^t Y_\tau$ and $V_t = \sum_{\tau=1}^t \mathbb{E}[Y_\tau^2 \mid \mathcal{F}_{\tau-1}]$. For any fixed scalar $v > 0$ and any threshold $x > 0$:
\begin{equation}
    \mathbb{P}\left( M_t \ge x \text{ and } V_t \le v \right) \le \exp\left( - \frac{x^2}{2(v + Rx/3)} \right)
\end{equation}
\end{lemma}

\begin{lemma}[Azuma-Hoeffding Inequality, \cite{Azuma}]
\label{lem:azuma}
Let $\{M_t\}_{t \ge 0}$ be a martingale with respect to a filtration $\{\mathcal{F}_t\}_{t \ge 0}$. If its increments $Z_t = M_t - M_{t-1}$ are bounded almost surely such that $|Z_t| \le c_t$ for some constants $c_t$, then for any $\epsilon > 0$:
\begin{equation}
    \mathbb{P}\left( M_t - M_0 \ge \epsilon \right) \le \exp\left( - \frac{\epsilon^2}{2 \sum_{\tau=1}^t c_\tau^2} \right)
\end{equation}
\end{lemma}
When this inequality is applied to sequences with uniformly bounded increments (i.e., $c_\tau = U$ for all $\tau$), the cumulative variance term simplifies to $\sum_{\tau=1}^t c_\tau^2 = t U^2$. This yields the simplified form utilized throughout our derivations:
\begin{equation*}
    \mathbb{P}\left( M_t - M_0 \ge \epsilon \right) \le \exp\left( - \frac{\epsilon^2}{2 t U^2} \right).
\end{equation*}
Furthermore, when this result is applied specifically to a Martingale Difference Sequence (MDS) $\{Z_\tau\}$ initialized at zero, the cumulative sum is given by $M_t = \sum_{\tau=1}^t Z_\tau$ (with $M_0 = 0$). Substituting these conditions yields:
\begin{equation*}
    \mathbb{P}\left( \sum_{\tau=1}^t Z_\tau \ge \epsilon \right) \le \exp\left( - \frac{\epsilon^2}{2 t U^2} \right).
\end{equation*}
\begin{lemma}[Doob's Optional Stopping Theorem, \cite{doob1953stochastic}]
\label{lem:doobs}
Let $\{M_t\}_{t \ge 0}$ be a submartingale with respect to a filtration $\{\mathcal{F}_t\}_{t \ge 0}$, and let $N$ be a valid stopping time. If the expected stopping time is finite ($\mathbb{E}[N] < \infty$) and the increments $|M_t - M_{t-1}|$ are bounded almost surely, then the expected value at the stopping time satisfies:
\begin{equation}
    \mathbb{E}[M_N] \ge \mathbb{E}[M_0].
\end{equation}
\end{lemma}

Doob's Optional Stopping Theorem admits several alternative sufficient conditions, such as an almost surely bounded stopping time. We explicitly invoke the condition requiring both a finite expected stopping time and bounded increments. In our framework, the strictly positive drift of the unconstrained sequence guarantees that the expected stopping time is finite, while the clipping mechanism inherently ensures that the step sizes of the test statistic remain strictly bounded.

\begin{lemma}[Hoeffding's Lemma]
\label{lem:hoeffding_lemma}
Let $X$ be a bounded random variable such that $X \in [a, b]$ almost surely, and let $\mathbb{E}[X] \le \mu$. Then, for any $\lambda > 0$, the moment-generating function of $X$ is bounded by:
\begin{equation}
    \mathbb{E}[e^{\lambda X}] \le \exp\left( \lambda \mu + \frac{\lambda^2 (b-a)^2}{8} \right).
\end{equation}
\end{lemma}

\begin{lemma}[Ville's  Inequality, \cite{VillePHD}]
\label{lem:ville}
Let $\{M_t\}_{t \ge 0}$ be a non-negative supermartingale with respect to a filtration $\{\mathcal{F}_t\}_{t \ge 0}$. For any constant $A > 0$, the probability that the process ever exceeds $A$ is bounded by:
\begin{equation}
    \mathbb{P}\left( \sup_{t \ge 0} M_t \ge A \right) \le \frac{\mathbb{E}[M_0]}{A}.
\end{equation}
\end{lemma}
\subsection{Theorem 1}
\label{app:proof_theorem1}
Recall from \eqref{eq:test_stat_SHT} that the test statistic is defined as:
\begin{equation}
    f_t = \frac{1}{t} \sum_{\tau=1}^t \phi_\tau,
\end{equation}
where $\phi_\tau = (e_\tau - \mu_N)^2 - \sigma_N^2$. By our established convention, $e_\tau$ is strictly bounded in $[-C, C]$, and its nominal mean $\mu_N \in [-C, C]$. Consequently, the squared error $(e_\tau - \mu_N)^2 \in [0, 4C^2]$, meaning each summand $\phi_\tau$ is bounded almost surely within an interval of width $4C^2$:
\begin{equation}
    \phi_\tau \in [-\sigma_N^2, 4C^2 - \sigma_N^2] 
\end{equation}
Under Assumption~1, the model errors $\{e_t\}_{t \ge 1}$ are independent and identically distributed, implying that $\{\phi_\tau\}_{\tau=1}^t$ is a sequence of i.i.d. bounded random variables. 

\paragraph{Case 1: FP probability under $H_0$ ($y_t \sim \mathcal{D}_N$)}
Under the null hypothesis, nominal behavior guarantees that $\mathbb{E}_{\mathcal{D}_N}[\phi_\tau] = 0$, and therefore $\mathbb{E}_{\mathcal{D}_N}[f_t] = 0$. Evaluating the probability of exceeding the midpoint decision boundary $\Delta_A / 2$:
\begin{equation}
    \mathbb{P}\left(f_t > \frac{\Delta_A}{2} \;\middle|\; y_t \sim \mathcal{D}_N\right) = \mathbb{P}\left(f_t - \mathbb{E}_{\mathcal{D}_N}[f_t] > \frac{\Delta_A}{2} \;\middle|\; y_t \sim \mathcal{D}_N\right).
\end{equation}
Applying the one-sided upper-tail Hoeffding inequality with $\epsilon = \Delta_A / 2$:
\begin{align}
    \mathbb{P}\left(f_t > \frac{\Delta_A}{2} \;\middle|\; y_t \sim \mathcal{D}_N\right) &\le \exp\left(-\frac{2 t^2 (\Delta_A / 2)^2}{\sum_{\tau=1}^t (4C^2)^2}\right) \nonumber \\
    &= \exp\left(-\frac{\Delta_A^2}{32C^4} t\right).
\end{align}

\paragraph{Case 2: FN Probability under $H_1$ ($y_t \sim \mathcal{D}_A$)}
Under the alternative hypothesis, the expected score satisfies $\mathbb{E}_{\mathcal{D}_A}[\phi_\tau] = \Delta_A > 0$, giving $\mathbb{E}_{\mathcal{D}_A}[f_t] = \Delta_A$. The condition $f_t \le \Delta_A / 2$ is centered relative to its expectation under $H_1$ as follows:
\begin{align}
    f_t \le \frac{\Delta_A}{2} \iff f_t - \mathbb{E}_{\mathcal{D}_A}[f_t] \le -\frac{\Delta_A}{2}.
\end{align}
Applying the one-sided lower-tail Hoeffding inequality with deviation $\epsilon = \Delta_A / 2$:
\begin{align}
    \mathbb{P}\left(f_t \le \frac{\Delta_A}{2} \;\middle|\; y_t \sim \mathcal{D}_A \right) &= \mathbb{P}\left(f_t - \mathbb{E}_{\mathcal{D}_A}[f_t] \le -\frac{\Delta_A}{2} \;\middle|\; y_t \sim \mathcal{D}_A\right) \nonumber \\
    &\le \exp\left(-\frac{2 t^2 (\Delta_A / 2)^2}{\sum_{\tau=1}^t (4C^2)^2}\right) \nonumber \\
    &= \exp\left(-\frac{\Delta_A^2}{32C^4} t\right).
\end{align}
Combining both cases completes the proof of Theorem~1.


\subsection{Proof of Theorem 2}
\label{app:proof_thm_bernstein_FP}

Under the null hypothesis $H_0$, the nominal system dynamics ensure that the centered squared error satisfies $\mathbb{E}_{\mathcal{D}_N}[\phi_\tau] = 0$. By the linearity of expectation, the mean of the test statistic is:
\begin{equation}
    \mathbb{E}_{\mathcal{D}_N}[f_t] = \mathbb{E}_{\mathcal{D}_N}\left[\frac{1}{t}\sum_{\tau=1}^t \phi_\tau\right] = 0.
\end{equation}
By definition, because the clipped error $e_\tau \in [-C, C]$ and its mean $\mu_N \in [-C, C]$, the squared deviation $(e_\tau - \mu_N)^2$ lies within $[0, 4C^2]$. Subtracting the variance shifts $\phi_\tau$ to $[-\sigma_N^2, 4C^2 - \sigma_N^2]$. Thus, the total range of variation of each summand is:
\begin{equation}
    R = 4C^2.
\end{equation}
Under Assumption~1, $\{\phi_\tau\}_{\tau=1}^t$ constitutes an i.i.d. sequence of bounded random variables with empirical mean $f_t$ and empirical variance $\hat{\sigma}_t^2 = \frac{1}{t}\sum_{\tau=1}^t (\phi_\tau - f_t)^2$. 

We evaluate the probability of committing an FP at time $t$, corresponding to the event $f_t \ge \gamma_t$ under $H_0$:
\begin{equation}
    \mathbb{P}[f_t \ge \gamma_t \mid H_0] = \mathbb{P}\left(f_t - \mathbb{E}_{\mathcal{D}_N}[f_t] \ge \gamma_t \;\middle|\; H_0\right).
\end{equation}
Substituting the dynamic upper threshold $\gamma_t = \sqrt{\frac{2 \hat{\sigma}_t^2 \ln(3/\alpha)}{t}} + \frac{12 C^2 \ln(3/\alpha)}{t}$, and using the effective range $R = 4C^2$, we obtain:
\begin{equation}
    \gamma_t = \hat{\sigma}_t \sqrt{\frac{2 \ln(3/\alpha)}{t}} + \frac{3 (4C^2) \ln(3/\alpha)}{t} = \hat{\sigma}_t \sqrt{\frac{2 \ln(3/\alpha)}{t}} + \frac{3 R \ln(3/\alpha)}{t}.
\end{equation}
Direct application of the upper-tail deviation bound \eqref{eq:eb_upper_tail} from Lemma~\ref{lem:empirical_bernstein} with error parameter $\delta = \alpha$ yields:
\begin{equation}
    \mathbb{P}[f_t \ge \gamma_t \mid H_0] = \mathbb{P}\left(f_t - \mathbb{E}_{\mathcal{D}_N}[f_t] \ge \hat{\sigma}_t \sqrt{\frac{2 \ln(3/\alpha)}{t}} + \frac{3 R \ln(3/\alpha)}{t} \;\middle|\; H_0\right) \le \alpha,
\end{equation}
which establishes the FP guarantee.

\subsection{Proof of Theorem 3}
\label{app:proof_thm_bernstein_FN}
Under the alternative hypothesis $H_1$, the true expected shift of the score is $\mathbb{E}_{\mathcal{D}_A}[\phi_\tau] = \Delta_A$. Consequently, the true expectation of the test statistic is:
\begin{equation}
    \mathbb{E}_{\mathcal{D}_A}[f_t] = \frac{1}{t}\sum_{\tau=1}^t \mathbb{E}_{\mathcal{D}_A}[\phi_\tau] = \Delta_A.
\end{equation}
Consider an alternative regime $\mathcal{D}_A$ whose true inflation satisfies the detectability condition $\Delta_A \ge \Delta_{\text{target}}$. An FN occurs if the test terminates and erroneously accepts $H_0$, which happens when $f_t \le \xi_t$. 

Recalling the definition of the safety boundary $\xi_t$:
\begin{equation}
    \xi_t = \Delta_{\text{target}} - \left( \hat{\sigma}_t \sqrt{\frac{2 \ln(3/\beta)}{t}} + \frac{12 C^2 \ln(3/\beta)}{t} \right).
\end{equation}
Using the condition $\Delta_A \ge \Delta_{\text{target}}$, we can upper-bound $\xi_t$ in terms of the true expected shift $\Delta_A$:
\begin{equation}
    \xi_t \le \Delta_A - \left( \hat{\sigma}_t \sqrt{\frac{2 \ln(3/\beta)}{t}} + \frac{12 C^2 \ln(3/\beta)}{t} \right).
\end{equation}
Therefore, the event $f_t \le \xi_t$ implies:
\begin{align}
    f_t &\le \Delta_A - \left( \hat{\sigma}_t \sqrt{\frac{2 \ln(3/\beta)}{t}} + \frac{12 C^2 \ln(3/\beta)}{t} \right) \nonumber \\
    \iff \Delta_A - f_t &\ge \hat{\sigma}_t \sqrt{\frac{2 \ln(3/\beta)}{t}} + \frac{12 C^2 \ln(3/\beta)}{t} \nonumber \\
    \iff \mathbb{E}_{\mathcal{D}_A}[f_t] - f_t &\ge \hat{\sigma}_t \sqrt{\frac{2 \ln(3/\beta)}{t}} + \frac{3 R \ln(3/\beta)}{t},
\end{align}
where we have substituted $R = 4C^2$.

Evaluating the probability under $H_1$ and applying the lower-tail deviation bound \eqref{eq:eb_lower_tail} from Lemma~\ref{lem:empirical_bernstein} with $\delta = \beta$:
\begin{align}
    \mathbb{P}[f_t \le \xi_t \mid H_1] &\le \mathbb{P}\left(\mathbb{E}_{\mathcal{D}_A}[f_t] - f_t \ge \hat{\sigma}_t \sqrt{\frac{2 \ln(3/\beta)}{t}} + \frac{3 R \ln(3/\beta)}{t} \;\middle|\; H_1\right) \nonumber \\
    &\le \beta.
\end{align}
This confirms that for any alternative data generating model satisfying $\Delta_A \ge \Delta_{\text{target}}$, the probability of premature safe termination is bounded above by $\beta$, completing the proof.


\subsection{Proof of Theorem 4}

Because the increments $X_t = \phi_t - c$ are i.i.d. under the null hypothesis $\mathcal{D}_N$, the CUSUM statistic $W_t = \max(0, W_{t-1} + X_t)$ acts as a renewal process. It initiates a sequence of statistically identical excursions, resetting at stopping times $T_m = \inf\{t > T_{m-1} : W_t = 0\}$, with $T_0 = 0$.

During the $m$-th excursion (for $t \in (T_{m-1}, T_m]$), the reflecting boundary is inactive (i.e., $W_t>0$), and the statistic is equivalent to a local unconstrained random walk: \begin{equation}W_{T_{m-1}+t} = \sum_{i=1}^t X_{T_{m-1}+i}\triangleq  S_t^{(m)}.\end{equation}
Let $p$ be the probability that an excursion crosses the threshold $h$ in an interval $(T_{m-1},T_m] $. We can rigorously upper-bound $p$ by extending the maximum over the finite excursion duration to the supremum over an infinite horizon $l \ge 1$:
\begin{align}
    p &= \mathbb{P}_{\mathcal{D}_N}\left( \max_{1 \le t \le T_m - T_{m-1}} S_t^{(m)} \ge h \right) \nonumber \\
    &\le \mathbb{P}_{\mathcal{D}_N}\left( \sup_{t \ge 1} S_t^{(m)} \ge h \right).
\end{align}

By the i.i.d. assumption, $S_t^{(m)}$ is equal in distribution to a generic unconstrained random walk $S_l = \sum_{i=1}^l X_i$ (since it is never reset to $0$) . Thus, the probability of a false alarm during any single excursion is bounded by the supremum of this generic walk:
\begin{equation}
    p \le \mathbb{P}_{\mathcal{D}_N}\left( \sup_{l \ge 1} S_l \ge h \right).
\end{equation}

By our definition, the test statistic is bounded in $[-\sigma_N^2, 4C^2 - \sigma_N^2]$ with a total range width of $4C^2$. Accounting for the drift slack parameter $c$, the increments are bounded within $X_i \in [-\sigma_N^2 - c, 4C^2 - \sigma_N^2 - c]$, yielding a total interval width of $B = 4C^2$. By the configured drift, $\mathbb{E}_{\mathcal{D}_N}[X_i] = -c$. Applying Hoeffding's Lemma (Lemma~\ref{lem:hoeffding_lemma}) to this bounded increment yields the bound on its moment-generating function for any $\lambda > 0$:
\begin{equation}
    \mathbb{E}_{\mathcal{D}_N}[e^{\lambda X_i}] \le \exp\left( -\lambda c + \frac{\lambda^2 B^2}{8} \right).
\end{equation}
We select $\lambda = \frac{8c}{B^2}$ to exactly zero the exponent, which guarantees $\mathbb{E}_{\mathcal{D}_N}[e^{\lambda X_i}] \le 1$. 

Using this $\lambda$, we construct the exponential process $M_l = \exp(\lambda S_l) = \exp\left( \lambda \sum_{i=1}^l X_i \right)$. Because the increments are independent and $\mathbb{E}_{\mathcal{D}_N}[e^{\lambda X_i}] \le 1$, the sequence $\{M_l\}_{l \ge 1}$ forms a non-negative supermartingale with initial value $M_0 = \exp(0) = 1$. 

Due to the monotonicity of the exponential function,
the event that the unconstrained sum ever crosses $h$ is equivalent to the supermartingale crossing $\exp(\lambda h)$. Applying Ville's Inequality (Lemma~\ref{lem:ville}) provides a uniform, time-independent upper bound on the false alarm probability for any single excursion:
\begin{align}\label{eq:theroem4_Excursion_Prob}
    \mathbb{P}_{\mathcal{D}_N}\left( \sup_{l \ge 1} S_l \ge h \right) &= \mathbb{P}_{\mathcal{D}_N}\left( \sup_{l \ge 1} M_l \ge \exp(\lambda h) \right) \nonumber \\
    &\le \frac{\mathbb{E}_{\mathcal{D}_N}[M_0]}{\exp(\lambda h)} \nonumber \\
    &= \exp\left( -\frac{8ch}{B^2} \right).
\end{align}

Because the CUSUM statistic resets to zero after each failed excursion, the sequence of excursions forms a series of independent trials. The total number of excursions until the first alarm follows a geometric distribution with success probability $p$, yielding an expected count of $1/p$. Since every excursion takes at least one time step to complete, the overall ARL ($\text{ARL}_0 = \mathbb{E}_{\mathcal{D}_N}[\hat{\tau}]$) is strictly bounded below by this expected number of excursions:
\begin{equation}\label{eq:Theorem4_final_part}
    \text{ARL}_0 \ge  \frac{1}{p} \ge \exp\left( \frac{8ch}{B^2} \right).
\end{equation}

Setting this theoretical lower bound equal to the desired target $\text{ARL}_{\text{target}}$ and isolating $h$ yields the required threshold configuration:
\begin{equation}
    h = \frac{B^2 \ln(\text{ARL}_{\text{target}})}{8c} = \frac{16C^4 \ln(\text{ARL}_{\text{target}})}{8c} = \frac{2C^4 \ln(\text{ARL}_{\text{target}})}{c}.
\end{equation}
This guarantees the expected run length under the pre-change regime meets or exceeds $\text{ARL}_{\text{target}}$, completing the proof.
\subsection{Proof of Theorem 5}
Under the post-change regime $\mathcal{D}_A$ ($t > \tau^*$), the expected value of the score shifts due to the structural divergence. Let the increment be $X_t = \phi_t - c$. We assume the drift parameter $c$ is configured such that the expected increment under the alternative distribution possesses a strictly positive drift:
\begin{equation}
    \nu_A = \mathbb{E}_{\mathcal{D}_A}[X_t] > 0.
\end{equation}

To rigorously bound the EDD $\mathbb{E}_{\mathcal{D}_A}[\hat{\tau} - \tau^* \mid \hat{\tau} > \tau^*]$, we consider the worst-case scenario where the CUSUM statistic is completely depleted exactly at the changepoint, meaning $W_{\tau^*} = 0$. If $W_{\tau^*} > 0$ then the statistic will cross the threshold faster, meaning that this constitutes a worst-case bound on the EDD.

For post-change steps $t \ge 1$, we define the unconstrained random walk of the post-change increments as $S_t = \sum_{i=1}^t X_{\tau^*+i}$. Because the CUSUM statistic is bounded below by zero, it strictly dominates the unconstrained walk: $W_{\tau^*+t} \ge S_t$. Let $\hat{t} = \inf\{t \ge 1 : S_t \ge h\}$ be the stopping time of this unconstrained walk. By domination, the actual detection delay is bounded by this unconstrained stopping time: $\hat{\tau} - \tau^* \le \hat{t}$.

Because $\nu_A > 0$, the unconstrained walk crosses $h$ in finite time almost surely. Applying Wald's identity to $S_t$ yields:
\begin{equation}\label{eq:WaldsApplicationForCPD}
    \mathbb{E}_{\mathcal{D}_A}[S_{\hat{t}}] = \nu_A \cdot \mathbb{E}_{\mathcal{D}_A}[\hat{t}].
\end{equation}

At the moment of crossing, the unconstrained sum satisfies $S_{\hat{t}} \ge h$. The extent to which the sum overshoots $h$ is strictly bounded by the maximum possible positive magnitude of a single increment, which is $4C^2 - c$. This provides the bounds $h \le S_{\hat{t}} < h + 4C^2 - c$. 

Taking expectations and substituting into Wald's identity yields:
\begin{equation}
    h \le \nu_A \cdot \mathbb{E}_{\mathcal{D}_A}[\hat{t}] < h + 4C^2 - c.
\end{equation}

Since the actual EDD is bounded above by the expected stopping time of the unconstrained walk ($\text{EDD} \le \mathbb{E}_{\mathcal{D}_A}[\hat{t}]$), dividing through by the positive drift $\nu_A$ establishes the bound:
\begin{equation}
    \text{EDD} \le \frac{h}{\nu_A} + \frac{4C^2 - c}{\nu_A}.
\end{equation}
Because the overshoot fraction is a finite constant independent of the threshold $h$, this yields the asymptotic upper bound on the delay:
\begin{equation}
    \text{EDD} \le \frac{h}{\nu_A} + \mathcal{O}(1),
\end{equation}
completing the proof.

\section{Theoretical Guarantees Beyond the I.I.D Assumption.}
\label{app:empirical_bernstein_extensions}

This appendix rigorously extends the theoretical guarantees of  CUREB-SHT and CUR-CUSUM relaxing Assumption 1. The clipped centered statistics are assumed to be a Martingale Difference Sequence (MDS). Before going into details about  why this assumption is reasonable, and explaining how the detection algorithms maintain their guarantees we present a novel general concentration result.
\subsection{Preliminaries and Foundational Theorems}

Let $(\Omega, \mathcal{F}, \mathbb{P})$ be a probability space equipped with a filtration $\{\mathcal{F}_t\}_{t \ge 0}$ representing the historical information up to time $t$. Let $\{X_t\}_{t \ge 1}$ be a sequence of random variables adapted to $\mathcal{F}_t$. We track the conditional mean $\mu_t = \mathbb{E}[X_t \mid \mathcal{F}_{t-1}]$ and define the centered increment as $Y_t = X_t - \mu_t$. 

By construction, $\{Y_t\}_{t \ge 1}$ is an MDS satisfying $\mathbb{E}[Y_t \mid \mathcal{F}_{t-1}] = 0$. We assume the increments are strictly bounded such that $|Y_t| \le R$ almost surely for some constant $R > 0$. (For our proposed statistic, this holds due to clipping in $[-C,C]$.)

Let the martingale sum be $M_t = \sum_{\tau=1}^t Y_\tau$. We define two measures of variance:
\begin{enumerate}
    \item \textbf{Predictable Variation (True Variance):} $V_t = \sum_{\tau=1}^t \mathbb{E}[Y_\tau^2 \mid \mathcal{F}_{\tau-1}]$
    \item \textbf{Empirical Variation (Sample Variance):} $\hat{V}_t = \sum_{\tau=1}^t Y_\tau^2$
\end{enumerate}

\subsection{Martingale Empirical Bernstein Bound}

\begin{theorem}[Martingale Empirical Bernstein Bound.] Let $\{Y_\tau\}_{\tau=1}^t$ be an MDS bounded by $R$ almost surely, such that $\mathbb{E}[Y_\tau | \mathcal{F}_{\tau-1}] = 0$. Let $\bar{Y}_t = \frac{1}{t} \sum_{\tau=1}^t Y_\tau$ denote the empirical running average, and let $\hat{\sigma}_t = \sqrt{\frac{1}{t}\sum_{\tau=1}^t Y_\tau^2}$ denote the empirical standard deviation. For any $\delta \in (0, 1)$, with probability at least $1 - \delta$, the following inequality holds:
$$ |\bar{Y}_t| \le \hat{\sigma}_t \sqrt{\frac{2 \ln(c_t/\delta)}{t}} + \frac{c'_t R \ln(c_t/\delta)}{t} $$
where $c_t$ and $c'_t$ are dynamic logarithmic penalty constants subsuming the union bounds over the variance space and sub-linear estimation errors.
\end{theorem}
\textit{Equivalently, the one-sided tail deviations satisfy:}
\begin{align}
\mathbb{P}\left( \bar{Y}_t \ge \hat{\sigma}_t \sqrt{\frac{2 \ln(c_t/\delta)}{t}} + \frac{c'_t R \ln(c_t/\delta)}{t} \right) &\le \delta, \\
\mathbb{P}\left( -\bar{Y}_t \ge \hat{\sigma}_t \sqrt{\frac{2 \ln(c_t/\delta)}{t}} + \frac{c'_t R \ln(c_t/\delta)}{t} \right) &\le \delta.
\end{align}
\subsection{Proof of the Theorem}

The proof proceeds in four steps: establishing a deterministic variance bound, applying a geometric peeling argument to handle random variance, utilizing measure concentration to swap the theoretical variance for the empirical variance, and translating the sum bound to an average.

\subsubsection*{Step 1: The Predictable Tail Bound}
Freedman's Inequality (Lemma \ref{lem:freedman}) bounds the tail of the martingale sum $M_t$. However, this inequality only holds when the variance limit $v$ is a fixed, deterministic scalar. In practice, the predictable variation $V_t$ is a random variable dependent on the data path, meaning Freedman cannot be evaluated directly on $V_t$.

\subsubsection*{Step 2: The Peeling Argument (Variance Grid)}
To apply Freedman's inequality to a random variable, we partition the space of all possible values of $V_t$ into a geometric grid. Because each increment is bounded by $R$, the predictable variation $V_t$ must lie in the interval $[0, tR^2]$.

We allocate half of our failure probability budget ($\delta/2$) to this step. We define a tiny base variance $v_0 = R^2$ and a grid growth parameter $\eta > 0$. The grid boundaries are defined as $v_k = v_0(1+\eta)^k$ for $k=1, \dots, K$, where $K = \lceil \ln_{1+\eta}(tR^2/v_0) \rceil$. The space $[0, tR^2]$ is thus divided into a base slice $[0, v_0]$ and $K$ geometric slices $(v_{k-1}, v_k]$.

We define an adaptive threshold function $x(v)$ that increases monotonically with variance. If the true random variance $V_t$ falls within a geometric slice $(v_{k-1}, v_k]$, we are guaranteed that $V_t > v_{k-1}$. Consequently, the  threshold satisfies $x(V_t) > x(v_{k-1})$. Because crossing a higher threshold is a strictly rarer event, the occurrence of $\{M_t \ge x(V_t)\}$ logically guarantees the occurrence of $\{M_t \ge x(v_{k-1})\}$.  Formally: \begin{equation}
\mathbb{P}(M_t \ge x(V_t)) \le \mathbb{P}(M_t \ge x(v_{k-1}))
\end{equation}
By the Law of Total Probability and the union bound over the base slice and all $K$ geometric slices, we decompose the total failure probability $\mathbb{P}\left( M_t \ge x(V_t) \right)$. Applying the threshold substitution above to safely overestimate the probability within each slice yields the following inequality:
\begin{equation} \mathbb{P}\left( M_t \ge x(V_t) \right) \le \mathbb{P}\left( M_t \ge x(v_0), V_t \le v_0 \right) + \sum_{k=1}^K \mathbb{P}\left( M_t \ge x(v_{k-1}), V_t \in (v_{k-1}, v_k] \right) \end{equation}
We allocate an equal probability budget $\delta' = \frac{\delta}{2(K+1)}$ to each slice. For the $k$-th geometric slice, the true variance is bounded by $v_k$. Because $\{V_t \in (v_{k-1}, v_k]\}$ strictly implies $\{V_t \le v_k\}$, we can upper-bound the slice's failure probability and then directly apply Freedman's inequality using the deterministic scalar $v_k$:
\begin{equation}\mathbb{P}\left( M_t \ge x(v_{k-1}), V_t \in (v_{k-1}, v_k] \right) \le \mathbb{P}\left( M_t \ge x(v_{k-1}), V_t \le v_k \right) \le \exp\left( - \frac{x(v_{k-1})^2}{2(v_k + R x(v_{k-1})/3)} \right) \end{equation}

To guarantee the failure probability for this slice does not exceed its allocated budget, we set this exponential upper bound exactly equal to $\delta'$:
$$ \exp\left( - \frac{x(v_{k-1})^2}{2(v_k + R x(v_{k-1})/3)} \right) = \delta' $$


Taking the natural logarithm of both sides and rearranging yields:
\begin{equation*}
\frac{x(v_{k-1})^2}{2(v_k + R x(v_{k-1})/3)} = \ln(1/\delta')
\end{equation*}

Multiplying out the denominator to form a standard quadratic equation in $x(v_{k-1})$:
\begin{align*}
x(v_{k-1})^2 &= 2v_k \ln(1/\delta') + \frac{2R}{3}\ln(1/\delta') x(v_{k-1}) \\
x(v_{k-1})^2 &- \left(\frac{2R}{3}\ln(1/\delta')\right) x(v_{k-1}) - 2v_k \ln(1/\delta') = 0
\end{align*}

We apply the quadratic formula $\left(x = \frac{-b \pm \sqrt{b^2 - 4ac}}{2a}\right)$ with $a = 1$, $b = -\frac{2R}{3}\ln(1/\delta')$, and $c = -2v_k \ln(1/\delta')$:
\begin{align*}
x(v_{k-1}) &= \frac{ \frac{2R}{3}\ln(1/\delta') \pm \sqrt{\left(\frac{2R}{3}\ln(1/\delta')\right)^2 - 4(1)(-2v_k\ln(1/\delta'))} }{2} \\
x(v_{k-1}) &= \frac{R}{3}\ln(1/\delta') \pm \frac{1}{2} \sqrt{\frac{4R^2}{9}\ln^2(1/\delta') + 8v_k\ln(1/\delta')}
\end{align*}

Pulling the factor of $4$ out of the square root (as a $2$) to cancel the $1/2$ coefficient gives the exact positive root:
\begin{equation*}
x(v_{k-1}) = \frac{R}{3}\ln(1/\delta') + \sqrt{\frac{R^2}{9}\ln^2(1/\delta') + 2v_k\ln(1/\delta')}
\end{equation*}

To obtain a simplified threshold bound, we apply the subadditivity of the square root ($\sqrt{a+b} \le \sqrt{a} + \sqrt{b}$) to the term inside the radical:
\begin{align*}
x(v_{k-1}) &\le \frac{R}{3}\ln(1/\delta') + \sqrt{\frac{R^2}{9}\ln^2(1/\delta')} + \sqrt{2v_k\ln(1/\delta')} \\
x(v_{k-1}) &\le \frac{R}{3}\ln(1/\delta') + \frac{R}{3}\ln(1/\delta') + \sqrt{2v_k\ln(1/\delta')} 
\end{align*}

Consolidating the logarithmic terms yields the corrected upper bound for the threshold:
\begin{equation*}
x(v_{k-1}) \le \frac{2R}{3}\ln(1/\delta') + \sqrt{2v_k\ln(1/\delta')}
\end{equation*}

\color{black}
Solving for the threshold $x(v_{k-1})$ and using the grid definition $v_k = (1+\eta)v_{k-1}$ yields:
\begin{equation}\label{eq:interval_threshold_def_for_martingale_proof}
x(v_{k-1}) \le \sqrt{2(1+\eta)v_{k-1}\ln(1/\delta')} + \frac{2R}{3}\ln(1/\delta')\end{equation}

For the base slice ($k=0$), the true variance is bounded in the interval $V_t \in [0, v_0]$. Because the maximum possible variance in this slice ($v_0$) is a fixed  scalar, we directly apply Freedman's inequality using $v_0$:
\begin{equation} \mathbb{P}\left( M_t \ge x(v_0), V_t \le v_0 \right) \le \exp\left( - \frac{x(v_0)^2}{2(v_0 + R x(v_0)/3)} \right) = \delta'\end{equation}
Solving for the base threshold $x(v_0)$ yields a fixed scalar boundary:
\begin{equation} x(v_0) = \sqrt{2 v_0 \ln(1/\delta')} + \frac{2R}{3} \ln(1/\delta') \end{equation}

Having constrained the failure probability within each individual slice to $\delta'$, we aggregate these bounds across the entire variance space. By applying the union bound over all $K+1$ partitions (the base slice and the $K$ geometric slices), the total probability that $M_t$ exceeds its corresponding slice threshold is:
$$ \sum_{k=0}^K \delta' = (K+1) \left( \frac{\delta}{2(K+1)} \right) = \frac{\delta}{2} $$
Because the total probability of failure anywhere on the grid is strictly bounded by $\delta/2$, we can define a global success event $\mathcal{E}$ by taking the complement.  Formally $\mathbb{P}[\mathcal{E}] \geq 1- \delta/2$ where :
$$ \mathcal{E} = \left\{ M_t < \begin{cases} 
  x(v_0) & \text{if } V_t \in [0, v_0] \\ 
  x(v_{k-1}) & \text{if } V_t \in (v_{k-1}, v_k] \text{ for } k \ge 1 
\end{cases} \right\} $$

To construct a single, anytime-valid equation without piecewise conditions, we define a global envelope function that acts as an upper bound for all possible cases simultaneously.

When this successful event $\mathcal{E}$  with probability at least $1-\delta/2$ occurs, one of two cases must hold:
\begin{enumerate}
    \item \textbf{Case 1 (Base Slice):} If $V_t \in [0, v_0]$, then $M_t < x(v_0)$. Because the variance $V_t$ is inherently non-negative, the quantity $\sqrt{2(1+\eta)V_t \ln(1/\delta')}$ is always greater than or equal to zero. Adding this non-negative quantity to our baseline threshold $x(v_0)$ yields a mathematically valid, albeit looser, upper bound: $M_t < \sqrt{2(1+\eta)V_t \ln(1/\delta')} + x(v_0)$.
    \item \textbf{Case 2 (Other Slices):} If $V_t \in (v_{k-1}, v_k]$ for some $k \ge 1$, we established that $M_t < x(v_{k-1})$. Expanding this threshold gives $M_t < \sqrt{2(1+\eta)v_{k-1} \ln(1/\delta')} + \frac{R}{3}\ln(1/\delta')$. Within this specific slice, we are mathematically guaranteed that $V_t > v_{k-1}$. Replacing the fixed boundary $v_{k-1}$ with the larger true variance $V_t$ strictly increases the value of the square root term. Consequently, this substitution produces a valid, larger upper bound: $M_t < \sqrt{2(1+\eta)V_t \ln(1/\delta')} + \frac{R}{3}\ln(1/\delta')$.
\end{enumerate}
Combining both cases yields
\begin{equation} M_t \le \sqrt{2(1+\eta)V_t \ln(1/\delta')} + x(v_0). \end{equation}

By setting the base variance as $v_0=R^2$ and setting $L_t = \ln\frac{2(K+1)}{\delta} $ for notational convenience, we get the final adaptive threshold: 
\begin{equation}\label{eq:Martingale_step2} M_t \le \sqrt{2 (1+\eta) V_t L_t} + R\sqrt{2L_t}+\frac{2R}{3}L_t,\end{equation}
with probability at least $1-\delta/2$.

\subsubsection*{Step 3: Empirical Variance Substitution}
The threshold derived in Step 2 depends on the unobservable true variance $V_t$. We must replace it with the observable empirical variance $\hat{V}_t = \sum_{\tau=1}^t Y_\tau^2$.

We define a secondary sequence representing the difference between the true and empirical variance: $Z_\tau = \mathbb{E}[Y_\tau^2 \mid \mathcal{F}_{\tau-1}] - Y_\tau^2$. Because $Y_\tau \in [-R, R]$, its square $Y_\tau^2 \in [0, R^2]$. Therefore, $Z_\tau$ is an MDS strictly bounded by $R^2$.

We apply the Azuma-Hoeffding inequality (Lemma \ref{lem:azuma}) to the sum of $Z_\tau$, allocating the remaining $\delta/2$ probability budget:
\begin{equation} \mathbb{P}\left( \sum_{\tau=1}^t Z_\tau \ge \epsilon \right) \le \exp\left( - \frac{\epsilon^2}{2t R^4} \right) = \frac{\delta}{2} \end{equation}

Solving for $\epsilon$ reveals that with probability at least $1 - \delta/2$:
\begin{equation} V_t - \hat{V}_t \le R^2 \sqrt{2t \ln(2/\delta)} \implies V_t \le \hat{V}_t + R^2 \sqrt{2t \ln(2/\delta)} \end{equation}

We substitute this upper bound for $V_t$ into the threshold from Step 2 in eq. (\ref{eq:Martingale_step2}). This yields:
\begin{equation} M_t \le \sqrt{2 (1+\eta) \left( \hat{V}_t + R^2 \sqrt{2t \ln(2/\delta)} \right) L_t} +  R\sqrt{2L_t}+\frac{\textcolor{black}{2}R}{3}L_t\end{equation}

Applying the algebraic subadditivity of the square root, $\sqrt{a+b} \le \sqrt{a} + \sqrt{b}$, we decouple the empirical variance from its concentration error:
\begin{align}\label{eq:Marticale_Decompose_Mt}
M_t &\le \sqrt{2 (1+\eta) \hat{V}_t L_t}  + \sqrt{2 (1+\eta) R^2 \sqrt{2t \ln(2/\delta)} L_t} +  R\sqrt{2L_t}+\frac{{2}R}{3}L_t
\end{align}

The first term on the right-hand side of \eqref{eq:Marticale_Decompose_Mt} represents the dominant empirical variance bound; because $\hat{V}_t \le tR^2$, this term scales as $\mathcal{O}(t^{1/2})$. The second term represents the estimation error between the true and empirical variance. Because it contains $\sqrt{\sqrt{t}}$, this penalty scales as $\mathcal{O}(R t^{1/4} L_t^{3/4})$. While $t^{1/4}$ grows asymptotically slower than the dominant empirical term, it strictly increases with time and remains as an explicit additive sub-linear penalty. The remaining  is the fixed, exact algebraic residual generated by the base slice threshold $x(v_0)$.

To  reach the final  bound, we consolidate the  factor $(1+\eta)$ into the  logarithm. We define a new parameter $c_t$ exactly such that $(1+\eta) L_t = \ln(c_t/\delta)$. Solving this equality for $c_t$ yields the explicit assignment:
\begin{equation} c_t = \delta \left( \frac{2(K+1)}{\delta} \right)^{1+\eta} \end{equation}
Substituting this definition into the first term simplifies our dominant empirical variance bound to exactly $\sqrt{2 \hat{V}_t \ln(c_t/\delta)}$.

Next, we address the remaining additive terms. Because the estimation penalty contains $t^{1/4}$, it is mathematically invalid to absorb it into a static, time-independent constant. Instead, we rigorously group the time-dependent estimation penalty and the fixed base-slice residuals into a single composite penalty term. 

To express this penalty in the canonical form $c'_t R \ln(c_t/\delta)$, we define the explicit, time-dependent coefficient $c'_t$ as follows:
$$ c'_t = \frac{1}{\ln(c_t/\delta)} \left( \sqrt{2 (1+\eta) \sqrt{2t \ln(2/\delta)} L_t} + \sqrt{2 L_t} + \frac{\textcolor{black}{2}}{3}L_t \right) $$

By utilizing these exact definitions for $c_t$ and $c'_t$,  the fully empirical, adaptive bound on the martingale sum is rigorously given by:
\begin{equation}\label{eq:MartingaleHT_Step3_End}
 M_t \le \sqrt{2 \hat{V}_t \ln(c_t/\delta)} + c'_t R \ln(c_t/\delta), \end{equation} with probability $1-\delta$.
\subsubsection*{Step 4: Translation to Final Statistic}
To recover the running average, we divide the entire one-sided inequality by $t$:
\begin{equation}
     \bar{Y}_t = \frac{M_t}{t} \le \frac{\sqrt{2 \hat{V}_t \ln(c_t/\delta)}}{t} + \frac{c'_t R \ln(c_t/\delta)}{t}
\end{equation} 
Pushing $t$ inside the square root as $t^2$ allows us to recover the empirical standard deviation $\hat{\sigma}_t$:
\begin{equation} \frac{\sqrt{2 \hat{V}_t \ln(c_t/\delta)}}{\sqrt{t^2}} = \sqrt{\frac{\hat{V}_t}{t}} \sqrt{\frac{2 \ln(c_t/\delta)}{t}} = \hat{\sigma}_t \sqrt{\frac{2 \ln(c_t/\delta)}{t}} \end{equation}

Because the mirrored sequence $\{-Y_\tau\}_{\tau=1}^t$ is also an MDS bounded by $R$, applying the identical peeling and concentration arguments to the inverted sum $-M_t = \sum_{\tau=1}^t (-Y_\tau)$ establishes the symmetric lower bound. Combining these complementary bounds yields the final two-sided Martingale Empirical Bernstein inequality:
\begin{equation*} |\bar{Y}_t| \le \hat{\sigma}_t \sqrt{\frac{2 \ln(c_t/\delta)}{t}} + \frac{c'_t R \ln(c_t/\delta)}{t}, 
\end{equation*}
with probability $1-\delta$.
This concludes the proof.
\subsection{Consequence on SHT}
Even if we relax the strict i.i.d. requirement of Assumption 1, the theoretical guarantees of our framework are preserved with minor modifications.\\ 
\textbf{Assumption 2. (MDS)} \textit{Let $\{\mathcal{F}_t\}_{t \ge 0}$ denote the  filtration generated by the historical observations up to time $t$. Under the   regime $\mathcal{D}_N$, the sequence of test statistics $\{\phi_t\}_{t \ge 1}$ is adapted to this filtration, and its conditional expectation given the past evaluates to zero:}
\begin{equation}
\mathbb{E}_{\mathcal{D}_N}[\phi_t \mid \mathcal{F}_{t-1}] = 0.
\end{equation}

In practical streaming environments, expressive base models such as LSTMs or near-optimal Kalman-like estimators  capture the predictable temporal dynamics of the stream. Because our framework explicitly centers the squared errors using the nominal baseline $\mu_N$ and $\sigma_N^2$ (which are safely estimated from a large dataset), the conditional expected value of  $\phi_\tau$ given the historical data is engineered to be exactly zero under the null hypothesis. Consequently, the sequence of test statistics can be assumed to be an MDS, which is much more general than assuming i.i.d. errors.

By repeating the proofs for Theorems 2, and 3 using Theorem 6 in place of Lemma \ref{lem:empirical_bernstein} we can derive the following error-guaranteeing bounds.

\textbf{Corollary 1 (Robust Thresholds for MDS Data).} \textit{Let the sequence of scores $\{\phi_\tau\}_{\tau=1}^t$ be an MDS with respect to the filtration $\mathcal{F}_{\tau-1}$, bounded within an interval of total width $4C^2$. Under the null hypothesis $H_0$, let $\mathbb{E}_{\mathcal{D}_N}[\phi_\tau \mid \mathcal{F}_{\tau-1}] = 0$. For any target FP level $\alpha \in (0, 1)$, setting the dynamic upper threshold to:}
\begin{equation*} \gamma'_t = \sqrt{\frac{2\hat{\sigma}_t^2 \ln(c_{t,\alpha}/\alpha)}{t}} + \frac{2C^2 \, c'_{t,\alpha} \ln(c_{t,\alpha}/\alpha)}{t} \end{equation*}
\textit{guarantees that $\mathbb{P}(f_t \ge \gamma'_t \mid y_t \sim \mathcal{D}_N) \le \alpha$ at time $t$.}

\textit{Similarly, let $\Delta_{\text{target}} > 0$ denote a user-specified minimum detectable alternative margin. Under the alternative hypothesis $H_1$, if the true conditional drift satisfies $\mathbb{E}_{\mathcal{D}_A}[\phi_\tau \mid \mathcal{F}_{\tau-1}] \ge \Delta_{\text{target}}$, setting the dynamic lower threshold to:}
\begin{equation*} \xi'_t = \Delta_{\text{target}} - \left( \sqrt{\frac{2\hat{\sigma}_t^2 \ln(c_{t,\beta}/\beta)}{t}} + \frac{2C^2 \, c'_{t,\beta} \ln(c_{t,\beta}/\beta)}{t} \right) \end{equation*}
\textit{guarantees that $\mathbb{P}(f_t \le \xi'_t \mid y_t \sim \mathcal{D}_A)  \le \beta$ at time $t$, where $c_{t,\delta}$ and $c'_{t,\delta}$ are the adaptive penalty constants from Theorem 6 evaluated at $\delta = \alpha$ and $\delta = \beta$ respectively.}

The exact MDS thresholds ($\gamma'_t, \xi'_t$) share a similar scaling trajectory (i.e. $O(1/t)$ with $C^2$ and $O(1/\sqrt t)$ with the variance estimate) with the simpler i.i.d. thresholds ($\gamma_t, \xi_t$). Furthermore, because distribution-free concentration inequalities evaluate worst-case tail bounds, they are inherently conservative in practice. This theoretical conservatism acts as a natural buffer, effectively absorbing the slight mathematical gap between the static i.i.d. constants and the dynamic MDS penalties. This dynamic explains why deploying the i.i.d. CUREB-SHT stopping rules on complex, non-i.i.d. streams remains highly robust. As definitively validated by our empirical evaluations across diverse nonlinear and temporally correlated streams, the observed experimental error rates frequently fall (often significantly) below the user-prescribed $\alpha$ and $\beta$ tolerances, confirming that the lightweight i.i.d. thresholds serve as a highly effective and safe structural proxy.

\subsection{SCPD}
It turns out that the CUR-CUSUM statistic also maintains its theoretical guarantees under the more general MDS assumption. Let $\{\phi_t\}_{t \ge 1}$ be adapted to a filtration $\{\mathcal{F}_t\}_{t \ge 1}$ representing the historical observations up to time $t$. Under the null hypothesis, we again model $\phi_t$ as an MDS, satisfying $\mathbb{E}_{\mathcal{D}_N}[\phi_t \mid \mathcal{F}_{t-1}] = 0$. 

Regarding the proof of Theorem 4, the results up to eq. (\ref{eq:theroem4_Excursion_Prob}) utilize Hoeffding's lemma and Ville's
 inequality which are valid for Martingales and do not require Assumption 1.  The main issue lies in eq. (\ref{eq:Theorem4_final_part}), because the CUSUM excursions cannot be assumed i.i.d. geometric random variables. 

 Let $M$ be the total number of excursions until the first threshold crossing. The event $\{M > m\}$ implies that the first $m$ excursions sequentially failed to cross $h$. Because the conditional probability of success for any excursion is at most $p$ (where $p$ is given by eq. \eqref{eq:theroem4_Excursion_Prob}), the conditional probability of failure is at least $1 - p$. Chaining these conditional probabilities yields a rigorous lower bound on the survival function of $M$:
\begin{equation}
    \mathbb{P}_{\mathcal{D}_N}(M > m) \ge (1 - p)^m.
\end{equation}

By expressing the expected value of $M$ through the standard tail sum formula for expectation, we obtain a geometric series that serves as a strict lower bound:
\begin{equation}
    \mathbb{E}_{\mathcal{D}_N}[M] = \sum_{m=0}^\infty \mathbb{P}_{\mathcal{D}_N}(M > m) \ge \sum_{m=0}^\infty (1 - p)^m = \frac{1}{p}.
\end{equation}

Because every excursion requires at least one time step to complete, the expected time between false alarms is strictly bounded below by the expected number of excursions. Thus, this
proves that Theorem 4 holds rigorously under the generalized MDS assumption without relying on independent geometric trials.

Regarding the proof for Theorem 5, the main issue lies in using Wald's identity (to derive eq. (\ref{eq:WaldsApplicationForCPD})), which necessitates i.i.d. assumptions. We will now derive the equivalent bound under the more general MDS assumption using Doob's Optional Stopping Theorem, formalized in Lemma \ref{lem:doobs}.

Under the alternative regime $\mathcal{D}_A$ ($t > \tau^*$), we assume the conditional expected increment is strictly positive: $\mathbb{E}_{\mathcal{D}_A}[X_t \mid \mathcal{F}_{t-1}] \ge \nu_A > 0$, where $X_t = \phi_t - c$. 

Following the worst-case initialization $W_{\tau^*} = 0$, we define the local unconstrained random walk for post-change steps $t \ge 1$ as $S_t = \sum_{i=1}^t X_{\tau^*+i}$. We construct the compensated sum initialized at the change point as $M_t = S_t - t \nu_A = \sum_{i=1}^t (X_{\tau^*+i} - \nu_A)$, with $M_0 = 0$. The expected single-step evolution is:
\begin{equation}
    \mathbb{E}_{\mathcal{D}_A}[M_t - M_{t-1} \mid \mathcal{F}_{\tau^*+t-1}] = \mathbb{E}_{\mathcal{D}_A}[X_{\tau^*+t} - \nu_A \mid \mathcal{F}_{\tau^*+t-1}] \ge 0.
\end{equation}
Because the conditional expectation of each step is non-negative, the sequence $\{M_t\}_{t \ge 0}$  forms a submartingale.

Let $\hat{t} = \inf\{t \ge 1 : S_t \ge h\}$ be the stopping time of this unconstrained walk. Because the increments are strictly bounded and possess a strictly positive conditional drift ($\nu_A > 0$), the unconstrained walk reaches $h$ with a finite expected stopping time as a direct consequence of the Strong Law of Large Numbers ($\mathbb{E}_{\mathcal{D}_A}[\hat{t}] < \infty$). Because both the expected stopping time is finite and the increment magnitudes are almost surely bounded, Doob's Optional Stopping Theorem is satisfied. Applying Lemma 5 to $\hat{t}$, we get:
\begin{equation}
    \mathbb{E}_{\mathcal{D}_A}[M_{\hat{t}}] \ge \mathbb{E}_{\mathcal{D}_A}[M_0] = 0.
\end{equation}
Expanding the compensated summation isolates the expected unconstrained statistic $S_{\hat{t}}$:
\begin{align}\label{eq:RelaxedWaldResult}
    \mathbb{E}_{\mathcal{D}_A} [ S_{\hat{t}} - \hat{t} \nu_A ] &\ge 0 \nonumber \\
    \mathbb{E}_{\mathcal{D}_A}[S_{\hat{t}}] &\ge \nu_A \cdot \mathbb{E}_{\mathcal{D}_A}[\hat{t}].
\end{align}

Using the bound $$\mathbb{E}_{\mathcal{D}_A}[S_{\hat{t}}] < h + 4C^2 - c$$ derived in the initial proof for theorem 5, and the fact that $\hat{\tau} \leq \hat{t} $  we can now plug in \eqref{eq:RelaxedWaldResult} and conclude the proof without changing the ensuing steps.
\subsection{Experimental Comparison}
\begin{figure}[H]
    \centering
    \begin{subfigure}[b]{0.49\textwidth}
        \centering
        \includegraphics[width=\textwidth]{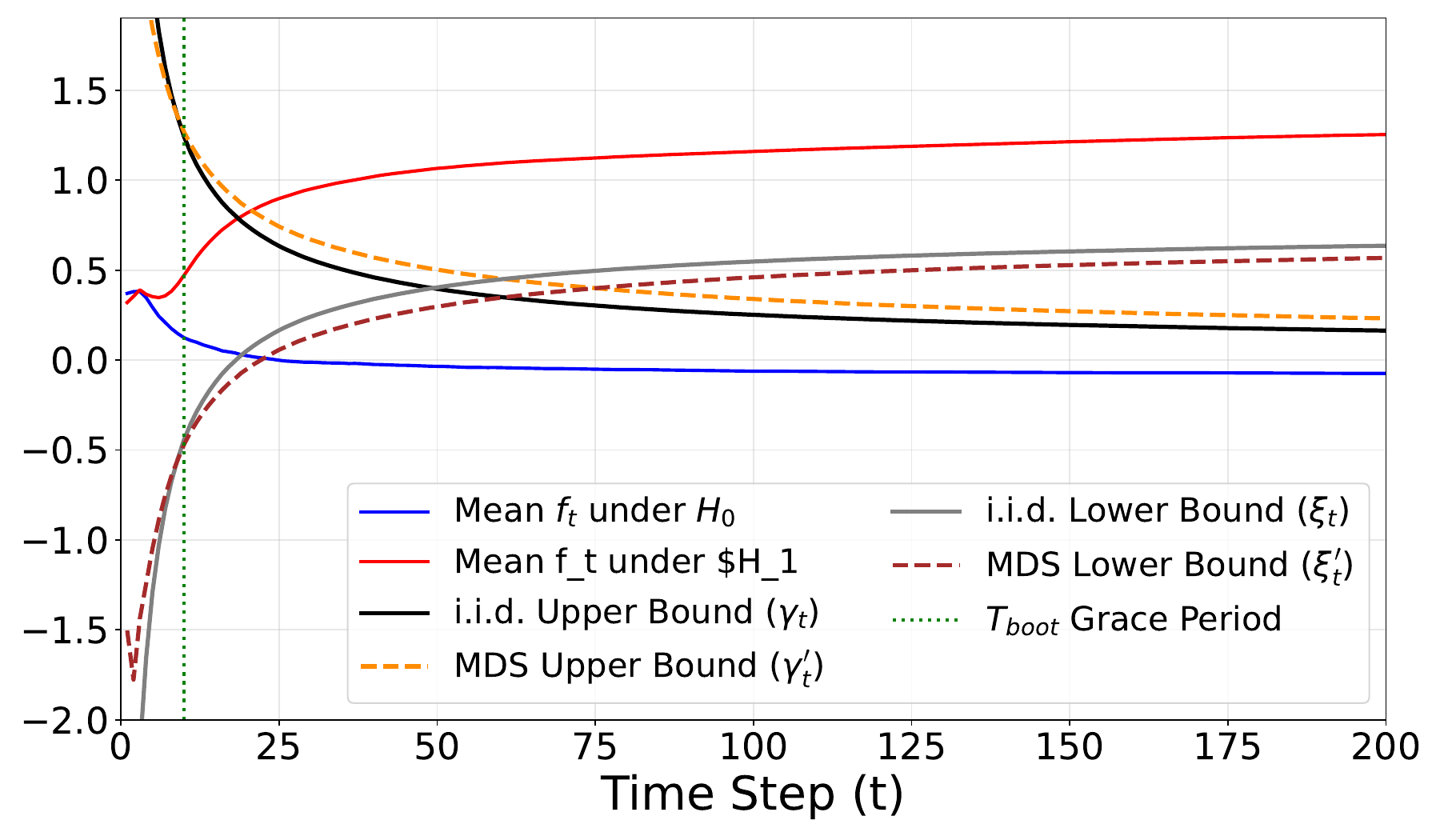}
        \caption{2D Linear Model (Standard KF)}
        \label{fig:mds_comparison_2d}
    \end{subfigure}
    \hfill
    \begin{subfigure}[b]{0.49\textwidth}
        \centering
        \includegraphics[width=\textwidth]{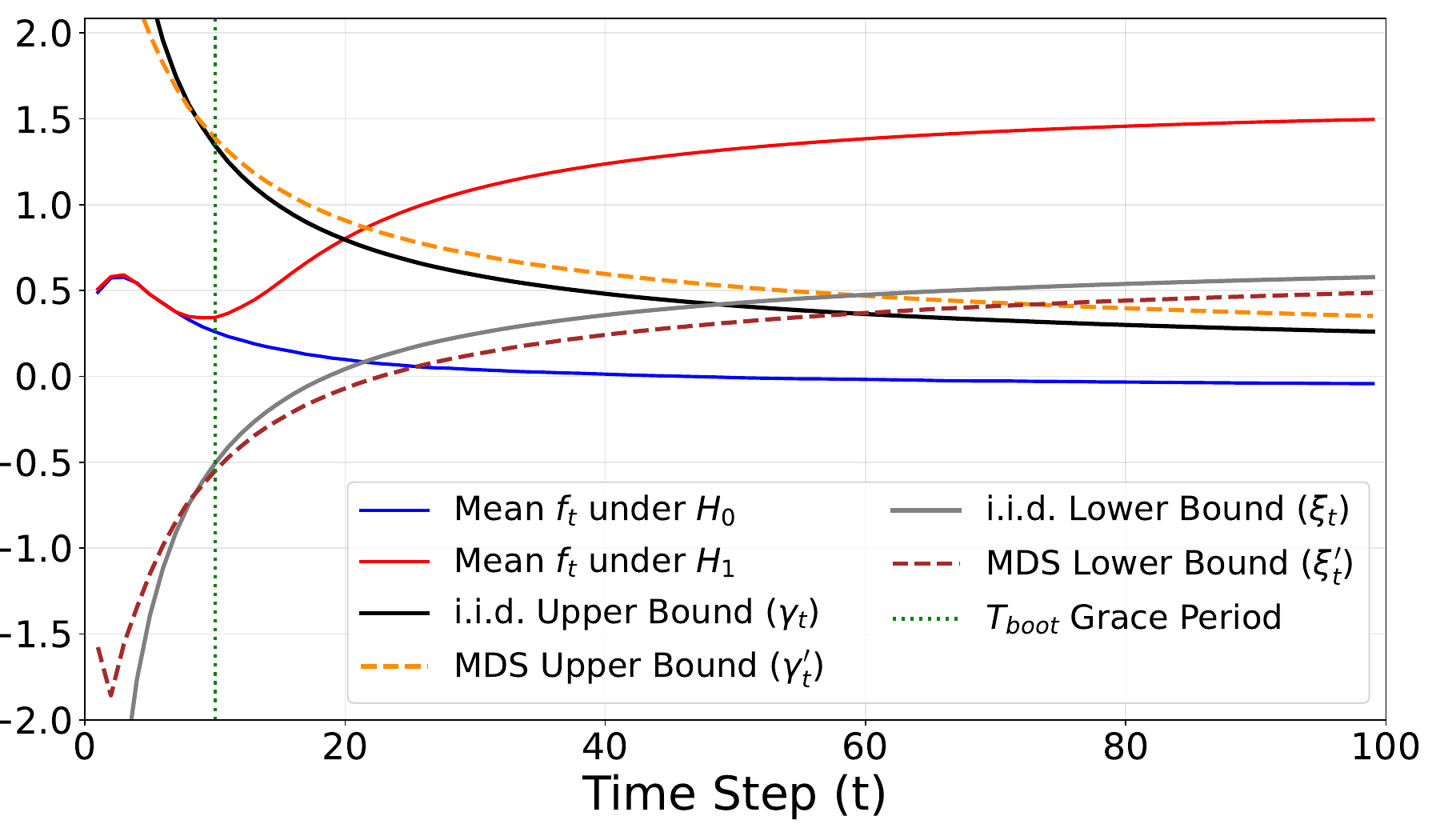}
        \caption{3D Nonlinear  Model (EKF)}
        \label{fig:mds_comparison_3d}
    \end{subfigure}
    \caption{Empirical comparison of the i.i.d. threshold boundaries ($\gamma_t, \xi_t$) versus the bounds ($\gamma'_t, \xi'_t$) derived under the relaxed MDS assumption. }
    \label{fig:mds_ablation_study}
\end{figure}

In Fig. \ref{fig:mds_ablation_study}, we compare the proposed threshold boundaries for the SHT under the i.i.d. premise (Assumption 1) and the relaxed MDS condition (Assumption 2), in the linear and nonlinear target tracking experiments of section \ref{sec:main_Results}. As is evident, the two formulations exhibit highly similar trajectories, resulting in near-identical sequential detection performance. While the dynamic MDS bounds become slightly more conservative (looser) for $t > 20$, this divergence leaves the detection capabilities virtually unaffected. By this stage, the test statistic $f_t$ has already decisively converged toward its respective steady-state expectation—decaying safely toward zero under the null regime, or compounding rapidly under the alternative.
\subsection{General Remark}
In conclusion,
the theoretical foundations of both detection frameworks remain rigorously valid under the relaxed MDS condition (Assumption 2). For the SCPD procedure, the exact analytical guarantees and threshold configurations derived under Assumption 1 hold mathematically without modification. For the SHT framework, while the exact concentration inequalities necessitate slightly different thresholds, the resulting operational boundaries exhibit the same convergence rate with $t$ and  near identical practical behavior. This, along with the detailed investigations of section \ref{sec:main_Results} confirms that the computationally efficient i.i.d. formulation serves as a safe and reliable proxy for real-time deployment. 

While formally extending these mathematical guarantees to even more relaxed data regimes—such as strong mixing or unbounded temporal dependencies—remains a compelling direction for future theoretical work, our comprehensive empirical evaluations suggest such expansions may not be strictly necessary for practical deployment. As demonstrated by the detailed numerical investigations across diverse non-linear tracking scenarios, both the i.i.d. and MDS formulations inherently yield highly accurate and robust detection performance. This empirical resilience indicates that the proposed theoretical frameworks already provide a more than sufficient structural foundation for reliable real-world operation.
\newpage
\section{CUMAD Results}
\subsection{ Target Tracking}\label{ap:cumad_res_tracking}
The performance evaluation of CUMAD for each considered $\theta_0$ $\theta_1$ is provided in table \ref{tab:cumad_appendix_fps_fns_tracking}. As you can see CUMAD consistently fails to satisfy both error constraints.
\begin{table}[H]
    \centering
    \caption{Performance evaluation of the CUMAD baseline across linear 2D, nonlinear 2D, and nonlinear 3D target tracking environments, showing False Positive (FP) and False Negative (FN) rates. Values exceeding 5\% are highlighted in red.}\label{tab:cumad_appendix_fps_fns_tracking}
    \begin{tabular}{l|c|c|c|c}
        \hline
        \textbf{Environment} & $\theta_0$ & $\theta_1$ & \textbf{FP (\%)} & \textbf{FN (\%)} \\ \hline
        \multirow{9}{*}{\textbf{Linear 2D}} 
         & 0.10 & 0.70 & \textcolor{red}{45.7} & \textcolor{red}{52.7} \\
         & 0.10 & 0.80 & \textcolor{red}{40.7} & \textcolor{red}{56.7} \\
         & 0.10 & 0.90 & \textcolor{red}{38.0} & \textcolor{red}{58.3} \\
         & 0.20 & 0.70 & \textcolor{red}{40.7} & \textcolor{red}{55.7} \\
         & 0.20 & 0.80 & \textcolor{red}{37.3} & \textcolor{red}{60.3} \\
         & 0.20 & 0.90 & \textcolor{red}{37.0} & \textcolor{red}{61.3} \\
         & 0.30 & 0.70 & \textcolor{red}{32.0} & \textcolor{red}{58.7} \\
         & 0.30 & 0.80 & \textcolor{red}{31.7} & \textcolor{red}{65.0} \\
         & 0.30 & 0.90 & \textcolor{red}{31.3} & \textcolor{red}{67.0} \\ \hline
        \multirow{9}{*}{\textbf{Nonlinear 2D}} 
         & 0.10 & 0.70 & \textcolor{red}{60.0} & \textcolor{red}{64.0} \\
         & 0.10 & 0.80 & \textcolor{red}{58.0} & \textcolor{red}{64.0} \\
         & 0.10 & 0.90 & \textcolor{red}{57.5} & \textcolor{red}{64.0} \\
         & 0.20 & 0.70 & \textcolor{red}{51.5} & \textcolor{red}{93.5} \\
         & 0.20 & 0.80 & \textcolor{red}{51.0} & \textcolor{red}{94.0} \\
         & 0.20 & 0.90 & \textcolor{red}{55.5} & \textcolor{red}{64.0} \\
         & 0.30 & 0.70 & \textcolor{red}{44.0} & \textcolor{red}{97.5} \\
         & 0.30 & 0.80 & \textcolor{red}{44.0} & \textcolor{red}{100.0} \\
         & 0.30 & 0.90 & \textcolor{red}{49.0} & \textcolor{red}{94.0} \\ \hline
        \multirow{9}{*}{\textbf{Nonlinear 3D}} 
         & 0.10 & 0.70 & \textcolor{red}{37.6} & \textcolor{red}{18.4} \\
         & 0.10 & 0.80 & \textcolor{red}{32.4} & \textcolor{red}{41.4} \\
         & 0.10 & 0.90 & \textcolor{red}{30.2} & \textcolor{red}{42.6} \\
         & 0.20 & 0.70 & \textcolor{red}{23.8} & \textcolor{red}{3.0} \\
         & 0.20 & 0.80 & \textcolor{red}{18.4} & \textcolor{red}{18.4} \\
         & 0.20 & 0.90 & \textcolor{red}{26.6} & \textcolor{red}{42.6} \\
         & 0.30 & 0.70 & \textcolor{red}{11.0} & \textcolor{red}{3.0} \\
         & 0.30 & 0.80 & \textcolor{red}{10.8} & \textcolor{red}{18.4} \\
         & 0.30 & 0.90 & \textcolor{red}{13.8} & \textcolor{red}{42.6} \\ \hline
    \end{tabular}
\end{table}
\newpage
\subsection{ Network Traffic Monitoring  Datasets}\label{ap:CUMAD_NetworkTraffic}
The Results of CUMAD for both CIC-IDS-2018 and CESNET 24 Datasets are provided in table \ref{tab:parameter_sweep_cumad_network_traffic} where its inconsistency is highlighted.\\
\begin{table}[H]
\centering
\caption{Parameter Sweep Performance Comparison for CUMAD across CIC-IDS-2018 and CESNET Datasets}
\label{tab:parameter_sweep_cumad_network_traffic}
\begin{tabular}{lccc}
\toprule
\textbf{Dataset} & \textbf{Parameters ($\theta_0, \theta_1$)} & \textbf{FP (\%)} & \textbf{FN (\%)} \\
\midrule
\multirow{9}{*}{\textbf{CIC-IDS-2018}} 
& 0.1, 0.7 & 0.00 & \textcolor{red}{54.60} \\
& 0.2, 0.7 & 0.00 & \textcolor{red}{55.20} \\
& 0.3, 0.7 & 0.00 & \textcolor{red}{54.80} \\
& 0.1, 0.8 & 0.00 & \textcolor{red}{55.60} \\
& 0.2, 0.8 & 0.00 & \textcolor{red}{57.80} \\
& 0.3, 0.8 & 0.00 & \textcolor{red}{60.40} \\
& 0.1, 0.9 & 0.00 & \textcolor{red}{51.80} \\
& 0.2, 0.9 & 0.00 & \textcolor{red}{61.40} \\
& 0.3, 0.9 & 0.00 & \textcolor{red}{68.40} \\
\midrule
\multirow{9}{*}{\textbf{CESNET}} 
& 0.1, 0.7 & \textcolor{red}{98.80} & 0.00 \\
& 0.1, 0.8 & \textcolor{red}{98.40} & 0.00 \\
& 0.1, 0.9 & \textcolor{red}{97.00} & 0.00 \\
& 0.2, 0.7 & \textcolor{red}{82.60} & 0.00 \\
& 0.2, 0.8 & \textcolor{red}{82.80} & 0.00 \\
& 0.2, 0.9 & \textcolor{red}{89.40} & 0.00 \\
& 0.3, 0.7 & \textcolor{red}{58.40} & 0.00 \\
& 0.3, 0.8 & \textcolor{red}{67.60} & 0.00 \\
& 0.3, 0.9 & \textcolor{red}{69.60} & 0.00 \\
\bottomrule
\end{tabular}
\end{table}

\subsection{H\'enon Map}\label{ap:CUMAD_res_Henon}
\begin{table}[H]
\centering
\caption{Standard Windowed CUMAD Performance across Grid Search Parameters (Mean $\pm$ Std. Dev. over 3 runs). Error rates exceeding the $5\%$ constraint are highlighted in \textcolor{red}{red}. \textbf{Hénon Map Experiment}}
\label{tab:cumad_window_results_henon}
\begin{tabular}{cc r@{$\pm$}l r@{$\pm$}l}
\toprule
$\theta_0$ & $\theta_1$ & \multicolumn{2}{c}{\textbf{FN (\%)}} & \multicolumn{2}{c}{\textbf{FP (\%)}} \\
\midrule
0.1 & 0.7 & \textcolor{red}{47.5} & 7.6 & \textcolor{red}{11.2} & 6.1 \\
0.1 & 0.8 & \textcolor{red}{50.3} & 7.1 & \textcolor{red}{10.5} & 6.3 \\
0.1 & 0.9 & \textcolor{red}{23.7} & 7.1 & \textcolor{red}{9.2} & 6.3 \\
\midrule
0.2 & 0.7 & \textcolor{red}{48.7} & 11.3 & \textcolor{red}{9.1} & 6.9 \\
0.2 & 0.8 & \textcolor{red}{56.6} & 10.3 & \textcolor{red}{7.7} & 6.4 \\
0.2 & 0.9 & \textcolor{red}{17.8} & 10.3 & 4.1 & 3.2 \\
\midrule
0.3 & 0.7 & \textcolor{red}{62.3} & 17.3 & 4.7 & 4.4 \\
0.3 & 0.8 & \textcolor{red}{69.1} & 15.6 & 4.0 & 4.3 \\
0.3 & 0.9 & \textcolor{red}{52.5} & 16.3 & 3.9 & 3.2 \\
\bottomrule
\end{tabular}
\end{table}
\subsection{Lorenz-Chen System}
\label{ap:CUMAD_res_Lorenz_Chen}
\begin{table}[H]
\centering
\caption{Standard Windowed CUMAD Performance across Grid Search Parameters (Mean $\pm$ Std. Dev. over 3 runs). Error rates exceeding the $5\%$ constraint are highlighted in \textcolor{red}{red}. \textbf{Lorenz-Chen Experiment}}
\label{tab:cumad_window_results_lorenz_chen}
\begin{tabular}{cc r@{$\pm$}l r@{$\pm$}l}
\toprule
$\theta_0$ & $\theta_1$ & \multicolumn{2}{c}{\textbf{FN (\%)}} & \multicolumn{2}{c}{\textbf{FP (\%)}} \\
\midrule
0.1 & 0.7 & \textcolor{red}{42.3} & 11.7 & \textcolor{red}{5.3} & 2.2 \\
0.1 & 0.8 & \textcolor{red}{42.3} & 11.6 & 
\textcolor{red}{5.3} & 2.2 \\
0.1 & 0.9 & \textcolor{red}{42.3} & 11.7 & 
\textcolor{red}{5.2} & 2.2 \\
\midrule
0.2 & 0.7 & \textcolor{red}{39.1} & 12.4 & \textcolor{red}{5.8} & 2.3 \\
0.2 & 0.8 & \textcolor{red}{42.4} & 11.7 & \textcolor{red}{5.3} & 2.2 \\
0.2 & 0.9 & \textcolor{red}{48.5} & 11.2 & 5.0 & 2.1 \\
\midrule
0.3 & 0.7 & \textcolor{red}{41.3} & 13.5 & \textcolor{red}{5.4} & 2.1 \\
0.3 & 0.8 & \textcolor{red}{44.4} & 12.7 & 5.0 & 2.1 \\
0.3 & 0.8 & \textcolor{red}{40.1} & 12.9 & 4.8 & 2.1 \\
\bottomrule
\end{tabular}
\end{table}
\section{Investigating Raw Scan-B and PCA-based Change Detectors in Target Tracking}\label{ap:raw_scanB_PCA_tracking}
\begin{figure*}[h!]
    \centering
    \begin{subfigure}[b]{0.32\textwidth}
        \centering
        \includegraphics[width=\textwidth]{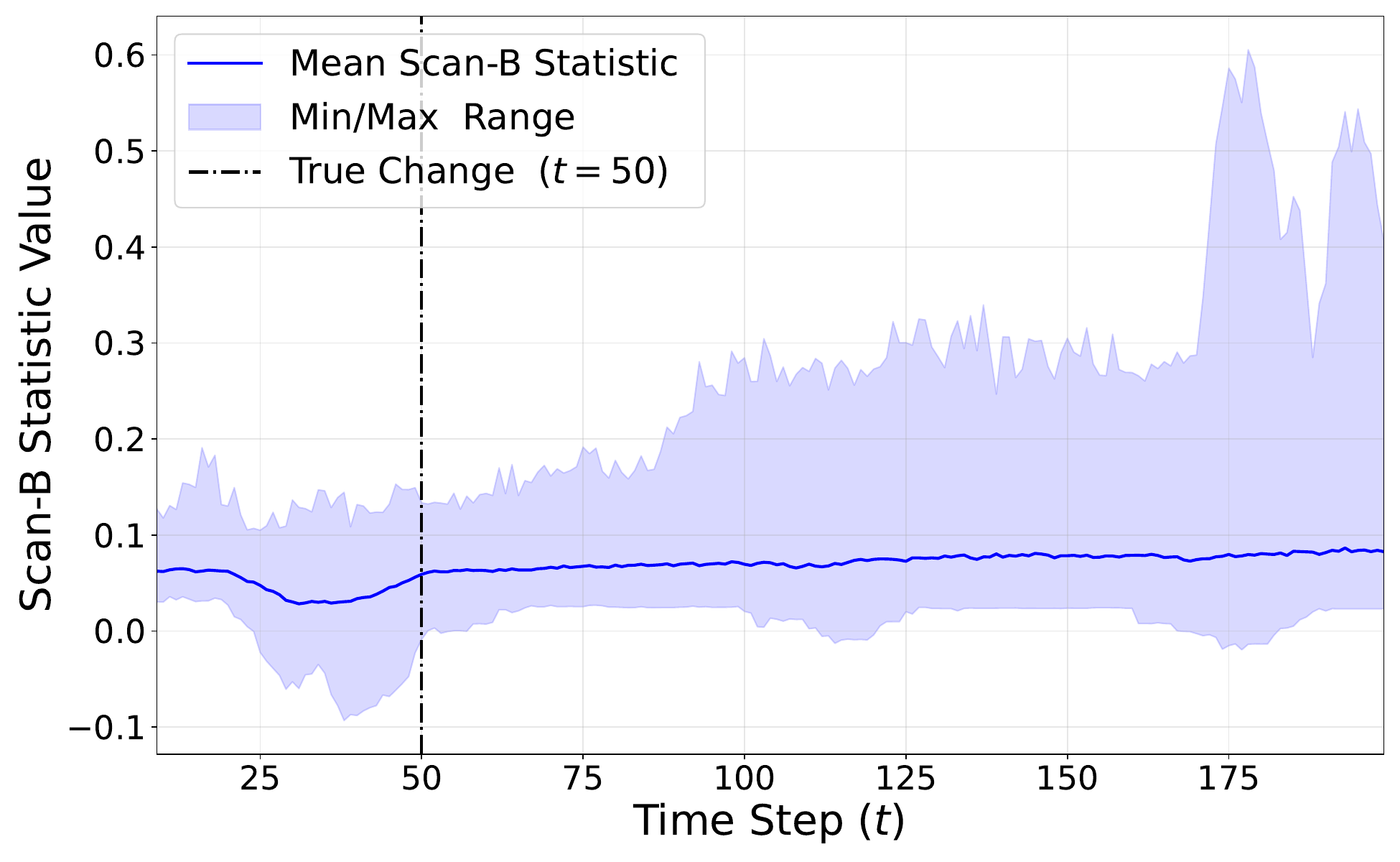}
        \caption{$B_0 = 10$}
        \label{fig:scan_b_raw_10}
    \end{subfigure}
    \hfill
    \begin{subfigure}[b]{0.32\textwidth}
        \centering
        \includegraphics[width=\textwidth]{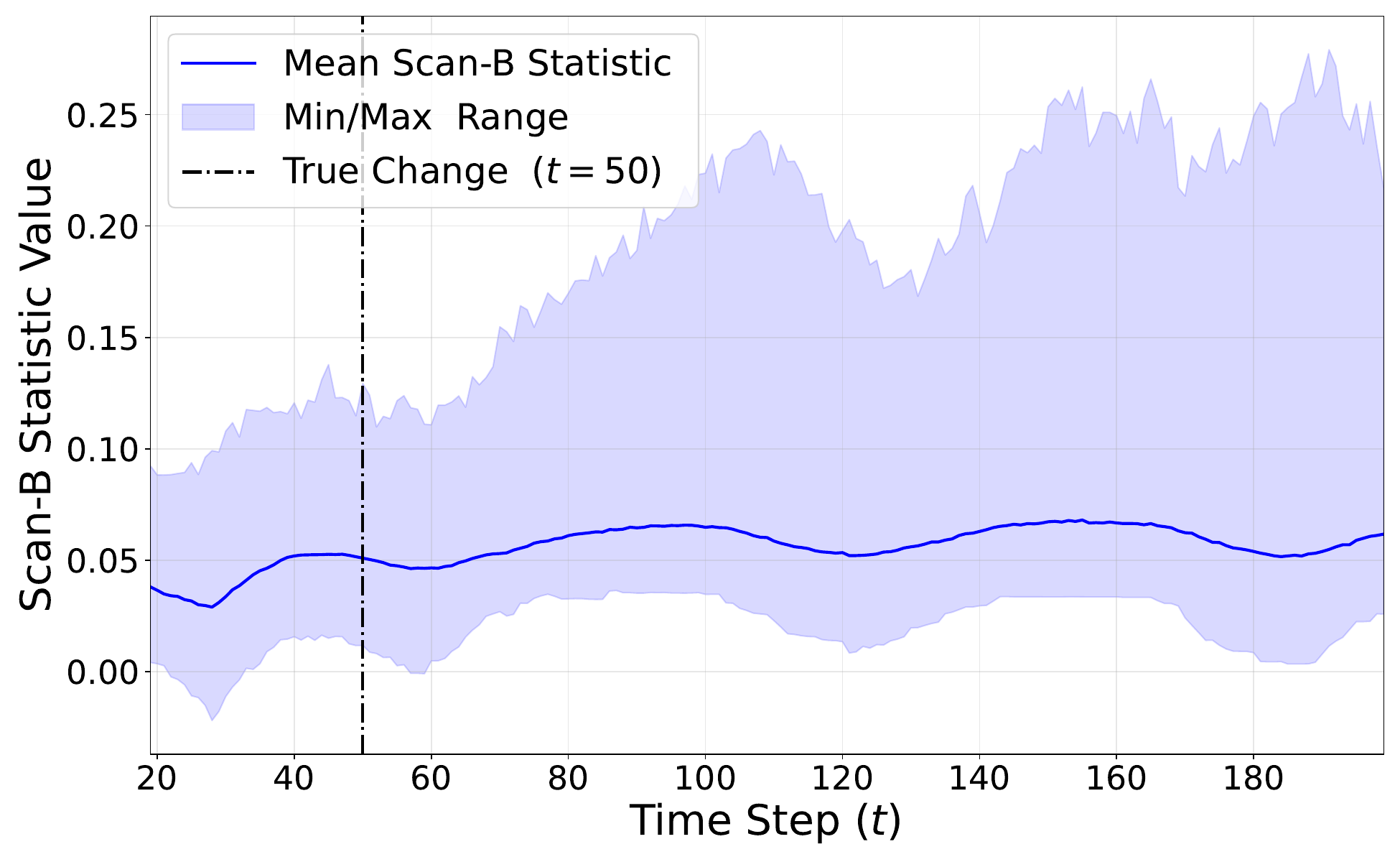}
        \caption{$B_0 = 20$}
        \label{fig:scan_b_raw_20}
    \end{subfigure}
    \hfill
    \begin{subfigure}[b]{0.32\textwidth}
        \centering
        \includegraphics[width=\textwidth]{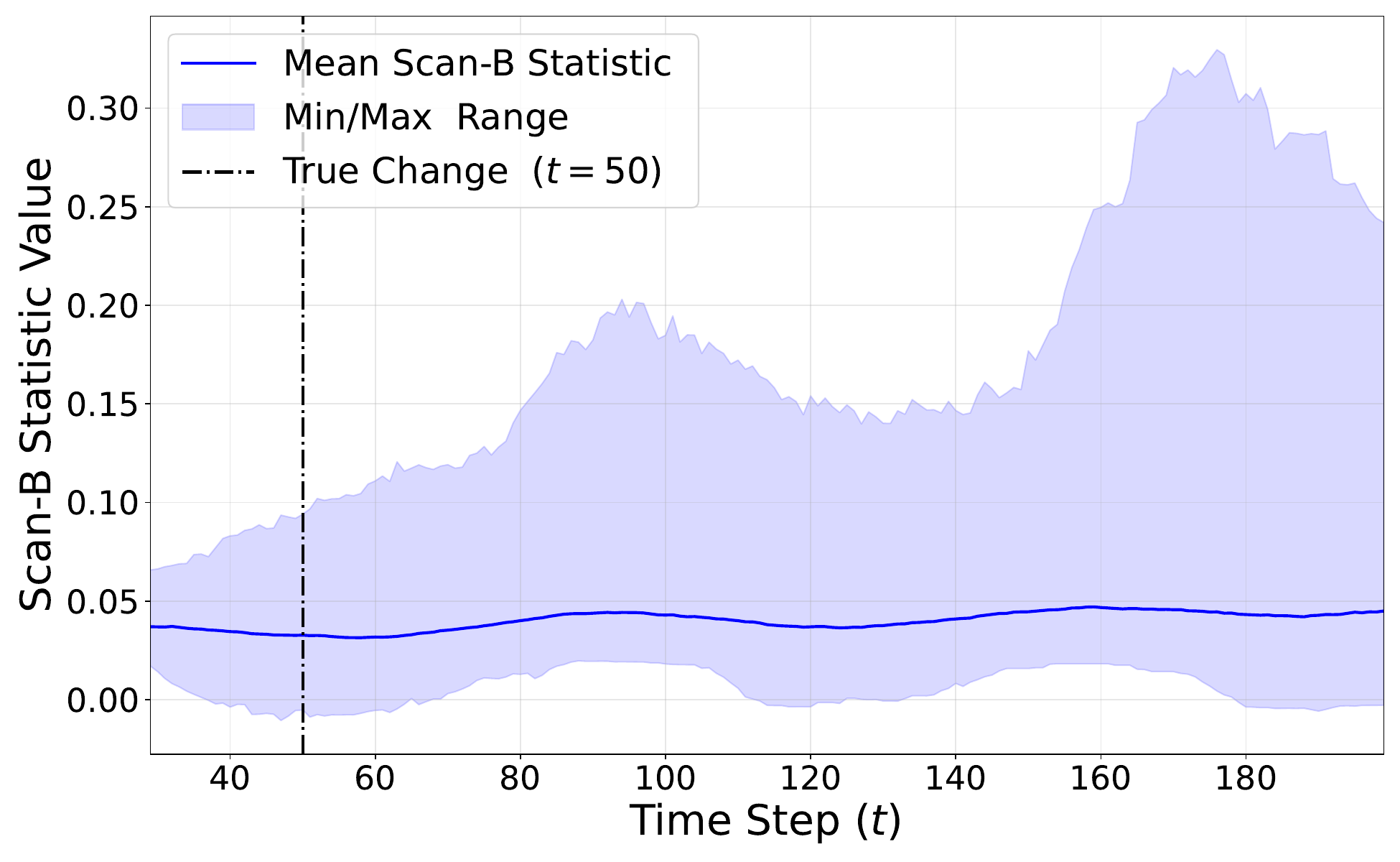}
        \caption{$B_0 = 30$}
        \label{fig:scan_b_raw_30}
    \end{subfigure}
    \caption{Scan-B test statistic applied directly to raw target tracking observations across varying block sizes ($B_0 \in \{10, 20, 30\}$).}
\label{fig:scan_b_tracking_raw}
\end{figure*}

First, we investigate the performance of the Scan-B method~\cite{ScanB} applied on the raw measurements of the target tracking simulation in section \ref{sec:CPD_main_res}. 
In Fig. \ref{fig:scan_b_tracking_raw}, for each $t$ we plot the average value of the statistic, along with the maximum and minimum values of all simulated trajectories in a shaded region. As is apparent, the post-change shift in the average value is negligible while the statistic displays extreme variance making change detection impossible. 

This failure can be attributed to the design of Scan-B as it  computes kernel distances in order to  detect changes between recent blocks of data. However, because we consider a moving target, the raw spatial coordinates in one time block will naturally be far away from the coordinates in the next block. Consequently, Scan-B interprets this natural shift as a change. Similar values of the statistic are computed for both the pre-change and the post-change target motion, and a separating threshold cannot be configured to distinguish between them.

\begin{figure}[H]
    \centering
    \begin{subfigure}[b]{0.48\columnwidth}
        \centering
\includegraphics[width=\textwidth]{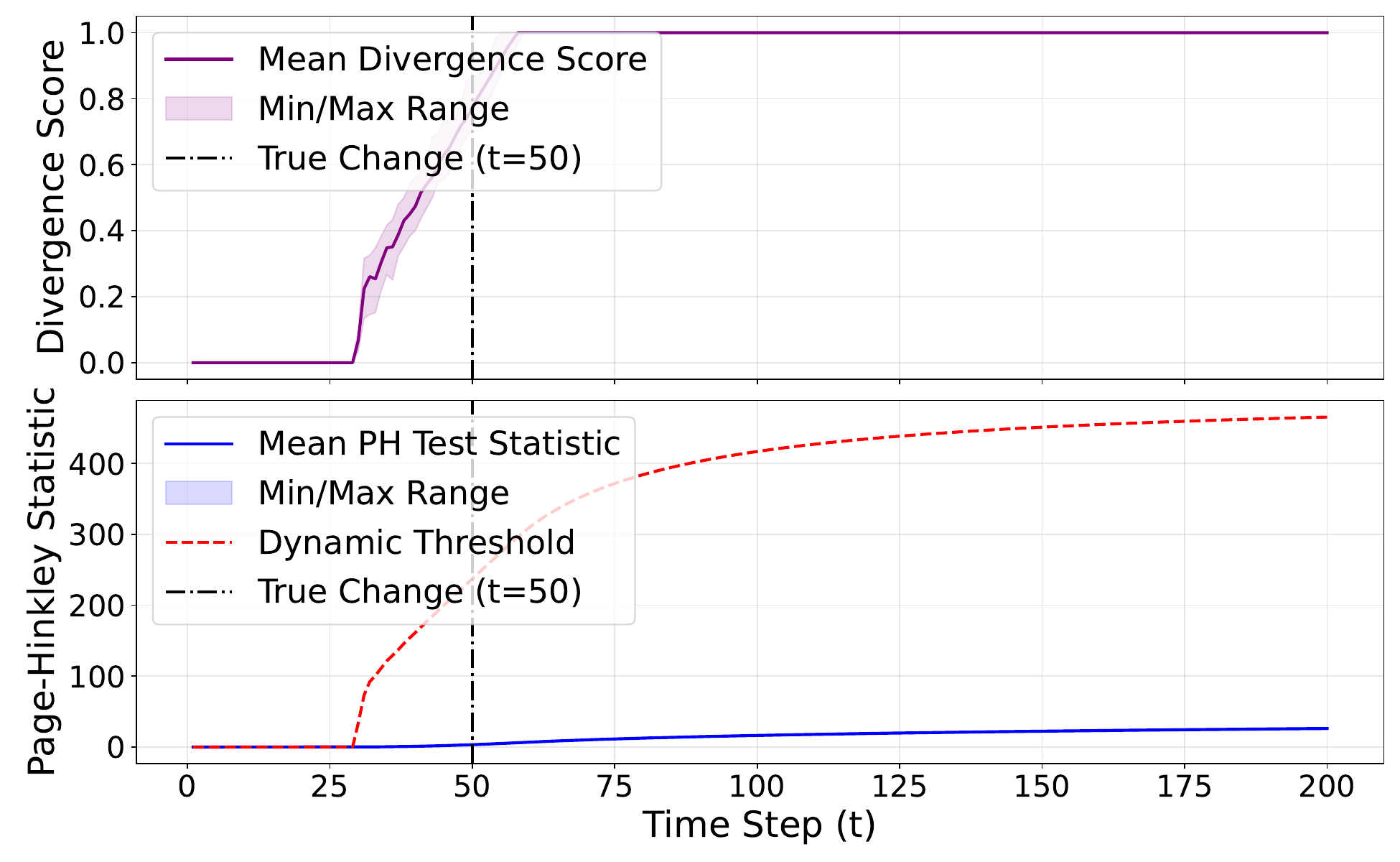}
        \caption{PCA-based Detector (KDD15)}
    \end{subfigure}
    \hfill
    \begin{subfigure}[b]{0.48\columnwidth}
        \centering
        \includegraphics[width=\textwidth]{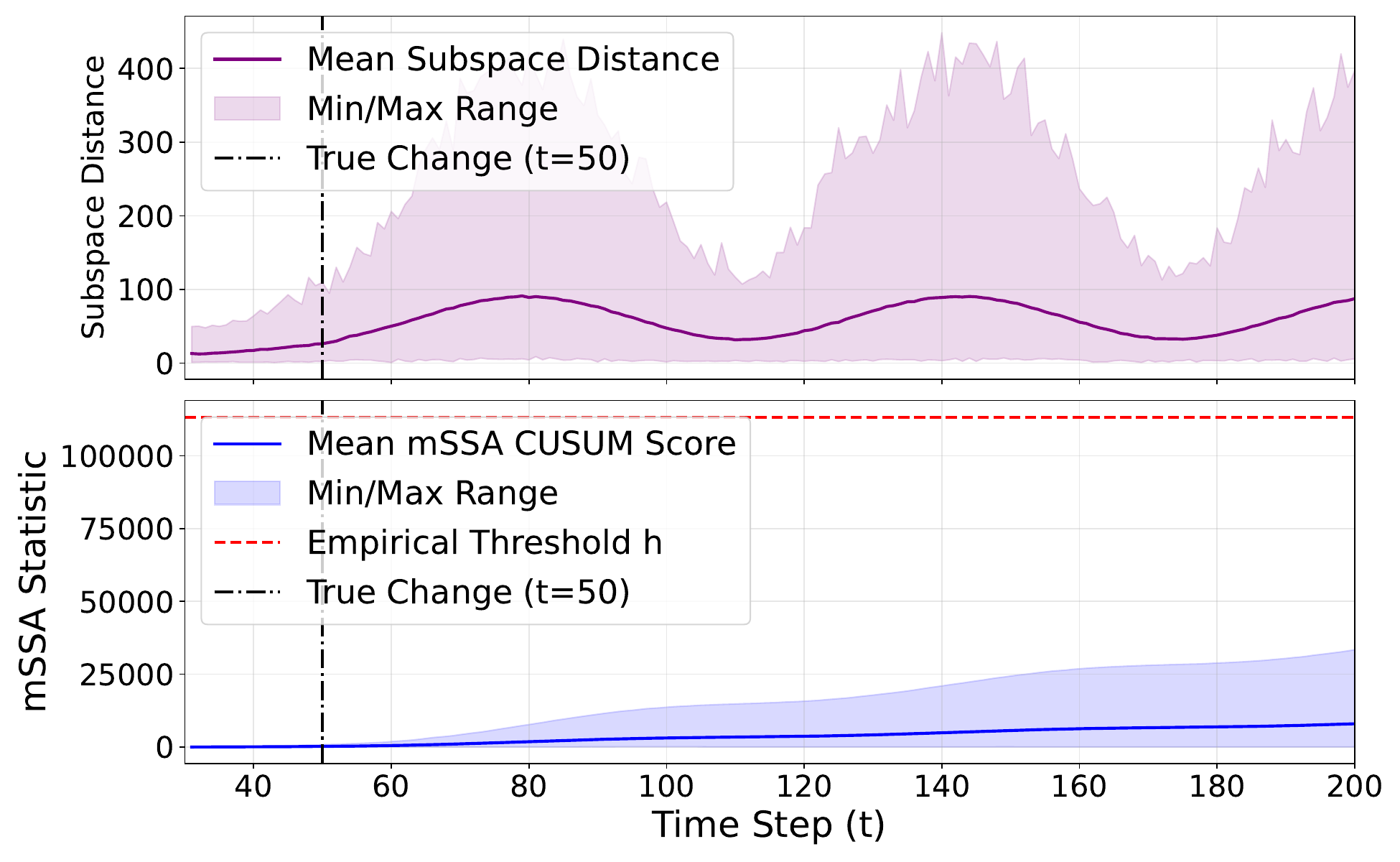}
        \caption{mSSA Detector}
 
    \end{subfigure}
    \caption{Diagnostic trajectories for the linear subspace SCPD methods on the 3D target tracking environment.}
\label{fig:linear_subspace_failures_target_tracking}
\end{figure}

A similar visualization for the test statistics of \cite{PCA_KDD_15} and \cite{msSA} is provided in Fig.  \ref{fig:linear_subspace_failures_target_tracking}. The PCA-based  framework of \cite{PCA_KDD_15} completely fails to detect the change because its dynamic PH threshold (dashed red line) exponentially inflates during the nominal regime, due to the proposed heuristic  rule. Furthermore,  both the divergence score and the  statistic actually  begin increasing well before the change time. Conversely, the mSSA detector  suffers from extreme variance in its subspace distance metric (shaded purple region) due to the highly nonlinear system evolution violating its core assumptions. This  variance forces a massive empirical threshold ($h > 120,000$), permanently suppressing the CUSUM statistic and resulting in complete  detection failure.
\section{Additional CPD Experiments}\label{ap:extra_cpd_experiments}
\subsection{Network Traffic Monitoring}
To evaluate the proposed error-driven CUSUM framework on high-dimensional network streams, we simulate $1000$ Monte Carlo episodes of length $T_{max}=200$ using the CIC-IDS-2018 dataset. Each trajectory begins with nominal traffic and abruptly transitions to a cyberattack at $\tau^*=50$. We construct these online evaluation episodes by first sampling a contiguous $50$-step sequence of 'benign' traffic, and immediately appending a $150$-step attack sequence, with the specific intrusion profile sampled randomly from the diverse pool of attacks available in the dataset. The same AE model as the SHT CIC-IDS-2018 experiments is used.

\begin{figure}[H]
    \centering
    \begin{subfigure}[b]{0.48\textwidth}
        \centering
\includegraphics[width=\textwidth]{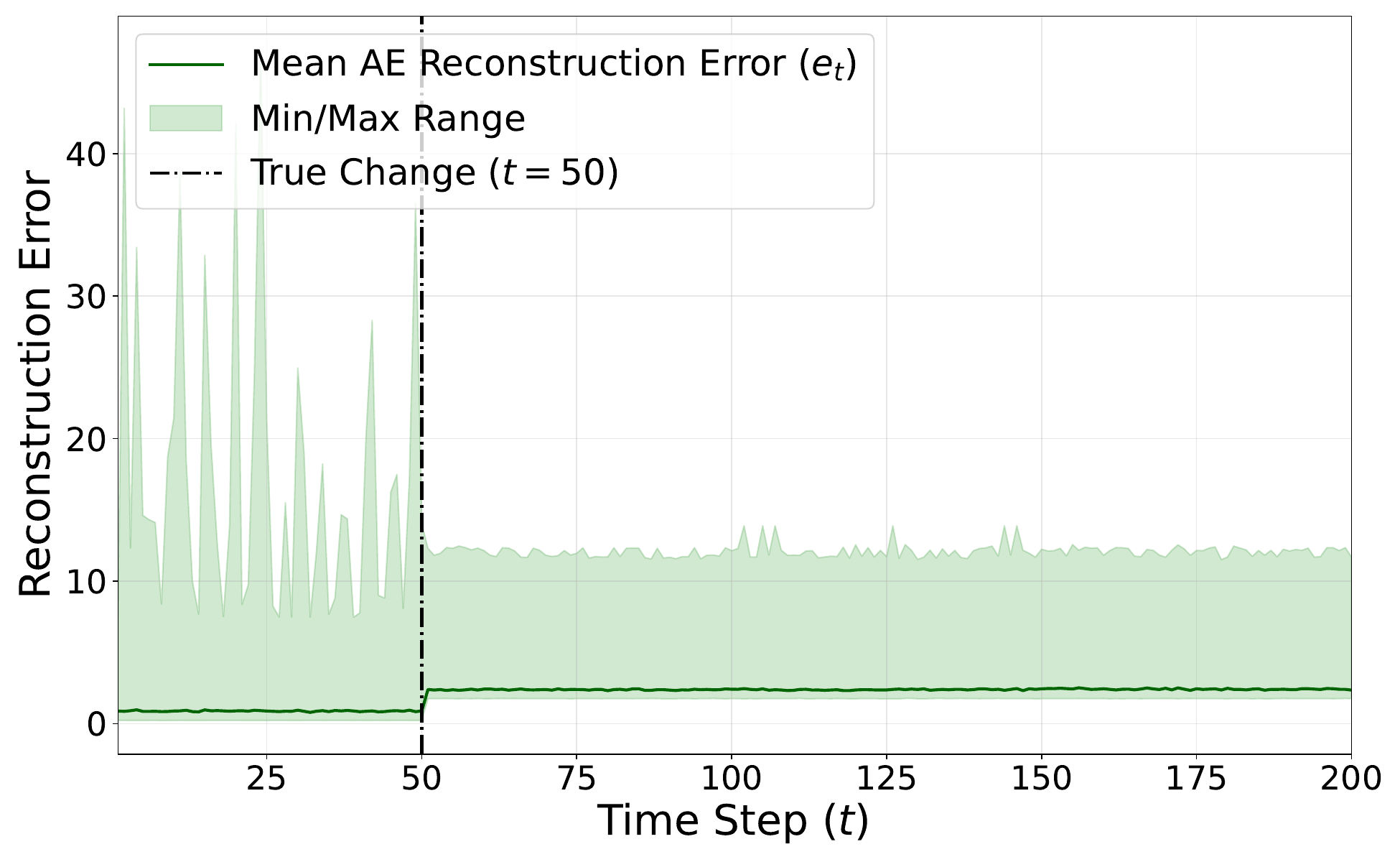}
     \caption{Reconstruction error}
   \end{subfigure}
    \hfill
    \begin{subfigure}[b]{0.48\textwidth}
    \centering    \includegraphics[width=\textwidth]{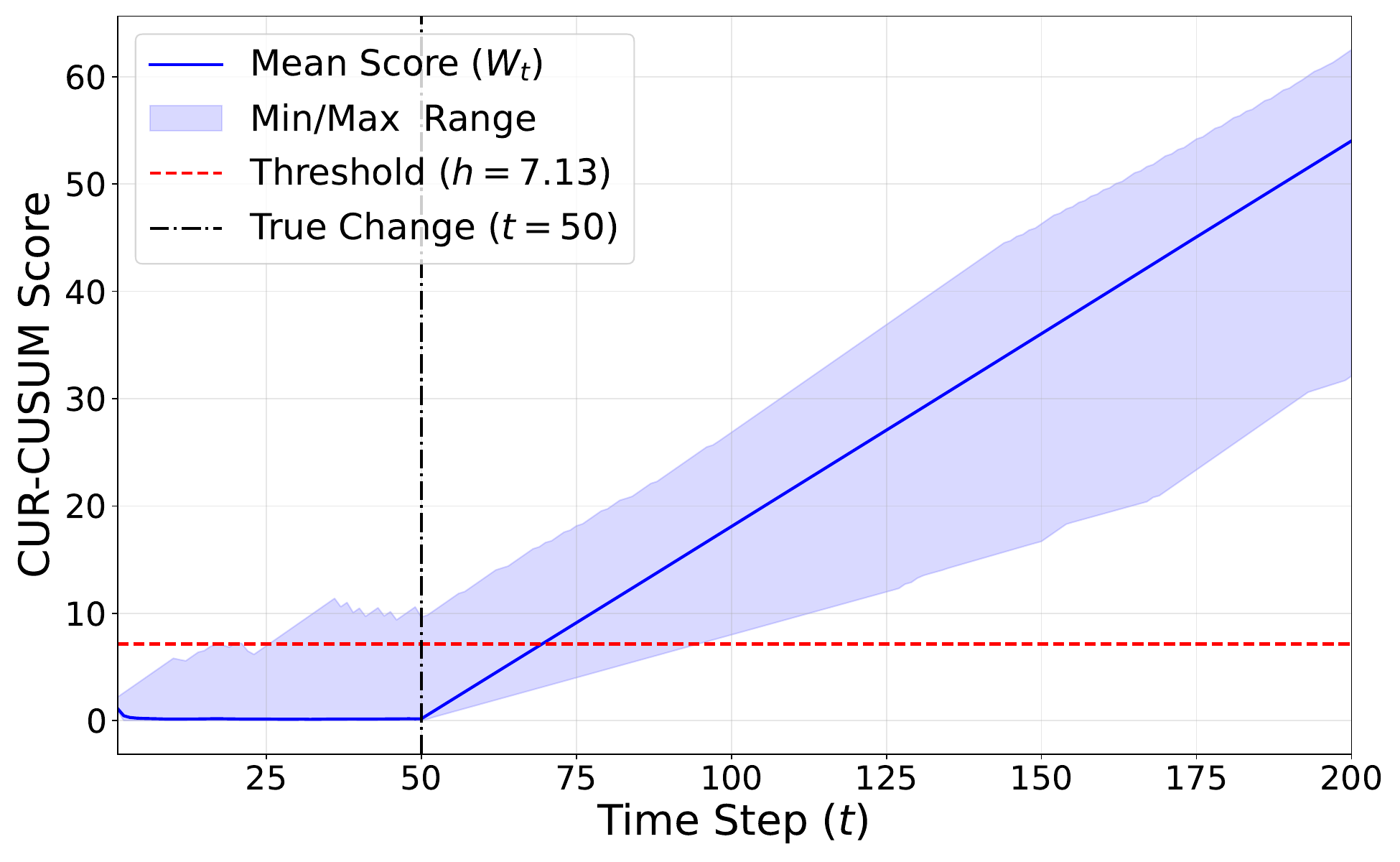}
        \caption{Proposed test statistic}
    \end{subfigure}
    \caption{Trajectory comparison of the AE reconstruction error and the derived test statistic.}
\label{fig:side_by_side_Network_CPD}
\end{figure}

CUR-CUSUM achieved an empirical early declaration  rate of $1.2\%$, zero misses, and an average detection delay of $18.52$ steps. Fig. \ref{fig:side_by_side_Network_CPD} plots the  reconstruction error alongside the proposed test statistic. During the pre-change regime, the error exhibits high variance around a small mean (highlighting the importance of clipping); post-change, the mean   increases while the variance stabilizes. The proposed  statistic remains  flat near $0$ during nominal operation and immediately exhibits a steady, linear accumulation following $\tau^*$.

\begin{figure}[H]
    \centering
    
    \begin{subfigure}[b]{0.48\textwidth}
        \centering
        \includegraphics[width=\textwidth]{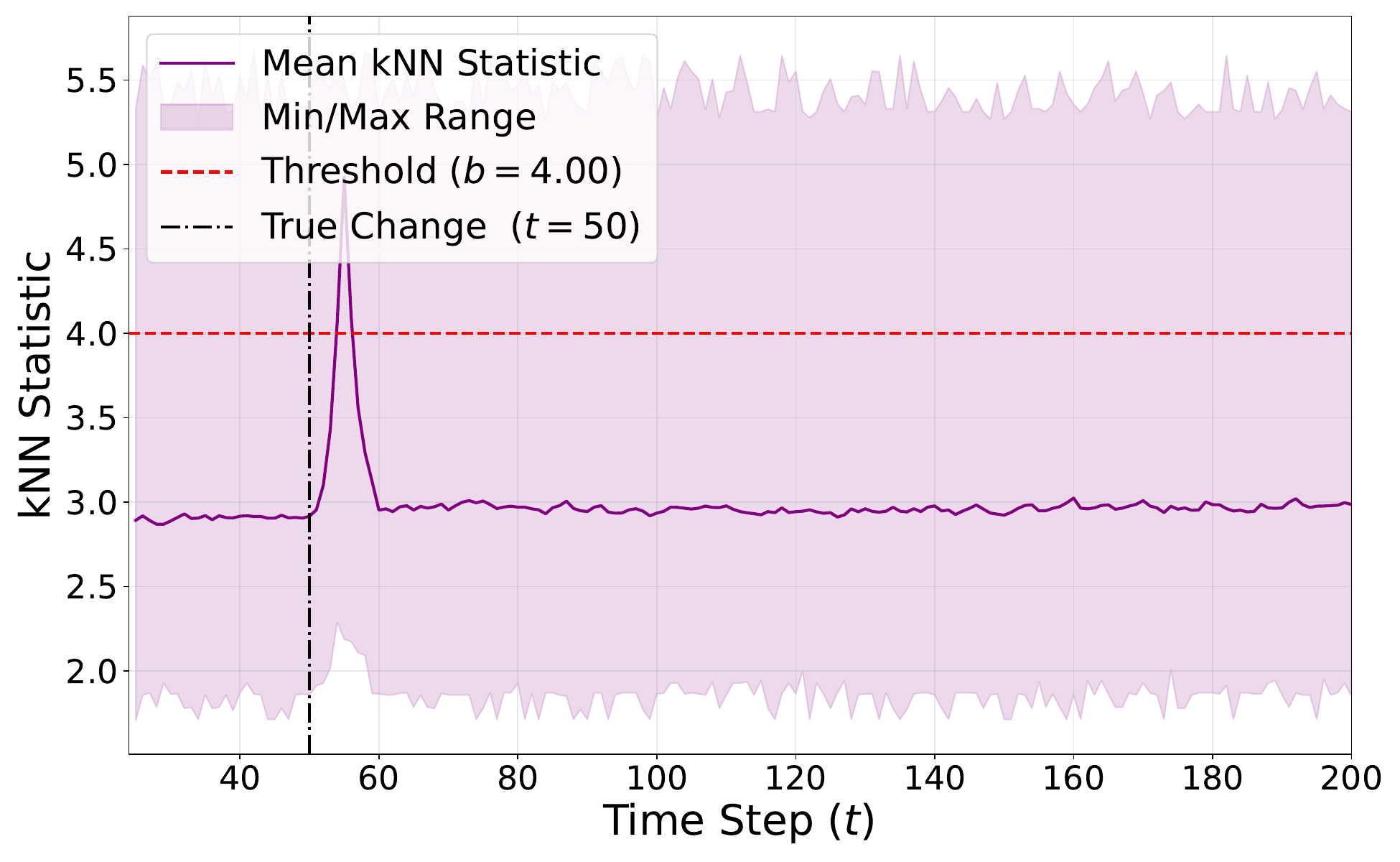}
        \caption{$L=10, k=3$}
        \label{fig:nn_cpd_10_k3}
    \end{subfigure}
    \hfill
    \begin{subfigure}[b]{0.48\textwidth}
        \centering
        \includegraphics[width=\textwidth]{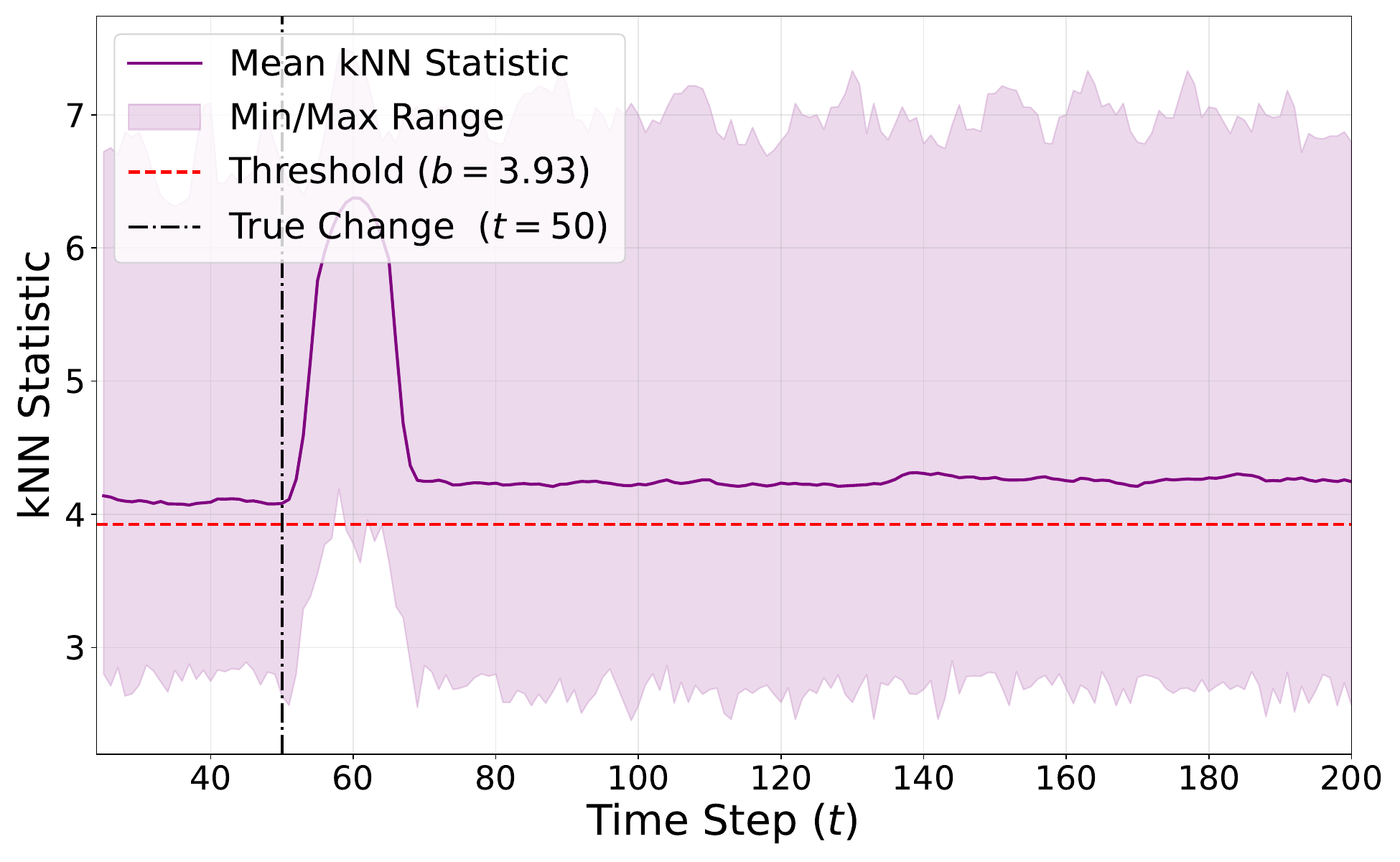}
        \caption{$L=20, k=3$}
        \label{fig:nn_cpd_20_k3}
    \end{subfigure}
    
    \vspace{0.4cm} 
    
    \begin{subfigure}[b]{0.48\textwidth}
        \centering
        \includegraphics[width=\textwidth]{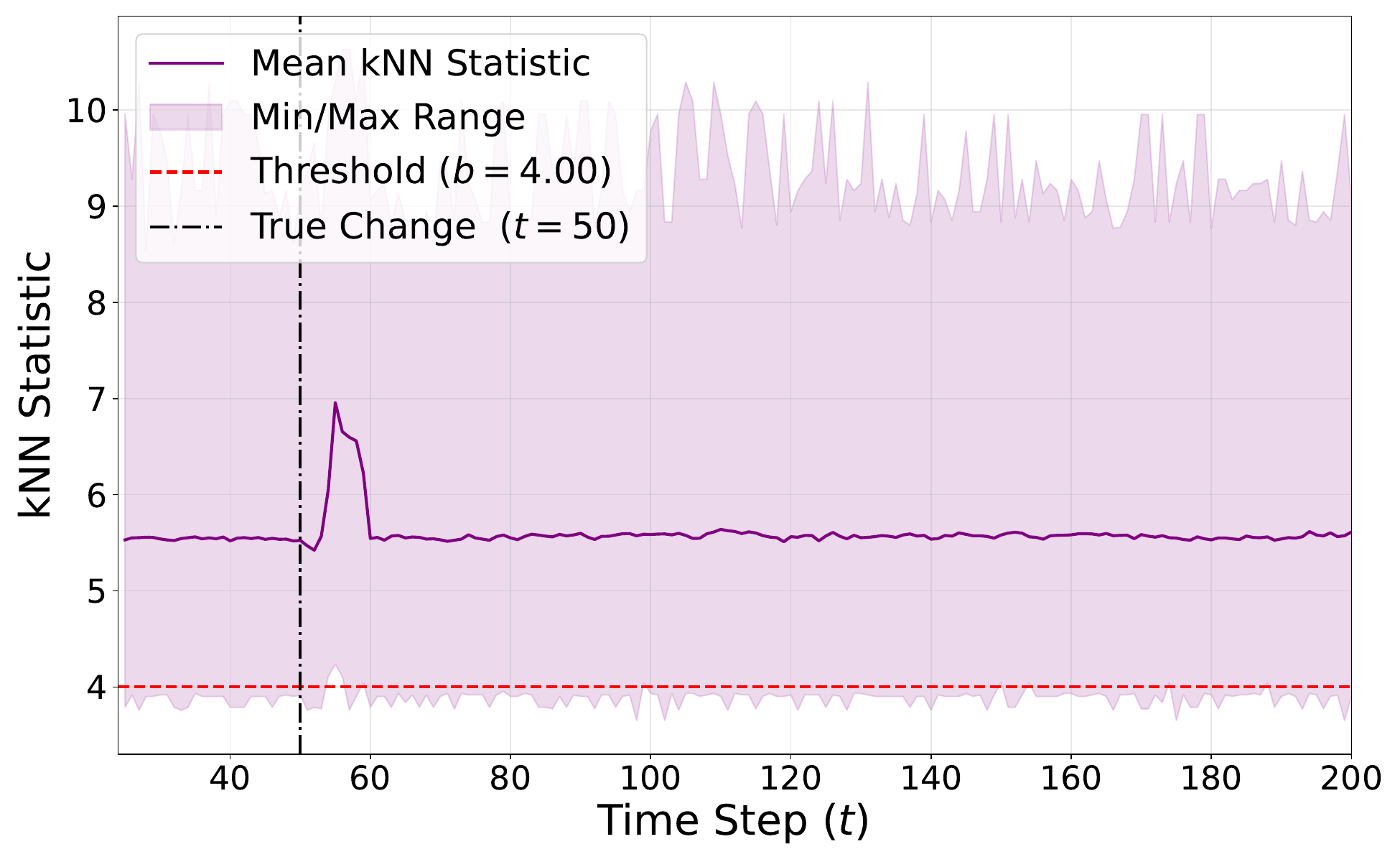}
        \caption{$L=10, k=5$}
        \label{fig:nn_cpd_10_k5}
    \end{subfigure}
    \hfill
    \begin{subfigure}[b]{0.48\textwidth}
        \centering
        \includegraphics[width=\textwidth]{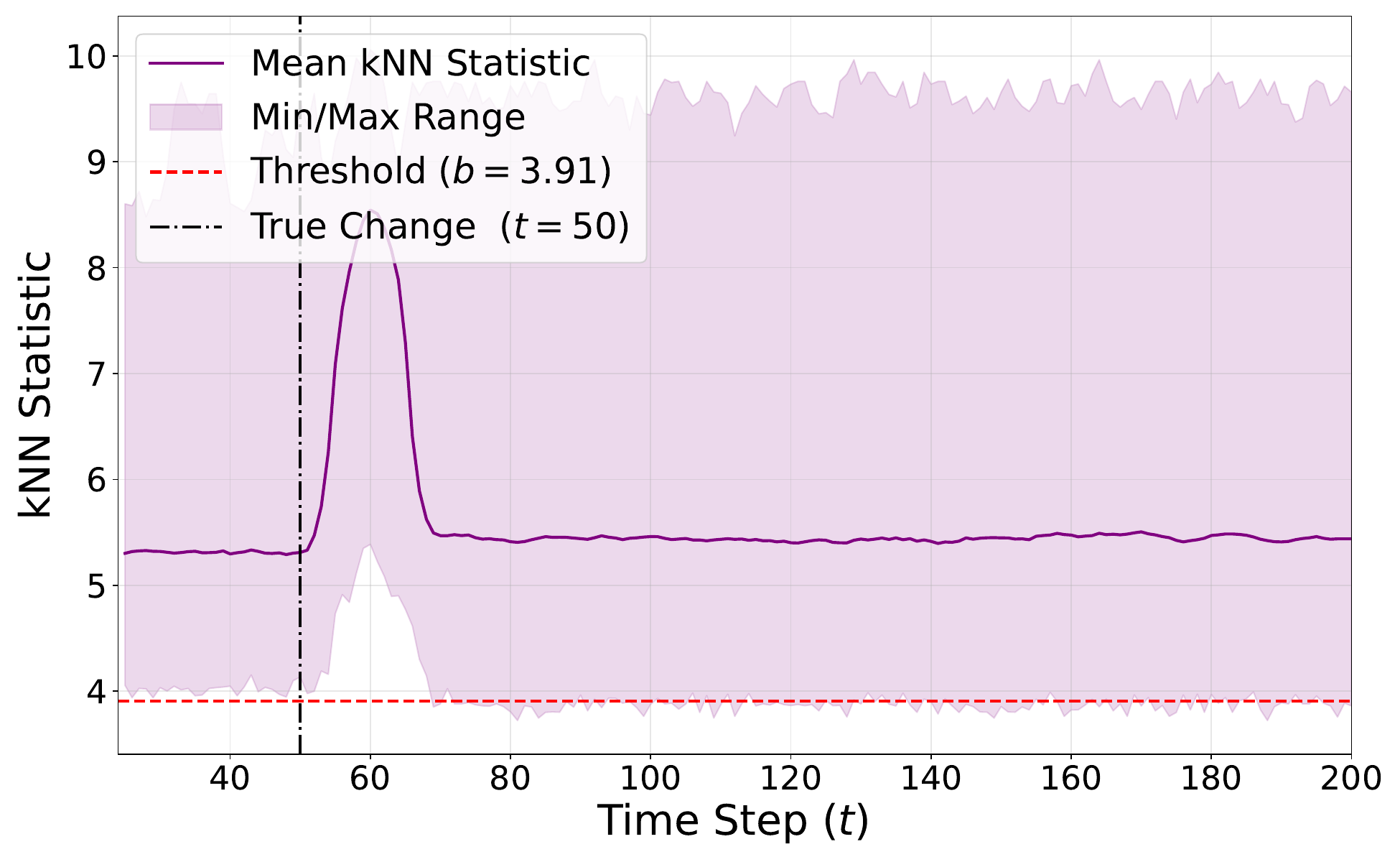}
        \caption{$L=20, k=5$}
        \label{fig:nn_cpd_20_k5}
    \end{subfigure}
    
    \caption{Visualization of the kNN-graph test statistic applied to AE reconstruction errors on the CIC-IDS-2018 dataset.} 
    \label{fig:nn_cpd_sensitivity_network}
\end{figure}

\begin{figure}[H]
    \centering
    \begin{subfigure}[b]{0.32\textwidth}
        \centering
        \includegraphics[width=\textwidth]{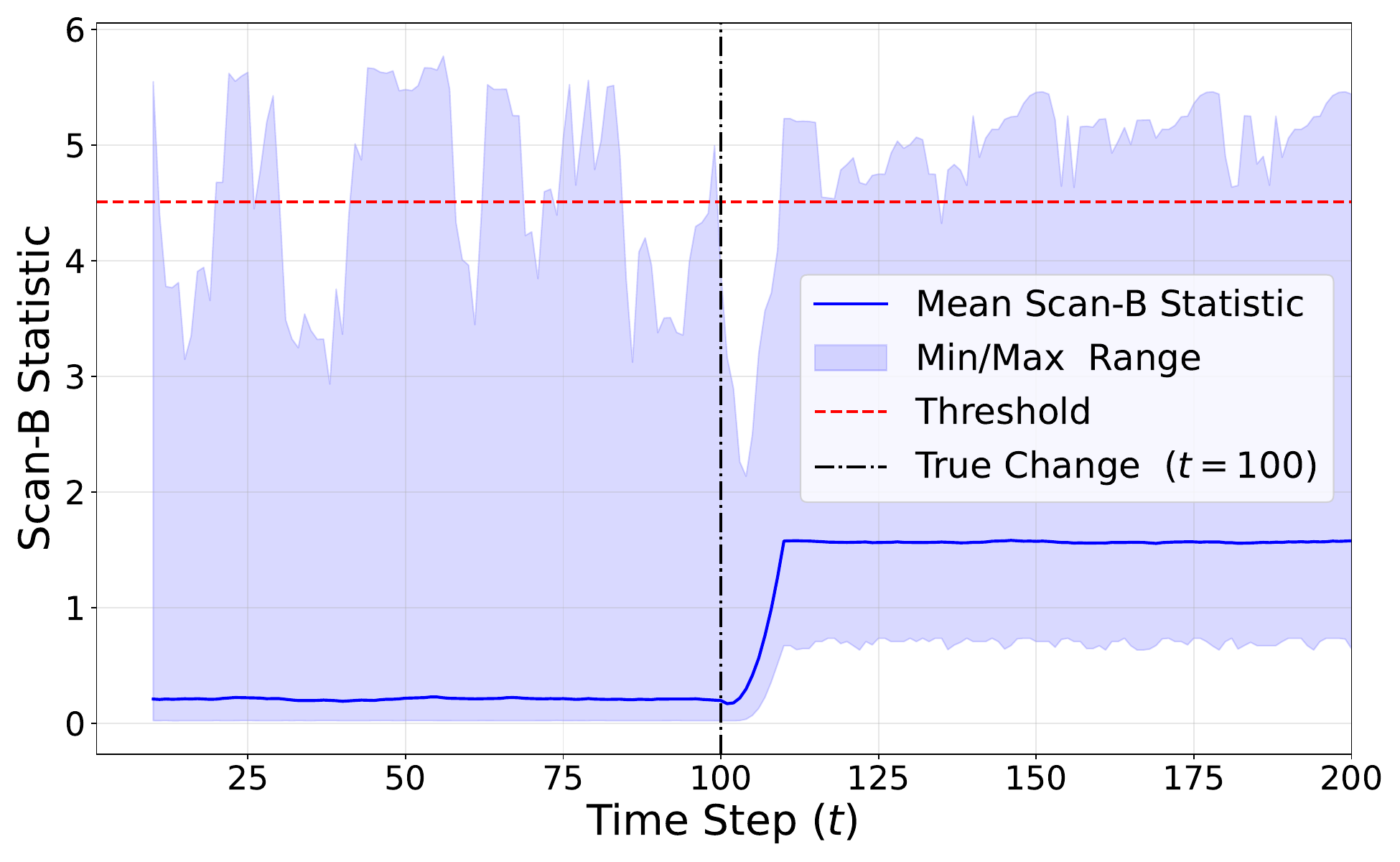}
        \caption{$B_0 = 10$}
        \label{fig:scan_b_net_10}
    \end{subfigure}
    \hfill
    \begin{subfigure}[b]{0.32\textwidth}
        \centering
        \includegraphics[width=\textwidth]{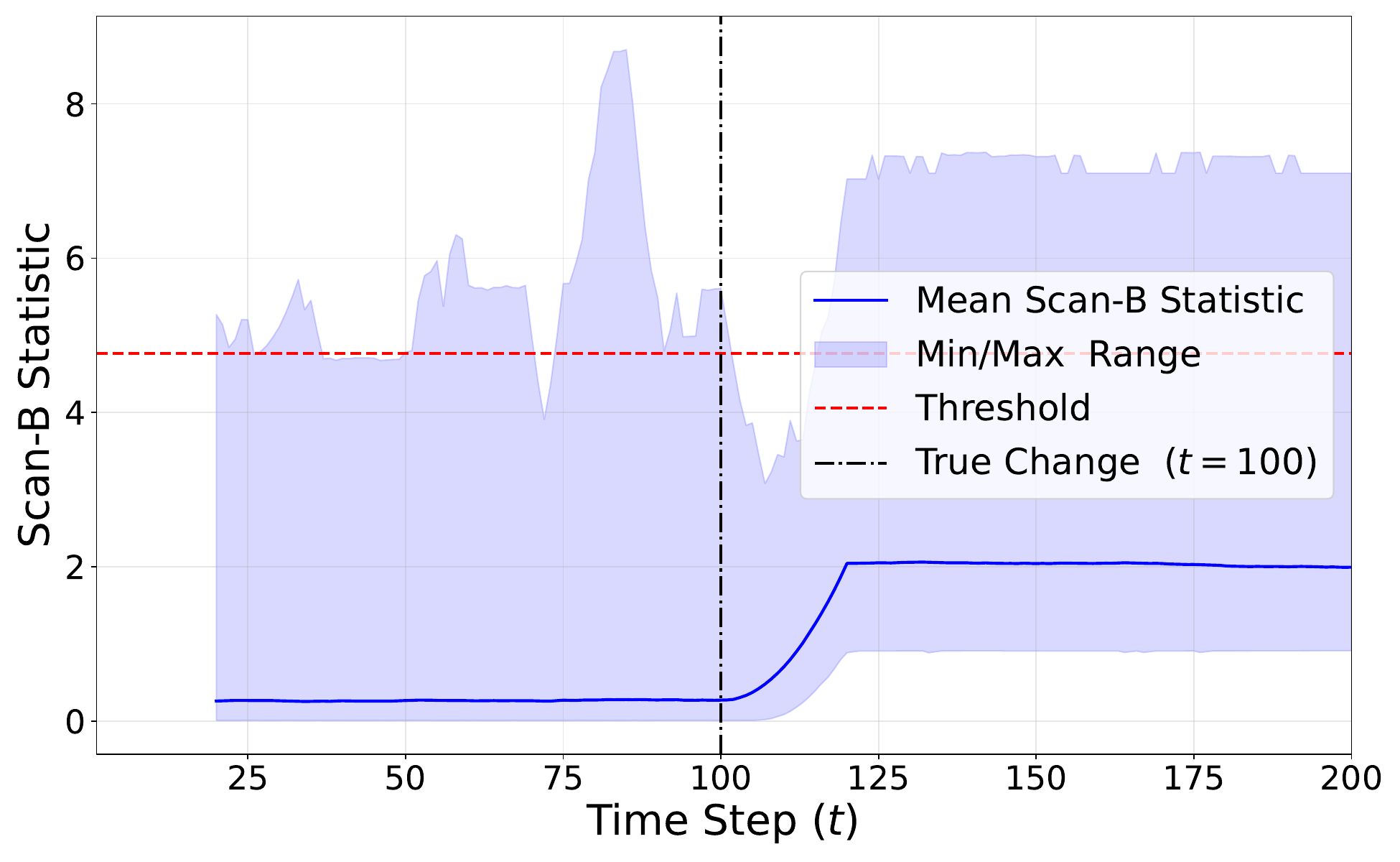}
        \caption{$B_0 = 20$}
        \label{fig:scan_b_net_20}
    \end{subfigure}
    \hfill
    \begin{subfigure}[b]{0.32\textwidth}
        \centering
        \includegraphics[width=\textwidth]{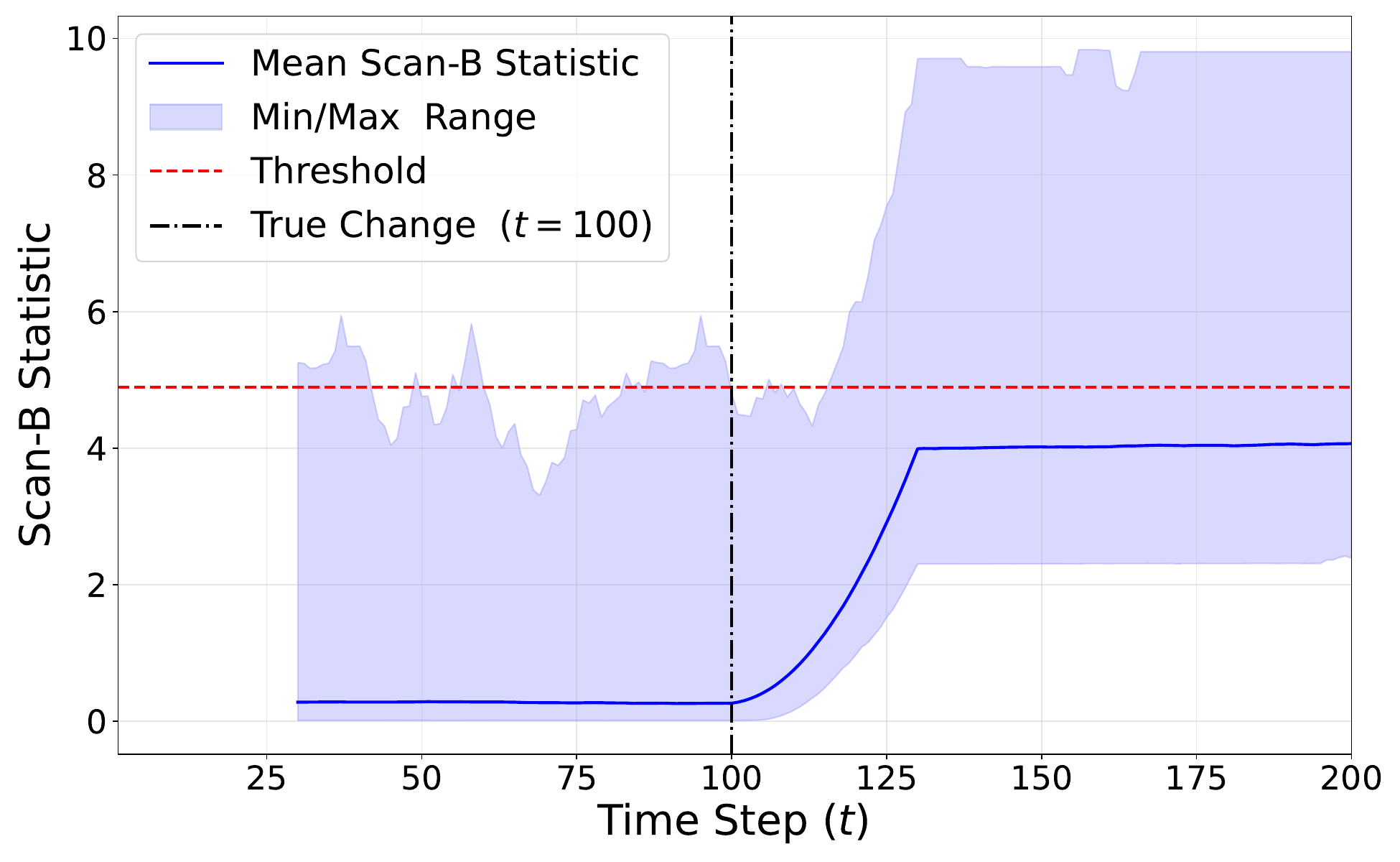}
        \caption{$B_0 = 30$}
        \label{fig:scan_b_net_30}
    \end{subfigure}
    \caption{Visualization of the Scan-B test statistic applied to AE reconstruction errors on the CIC-IDS-2018 dataset.}
\label{fig:scan_b_net_sensitivity}
\end{figure}

In contrast, Fig. \ref{fig:nn_cpd_sensitivity_network} plots the  $k$-NN test statistic of \cite{kNNCPD} applied to the identical AE reconstruction errors across various sliding window ($L$) and neighbor ($k$) configurations. The statistic exhibits extreme variance both pre-and post-change, accompanied by only a subtle and transient mean shift following $\tau^*$. Consequently, calibrating an effective decision boundary becomes practically infeasible; the  stopping rule proposed by \cite{kNNCPD} defaults to an excessively low threshold, triggering early terminations. Similarly, Fig. \ref{fig:scan_b_net_sensitivity} illustrates the Scan-B test statistic across varying reference block sizes ($B_0$). Although a marginal mean shift occurs post-change, it is entirely eclipsed by the statistic's extreme volatility. As a result, the standard Scan-B stopping rule computes an overly conservative threshold, completely failing to detect the anomaly.

\begin{figure}[htbp]
    \centering
    \begin{subfigure}[b]{0.48\textwidth}
        \centering
        \includegraphics[width=\textwidth]{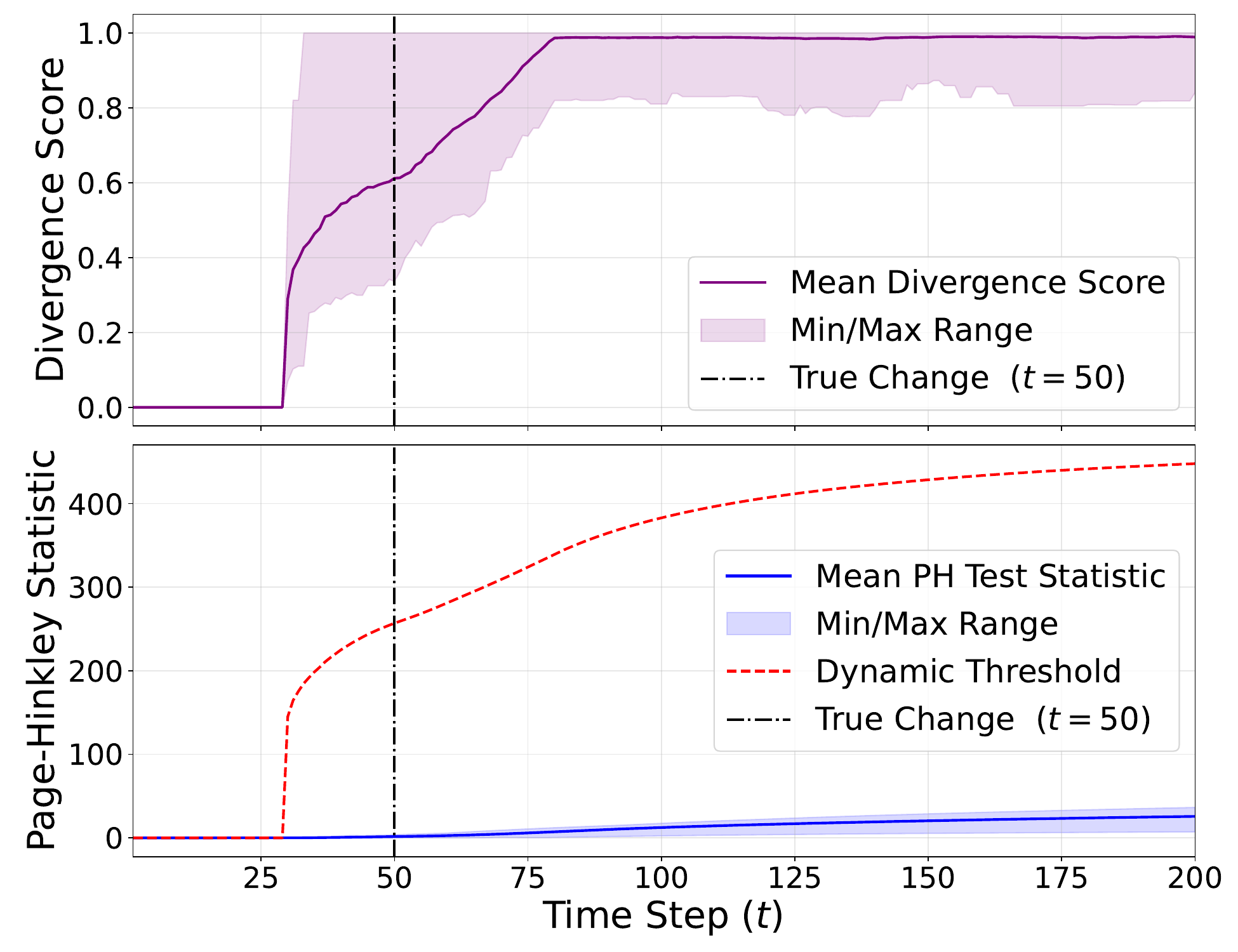}
        \caption{PCA-based Detector (KDD15)}
 \label{fig:nn_cpd_KDD}    \end{subfigure}
    \hfill
    \begin{subfigure}[b]{0.48\textwidth}
        \centering
        \includegraphics[width=\textwidth]{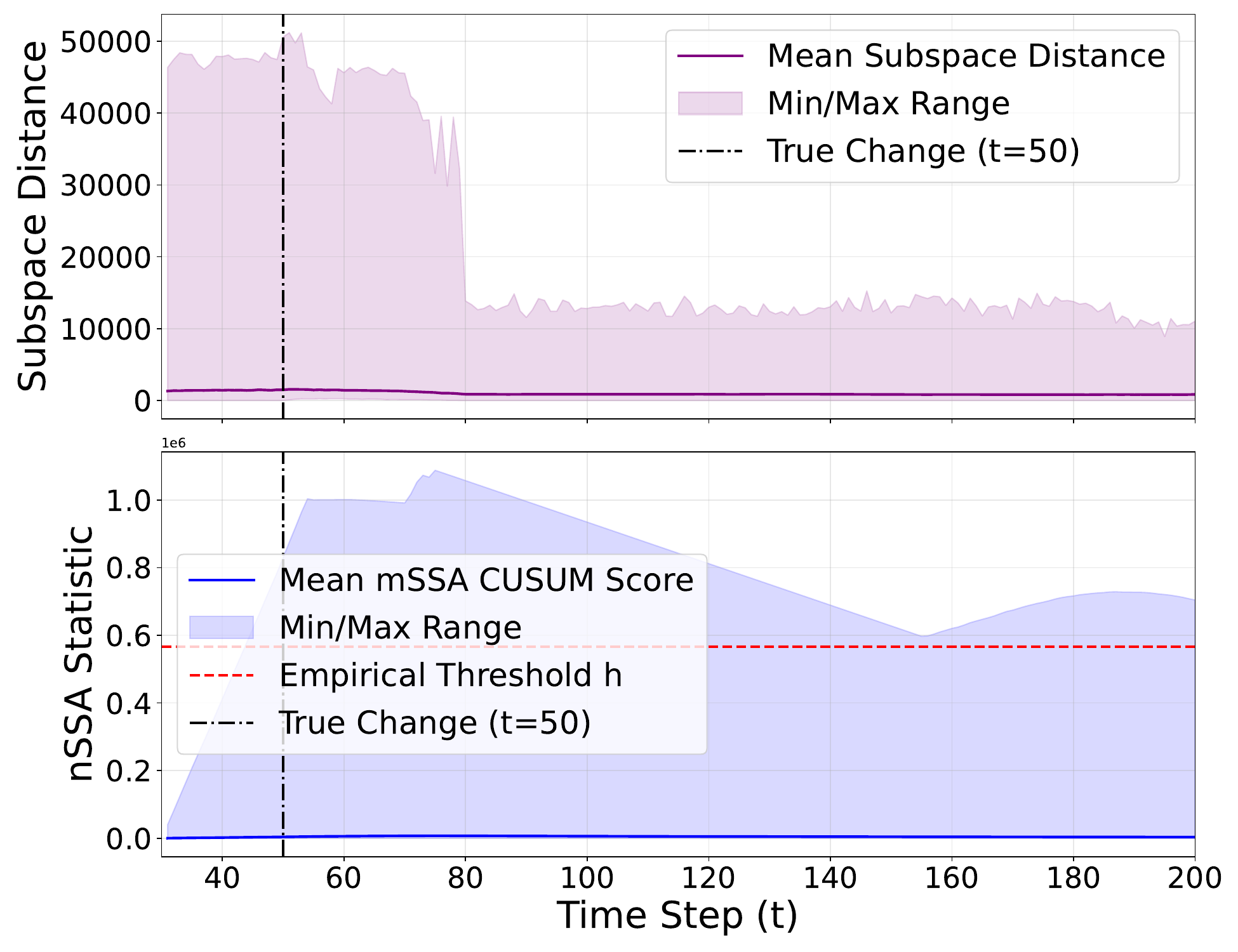}
        \caption{mSSA Detector}
    \label{fig:mssa_network}
    \end{subfigure}
    \caption{Visualization of the linear subspace test statistics applied to the CIC-IDS-2018 dataset.}
\label{fig:linear_subspace_network_failures}
\end{figure}

Furthermore, we evaluate the linear subspace-based methods of \cite{PCA_KDD_15} and \cite{msSA}. As reported in Fig. \ref{fig:nn_cpd_KDD}, the principal component divergence score from \cite{PCA_KDD_15} begins to rapidly inflate even prior to the true change point, while its  heuristic PH threshold remains too loose to trigger a valid detection. Likewise, the mSSA statistic (Fig. \ref{fig:mssa_network}) suffers from extreme variance, with the tracked subspace distances showing no discernible separation between the two regimes. 

\subsection{Chaotic Systems}
Finally, we  evaluate CUR-CUSUM on chaotic time-series trajectories. During online evaluation, each   trajectory with maximum horizon $T_{max}=200$ initiates under  Lorenz attractor dynamics before undergoing a  transition into the   Chen attractor at $\tau^*=100$. The same LSTM base model from the Lorenz-Chen SHT experiments of the main body is considered, trained only on sequences from the Lorenz (pre-change) attractor.

\begin{figure}[htbp]
    \centering
    \begin{subfigure}[b]{0.48\textwidth}
        \centering
        \includegraphics[width=\textwidth]{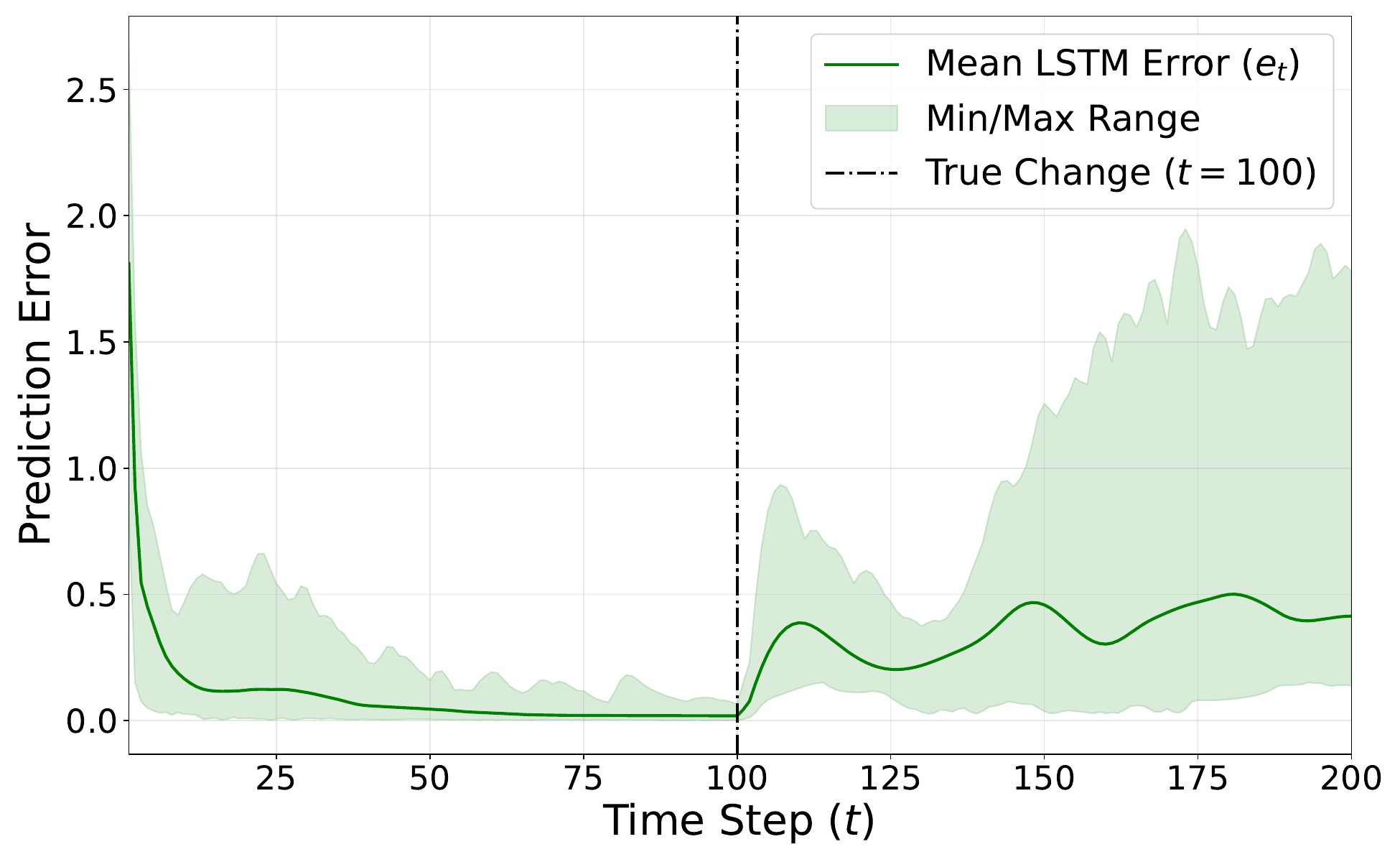}
        \caption{LSTM Prediction Errors.}
        \label{fig:chaos_lstm_errors}
    \end{subfigure}
    \hfill
    \begin{subfigure}[b]{0.48\textwidth}
        \centering
        \includegraphics[width=\textwidth]{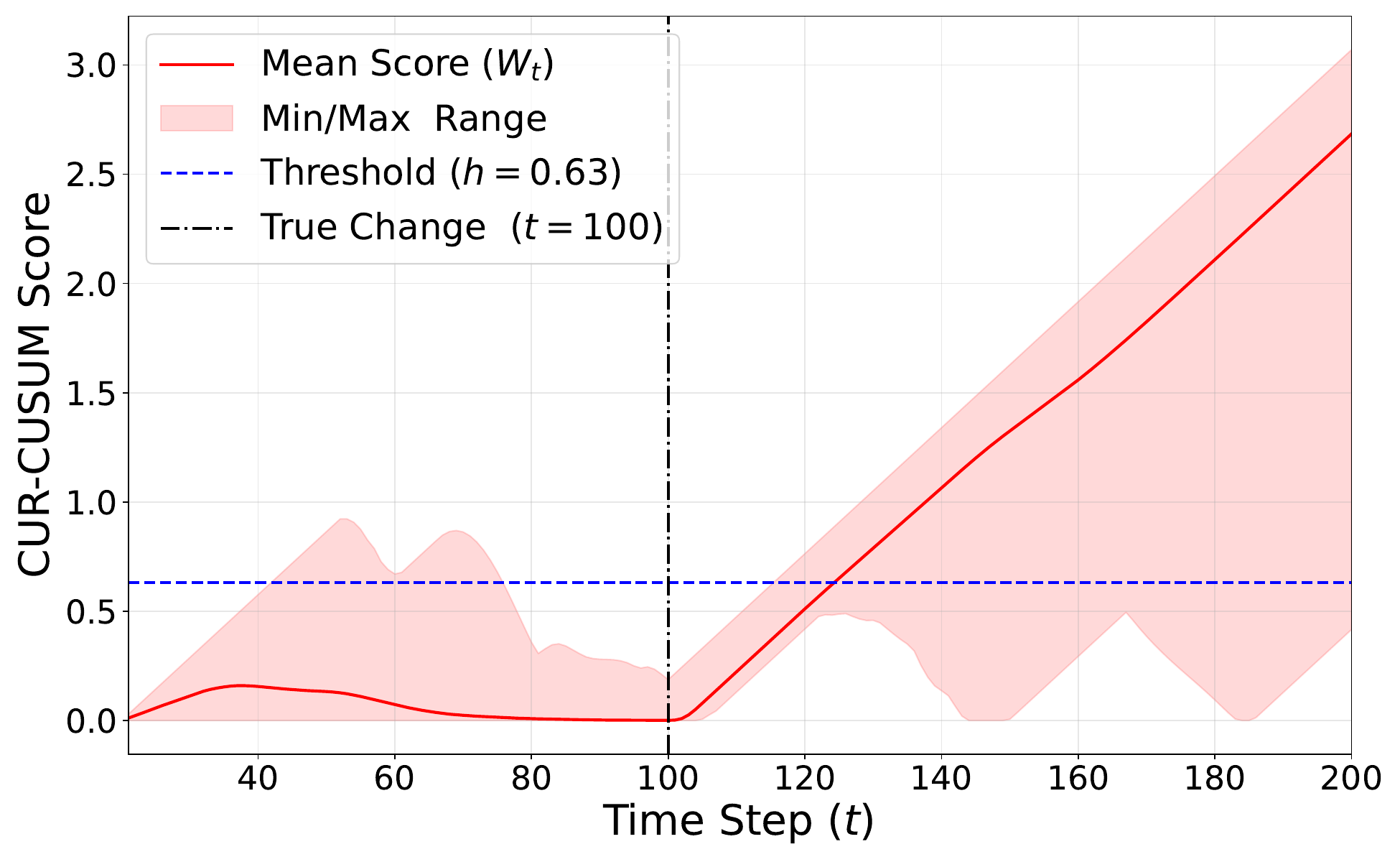}
        \caption{Proposed Test Statistic.}
        \label{fig:chaos_rcu_cusum}
    \end{subfigure}
    \caption{Trajectory comparison of the LSTM error and the derived test statistic.}
\label{fig:side_by_side_chaos_CPD}
\end{figure}

The proposed CUR-CUSUM detector yielded an early termination rate of $2.5\%$, zero missed detections, and an average detection delay of $24.52$ steps. Diagnostics are illustrated in Fig. \ref{fig:side_by_side_chaos_CPD}. While the predictive LSTM errors display only a moderate mean increase following the change, the transition induces a substantial variance inflation. CUR-CUSUM reliably captures this higher-order distributional shift, driving a steady, linear divergence in the test statistic immediately following the change point (Figure \ref{fig:chaos_rcu_cusum}).

\begin{figure}[H]
    \centering
    
    \begin{subfigure}[b]{0.48\textwidth}
        \centering
        \includegraphics[width=\textwidth]{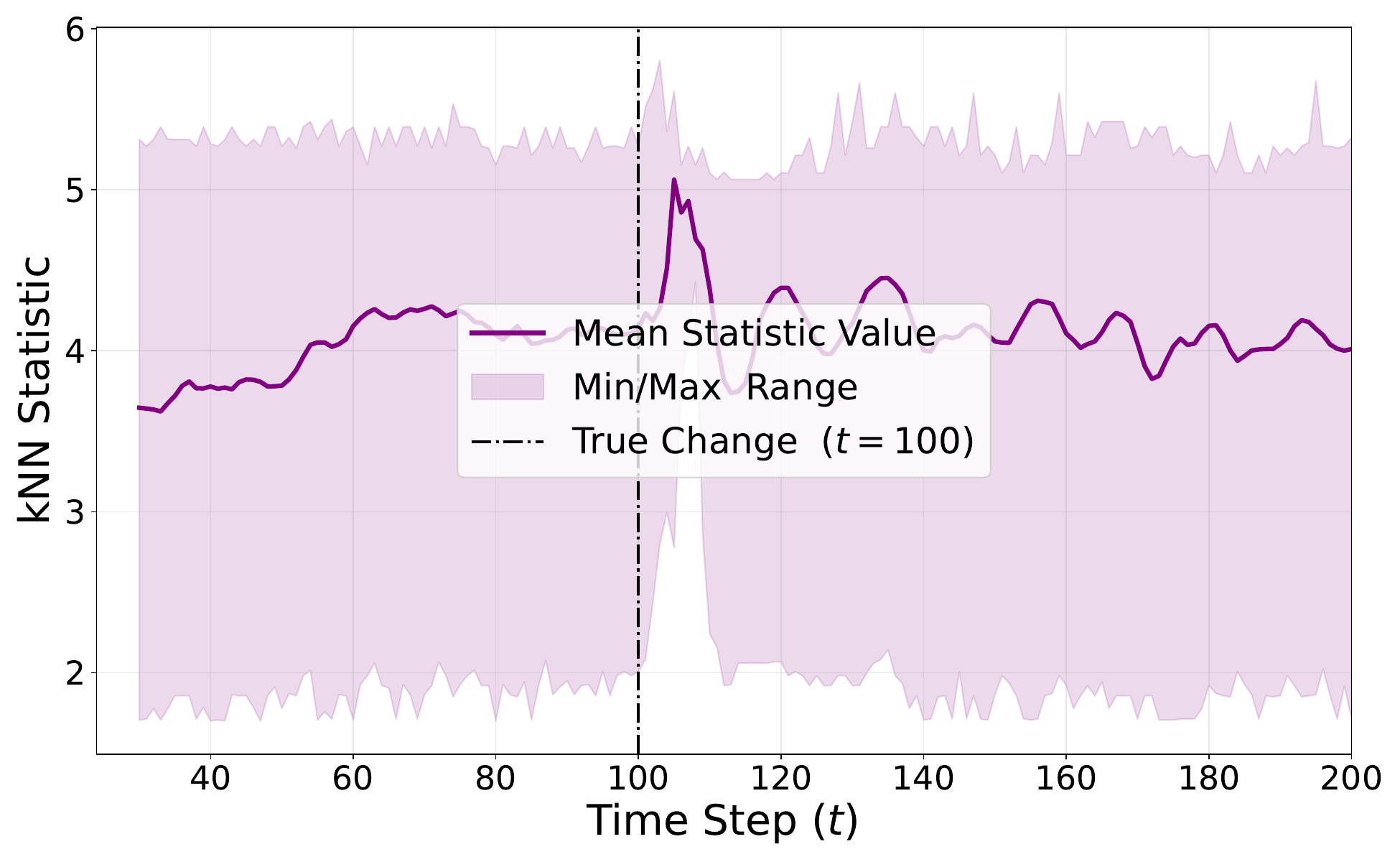}
        \caption{$L=10, k=3$}
        \label{fig:chaos_nn_cpd_10_k3}
    \end{subfigure}
    \hfill
    \begin{subfigure}[b]{0.48\textwidth}
        \centering
        \includegraphics[width=\textwidth]{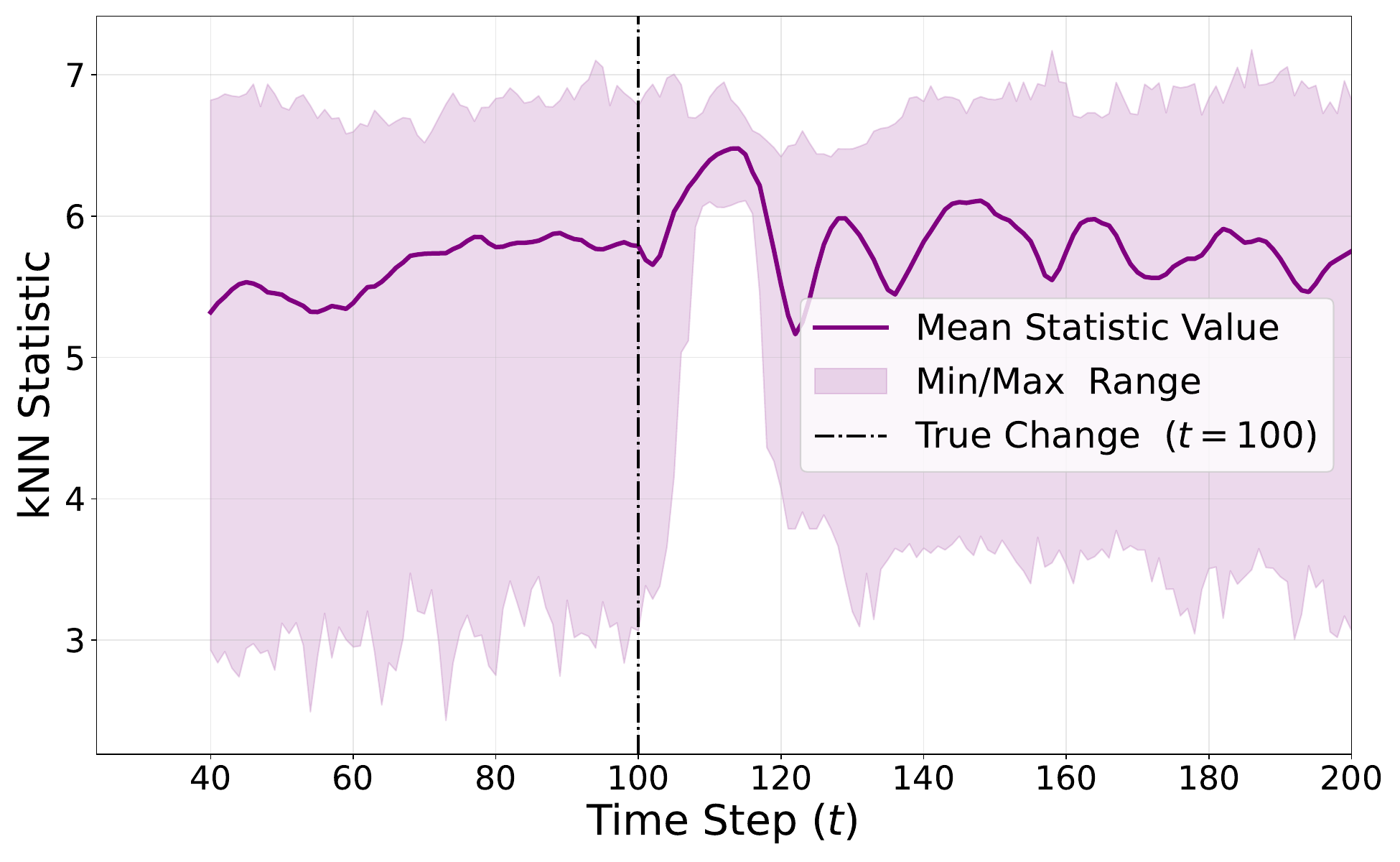}
        \caption{$L=20, k=3$}
        \label{fig:chaos_nn_cpd_20_k3}
    \end{subfigure}
    
    \vspace{0.4cm} 
    
    \begin{subfigure}[b]{0.48\textwidth}
        \centering
        \includegraphics[width=\textwidth]{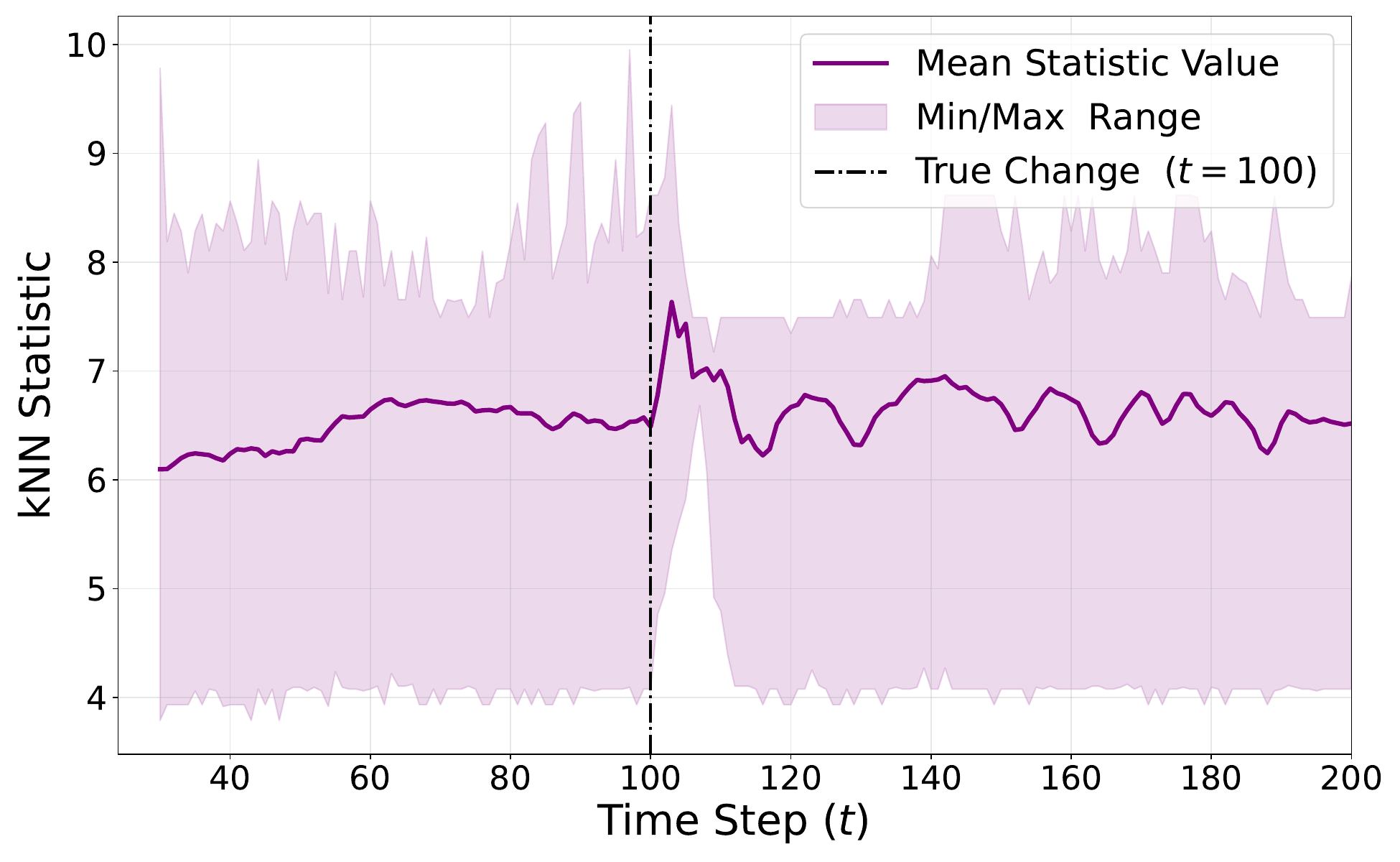}
        \caption{$L=10, k=5$}
        \label{fig:chaos_nn_cpd_10_k5}
    \end{subfigure}
    \hfill
    \begin{subfigure}[b]{0.48\textwidth}
        \centering
        \includegraphics[width=\textwidth]{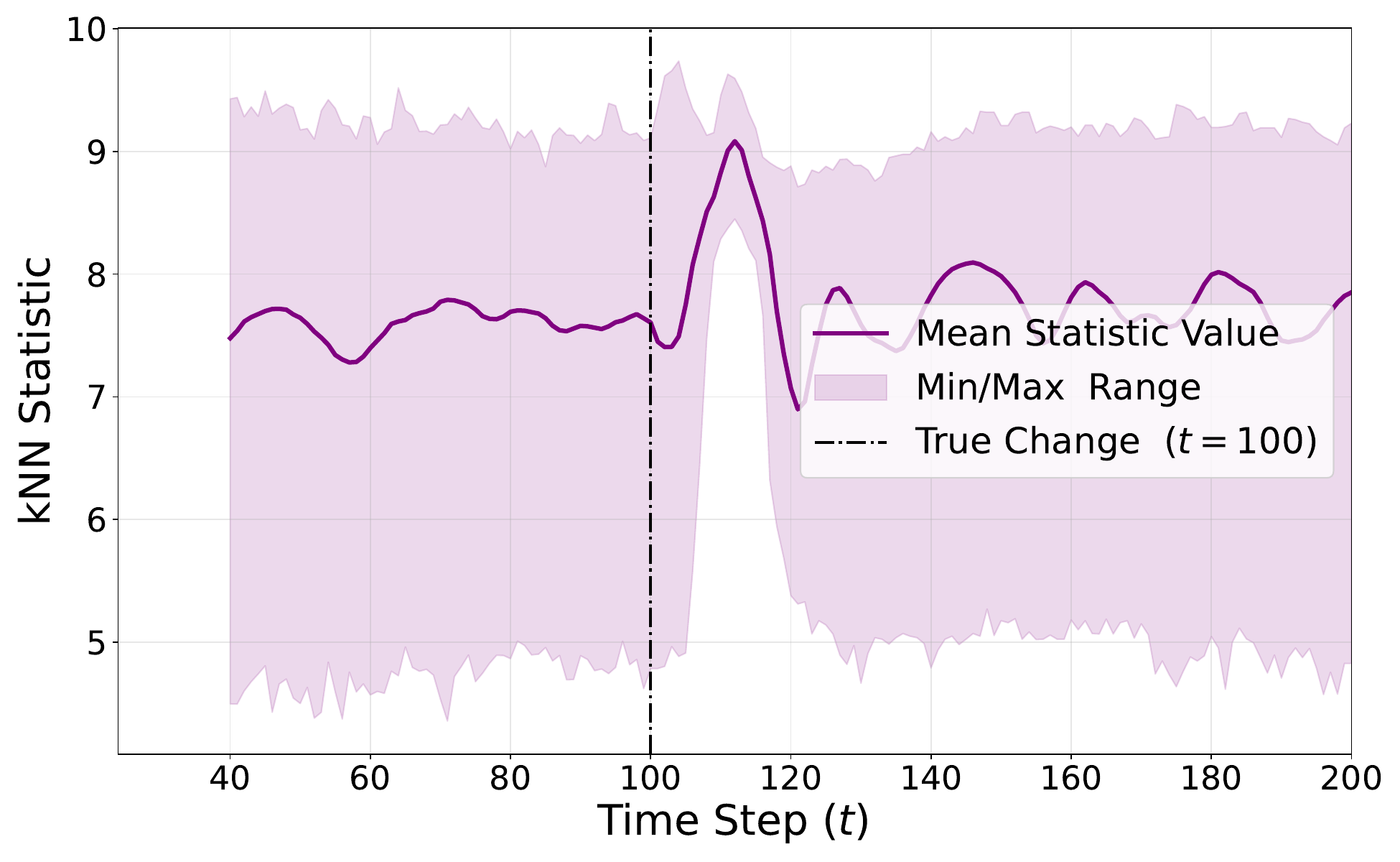}
        \caption{$L=20, k=5$}
        \label{fig:chaos_nn_cpd_20_k5}
    \end{subfigure}
    
    \caption{Visualization of the kNN test statistic applied to LSTM  errors on  chaotic change detection.} 
\label{fig:nn_cpd_sensitivity_chaos}
\end{figure}

\begin{figure}[htbp]
    \centering
    \begin{subfigure}[b]{0.32\textwidth}
        \centering
        \includegraphics[width=\textwidth]{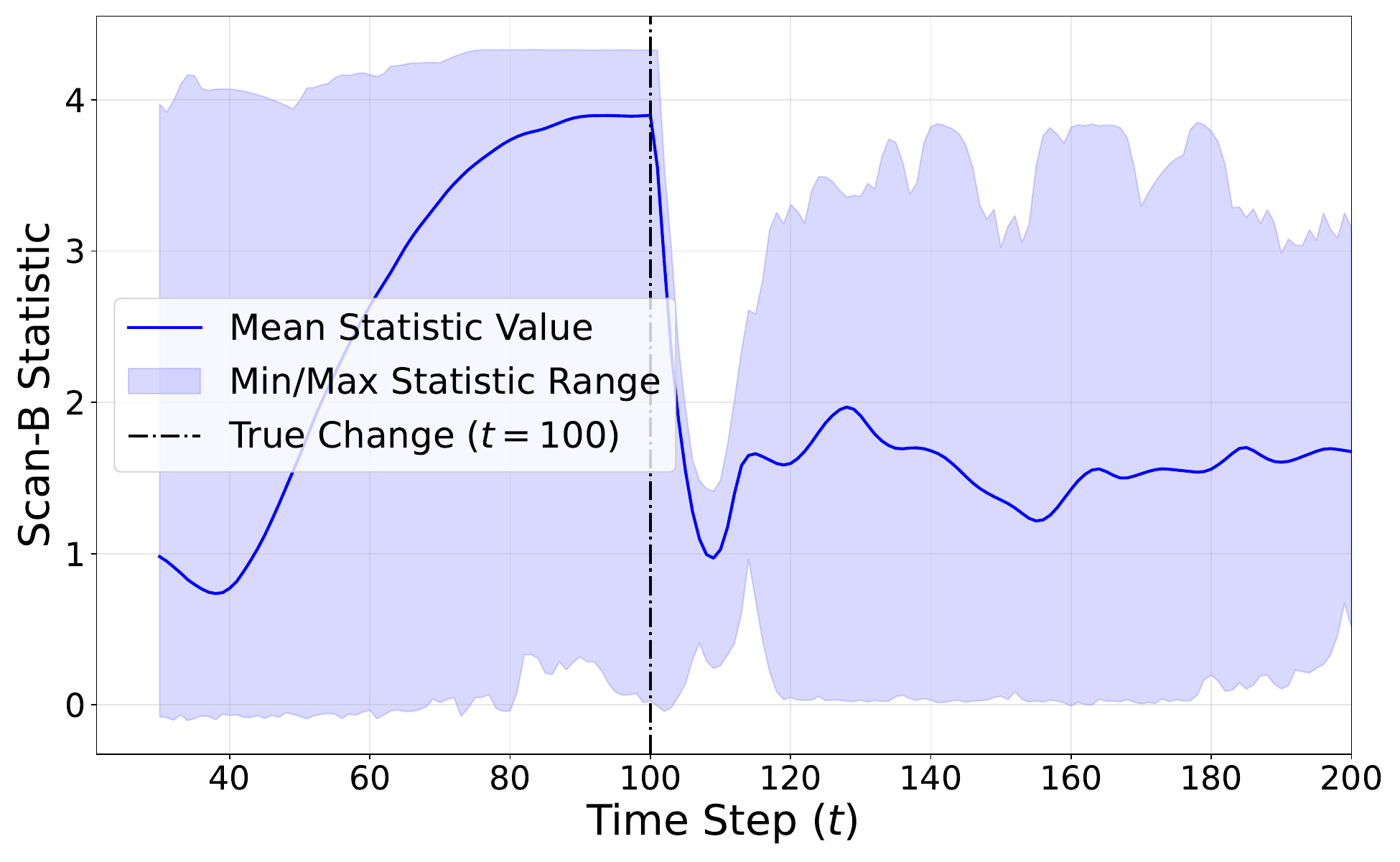}
        \caption{$B_0=10$}
        \label{fig:scanb_chaos_10}
    \end{subfigure}
    \hfill
    \begin{subfigure}[b]{0.32\textwidth}
        \centering
        \includegraphics[width=\textwidth]{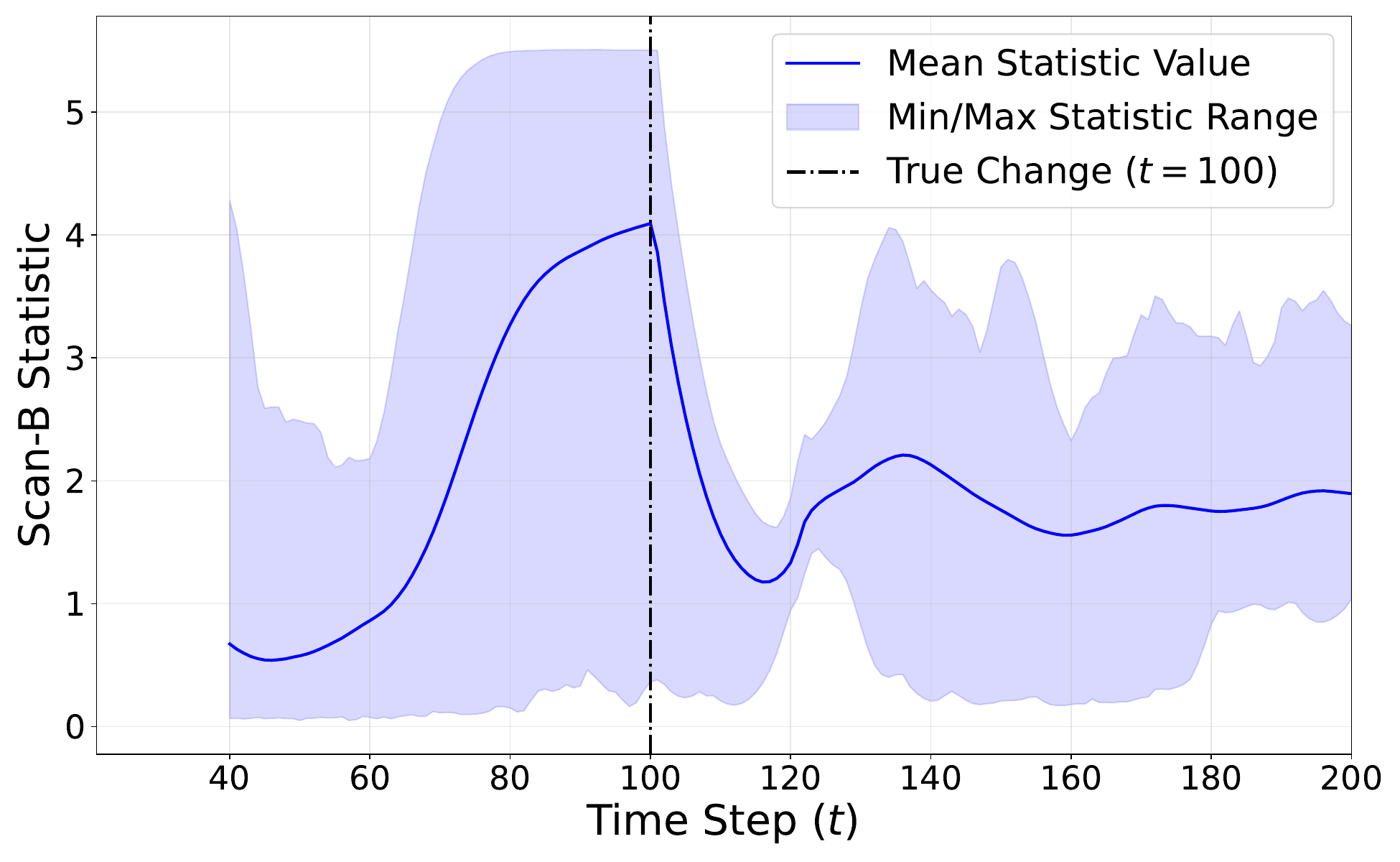}
        \caption{$B_0=20$}
        \label{fig:scanb_chaos_20}
    \end{subfigure}
    \hfill
    \begin{subfigure}[b]{0.32\textwidth}
        \centering
        \includegraphics[width=\textwidth]{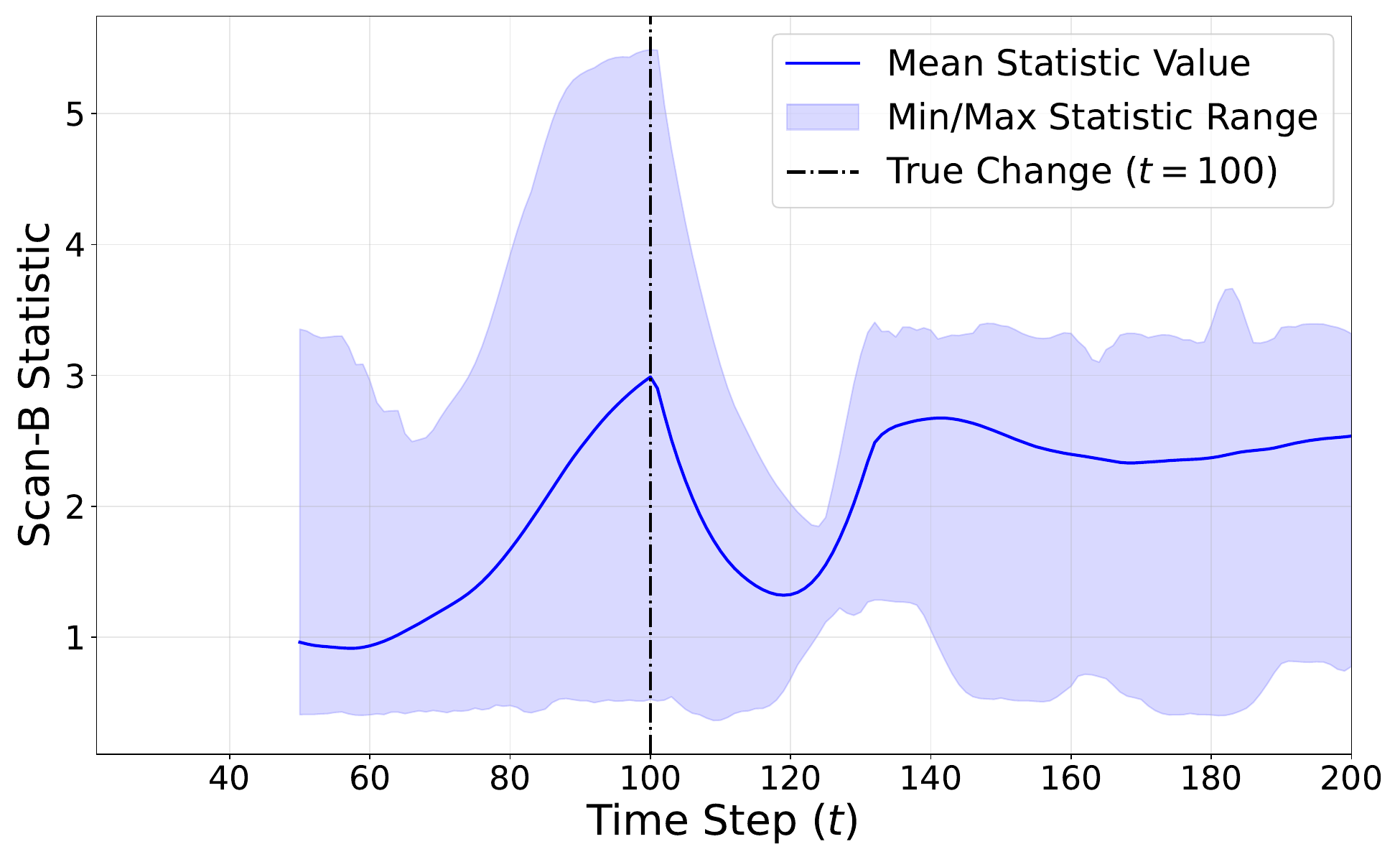}
        \caption{$B_0=30$}
        \label{fig:scanb_chaos_30}
    \end{subfigure}
    \caption{Visualization of the Scan-B test statistic applied to LSTM errors on chaotic change detection.}
    \label{fig:scanb_chaos_sensitivity}
\end{figure}

\begin{figure}[htbp]
    \centering
    \begin{subfigure}[b]{0.48\textwidth}
        \centering
        \includegraphics[width=\textwidth]{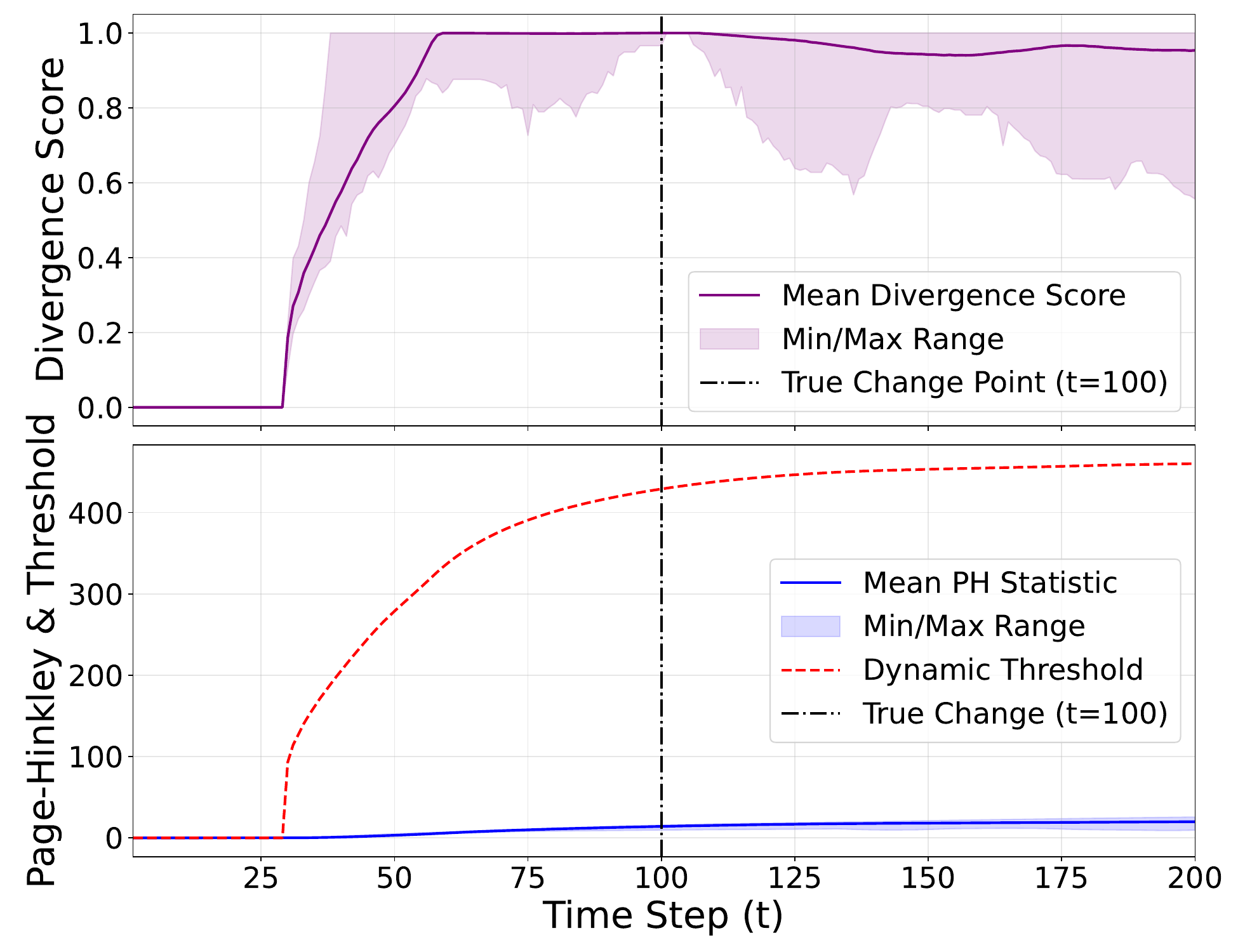}
        \caption{PCA-based Detector}
        \label{fig:pca_kdd15_chaos}
    \end{subfigure}
    \hfill
    \begin{subfigure}[b]{0.48\textwidth}
        \centering
        \includegraphics[width=\textwidth]{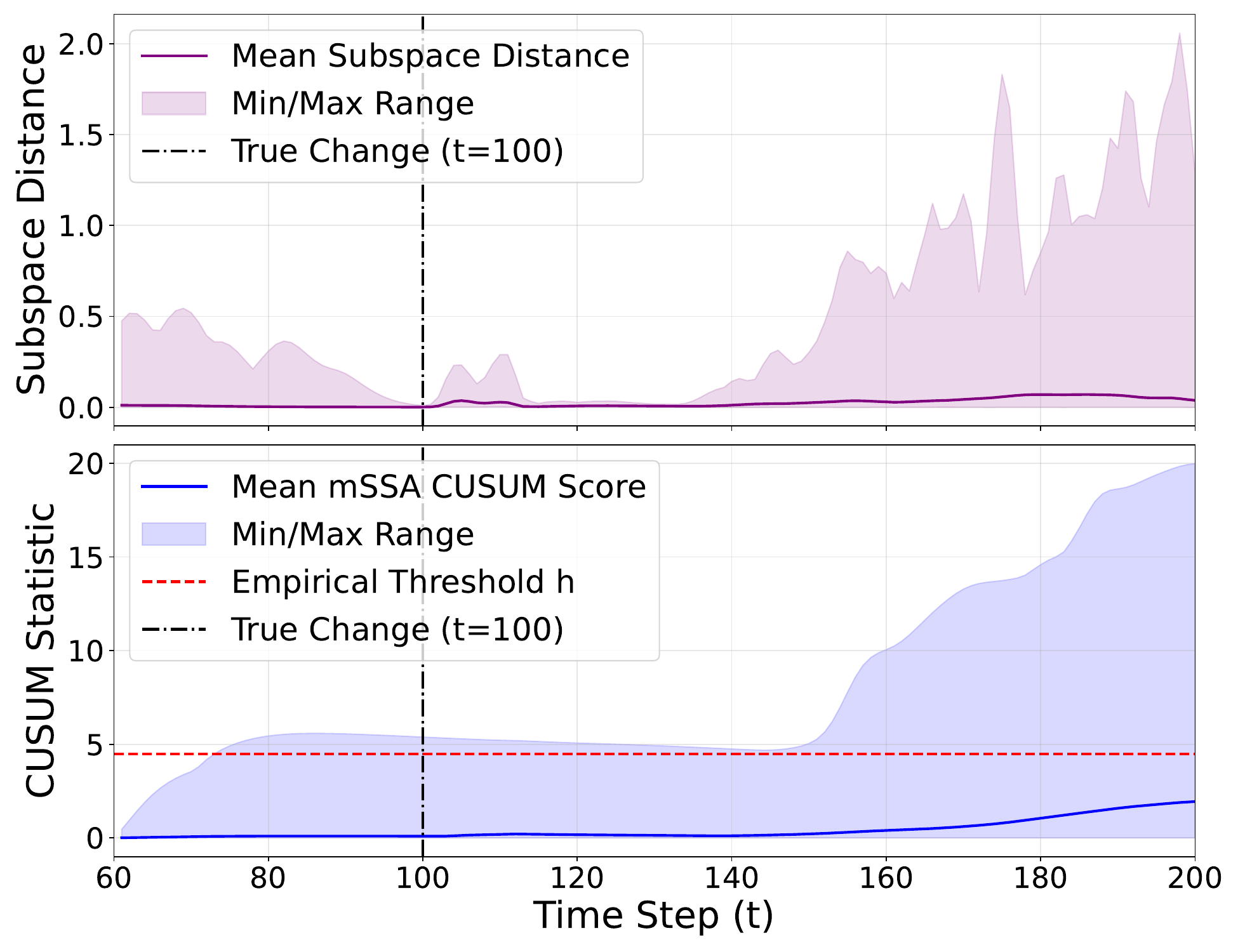}
        \caption{mSSA Detector}
        \label{fig:mssa_chaos}
    \end{subfigure}
    \caption{Visualization of the linear subspace baselines on chaotic change detection.}
    \label{fig:linear_subspace_chaos_failures}
\end{figure}

Both k-NN and Scan-B test statistics are highly ineffective for this task, as depicted in Fig. \ref{fig:nn_cpd_sensitivity_chaos}Fig. and \ref{fig:scanb_chaos_sensitivity} respectively.  For the former, there is a minimal and transient mean shift after the change, which is again masked by the variance. Interestingly, the Scan-B statistic climbs to a large value pre-change and decreases after change rendering detection infeasible.  
The behavior of the linear subspace baselines mirrors the behavior observed in the network traffic evaluation, as illustrated in Fig. \ref{fig:linear_subspace_chaos_failures}. The principle component detection method generates an excessively loose, rigid  threshold, while its divergence score prematurely surges well before the true transition at $\tau^*=100$. Meanwhile, the mSSA test statistic exhibits extreme variance and registers negligible increase post-change. Even this marginal mean increase is delayed until after $t=150$ (over $50$ steps past the change point), leaving the cumulative statistic suppressed far below the decision boundary. 

\newpage
\section{Additional CPD References}\label{sec:omitted_baselines}

In selecting representative benchmarks for our experimental evaluation, we focused on state-of-the-art methods spanning distinct theoretical paradigms: non-parametric kernel tests (Scan-B~\cite{ScanB}), graph-based  tests ($k$-NN~\cite{kNNCPD}), and linear subspace tracking (component-wise PCA \cite{PCA_KDD_15} and mSSA \cite{msSA}). 

While other unsupervised online detectors exist in the data stream literature, they were excluded from the primary empirical study due to documented systemic weaknesses:
\begin{itemize}
    \item \textbf{Spatial Partitioning (\textsf{kdq-tree}) \cite{dasu2006information}:} A windowed based approach that compares the distance of the current window and a reference pre-change window.  As established in \cite{PCA_KDD_15}, \textsf{kdq-tree} suffers acutely from the curse of dimensionality; in moderate-to-high dimensions ($d \ge 10$),  catastrophic miss rates are achieved.
    \item \textbf{Minor-Component PCA (\textsf{PCA-SPLL}) \cite{kuncheva2014pca}:} Discards leading components and projects observations exclusively onto principal components with the smallest eigenvalues, evaluating a Semi-Parametric Log-Likelihood (SPLL) criterion. While motivated by the hypothesis that minor axes are more sensitive to arbitrary perturbations, Qahtan et al.~\cite{PCA_KDD_15} demonstrated that discarding major components frequently eliminates the primary directions of variance and mean shifts. Consequently, \textsf{PCA-SPLL} suffers from elevated miss rates on high-dimensional data.
   \item \textbf{Direct Density-Ratio Estimation (\textsf{KLIEP / RuLSIF}) \cite{kawahara2009change, liu2013change}:} Estimates probability density ratios directly between reference and test windows via kernel models. However, repeatedly optimizing kernel weights and performing cross-validation at each time step incurs heavy computational complexity, forcing practitioners to restrict themselves to small sliding windows. Furthermore, these methods offer no theoretical ARL guarantees, leaving key parameters to be hand-tuned through empirical trial and error.
\end{itemize}
 Methods \cite{dasu2006information,kuncheva2014pca,kawahara2009change,liu2013change} are compared with component-wise PCA in \cite{PCA_KDD_15} with the latter emerging as a superior detector. Hence, it is the one considered in our experiments.

Another recent distribution-free work is the B-CAVE framework~\cite{B-Cave}. While computationally straightforward for small sliding windows, B-CAVE exhibits severe structural and statistical limitations that restrict its general applicability. First, the algorithm is fundamentally confined to one-dimensional streams: its core mechanism relies on sorting scalar observations within a centered window of length $W$ to evaluate between-class averages and variances, an operation that lacks a natural, canonical ordering in high-dimensional spaces ($d \gg 1$) without ad-hoc dimensionality reduction. 
Furthermore, B-CAVE is purely heuristic and lacks analytical distribution-free guarantees; it provides neither non-asymptotic bounds on the false alarm probability nor theoretical ARL formulas, forcing practitioners to rely on empirical trial-and-error to tune threshold parameters. Finally, because the test statistic is driven by scalar amplitude partitioning, the method exhibits significant vulnerability to structural shifts that preserve the marginal amplitude profile, such as frequency or phase modulations.

An alternative research thread examines SCPD with parametric uncertainty regarding the post change distribution.
The GLR CUSUM~\cite{GLR_CUSUM}  that combines the conventional CUSUM with the GLR principle achieves asymptotic optimality, albeit with unbounded computational costs. To alleviate this, the recently introduced $K$-Best rule~\cite{KBest} 
retains only the $K$ candidate change points with the highest test statistics,  maintaining zero asymptotic regret with a much reduced computational complexity. 
However, such approaches fundamentally rely on tractable parametric likelihoods and closed-form Maximum Likelihood Estimates (e.g., Gaussian shifts). 
They are  therefore inapplicable to our target setting—complex, high-dimensional streams governed by unknown nonlinear dynamics (such as deep representations, chaotic attractors, or network metadata) where parametric densities are unavailable.
\section{Ablation Analysis}
The key ideas of the proposed CUREB-SHT and CUR-CUSUM methods are 1)Use model errors instead of raw observations and 2)Use the specific test statistic. The latter has already been proven necessary by comparing Scan-B and k-NN on the same model's reconstruction errors in all SCDP experiments. We will now verify whether tracking the clipped residuals of a model is necessary or whether the model can be completely bypassed. 

In this ablation analysis focusing on network security (CIC-IDS-2018 experiments from section \ref{sec:main_Results}) we consider the raw variants of CUREB-SHT and CUR-CUSUM where only the threshold/stopping rule logic is implemented.
 Because the RCU-CUSUM detector operates on a one-dimensional scalar stream, we collapsed the raw multidimensional observation vector  using two classical distance metrics to serve as baseline inputs. The first baseline is the \textit{Standardized $L_2$ Norm}, defined as $s_t = ||\mathbf{x}_t||_2$, where $\mathbf{x}_t$ is the raw feature vector standardized using a standard scaler on the null/pre-change data. Our second ablation benchmark utilizes the \textit{Mahalanobis Distance}. Both 1D scalar streams were subsequently fed into the exact same CUREB-SHT and CUR-CUSUM detector formulations, including the identical   clipping logic and data-driven threshold calibrations. This experiment directly tests our hypothesis: that applying statistical tests directly to raw observational norms fails due to the retention of non-stationary, predictable system dynamics, whereas predictive residuals cleanly isolate unmodeled anomalies.
The results are reported in table \ref{tab:unified_ablation_appendix},  where the importance of the model utilization  instead of raw data is clearly highlighted. In both cases the raw tests can not detect the alternative/post-change regime.

\begin{table}[htbp]
    \centering
    \caption{Ablation Analysis Results.}
    \label{tab:unified_ablation_appendix}
    \begin{subtable}{\textwidth}
        \centering
        \caption{Sequential Hypothesis Testing (SHT)}
        \begin{tabular}{@{}lccc@{}}
            \toprule
            \textbf{Pipeline Configuration} & \textbf{FP Rate} & \textbf{Avg. Delay (steps)} & \textbf{ FN Rate } \\
            \midrule
            Raw $L_2$ Norm & 0 $\%$ & 18.00 & 99.80\% \\
            Raw Mahalanobis Distance & 0 \% & 98.65 & 81.40\% \\
         
            \bottomrule
        \end{tabular}
        \label{tab:ablation_sht}
    \end{subtable}
        \vspace{1em}

    \begin{subtable}{\textwidth}
        \centering
        \caption{Sequential Change Point Detection (SCPD)}
        \begin{tabular}{@{}lccc@{}}
            \toprule
            \textbf{Pipeline Configuration} & \textbf{Early Termination Rate} & \textbf{Avg. Delay (steps)} & \textbf{Missed Probability} \\
            \midrule
            Raw $L_2$ Norm & $0.67 \%$ & $51.79 $ & 99.46\% \\
            Raw Mahalanobis Distance & 0.67 \% & $61.69$ & 32.87\% \\
           
            \bottomrule
        \end{tabular}
        \label{tab:ablation_scpd}
    \end{subtable}


\end{table}
\section{Sensitivity Analysis}\label{ap:Sensitivity}
To empirically validate the robustness of the proposed framework to hyperparameter selection, we conduct a targeted sensitivity analysis using the high-dimensional CIC-IDS-2018 network traffic dataset. Throughout our primary evaluations, the framework employs a universal, alternative-agnostic calibration strategy: the clipping threshold $C$ is set to the 95th percentile of the nominal validation errors, and the target alternative margin is fixed at $\Delta_{\text{target}} = \mu C$, with $\mu=0.8$. In this section, we systematically evaluate the individual impact of these two core parameters on both SHT and SCP performance. We structure our analysis in two phases: first, we fix the target margin multiplier $\mu$ at 0.8 and vary the validation percentile used to define the clipping boundary $C$; second, we hold $C$ constant at the default 95th percentile and vary $\mu$ to test the framework's sensitivity to the expected anomaly magnitude.

\begin{figure}[H]
    \centering
    \begin{subfigure}{0.32\textwidth}
        \centering
        \includegraphics[width=\textwidth]{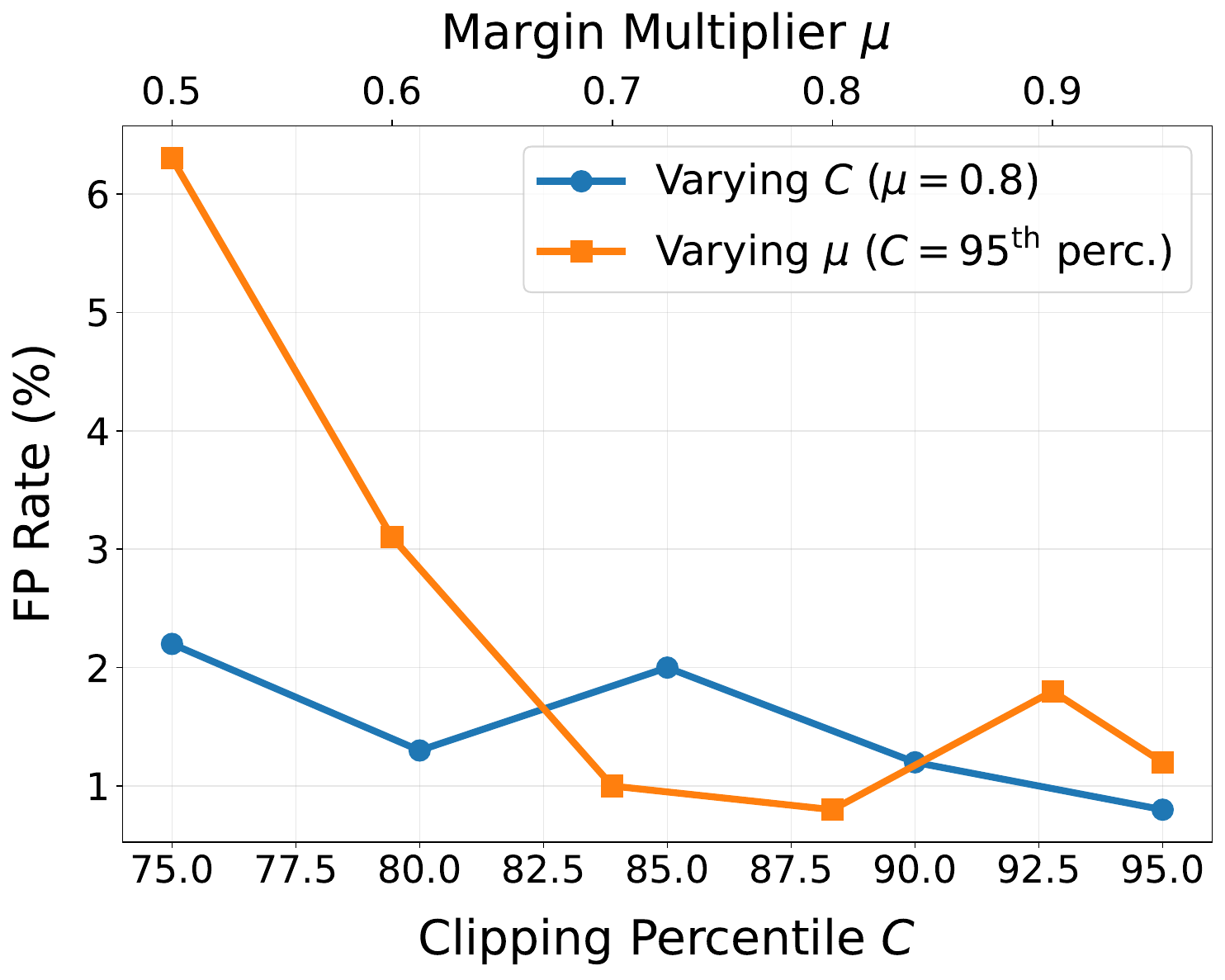}
        \caption{FP Rate}
    \end{subfigure}\hfill
    \begin{subfigure}{0.32\textwidth}
        \centering
        \includegraphics[width=\textwidth]{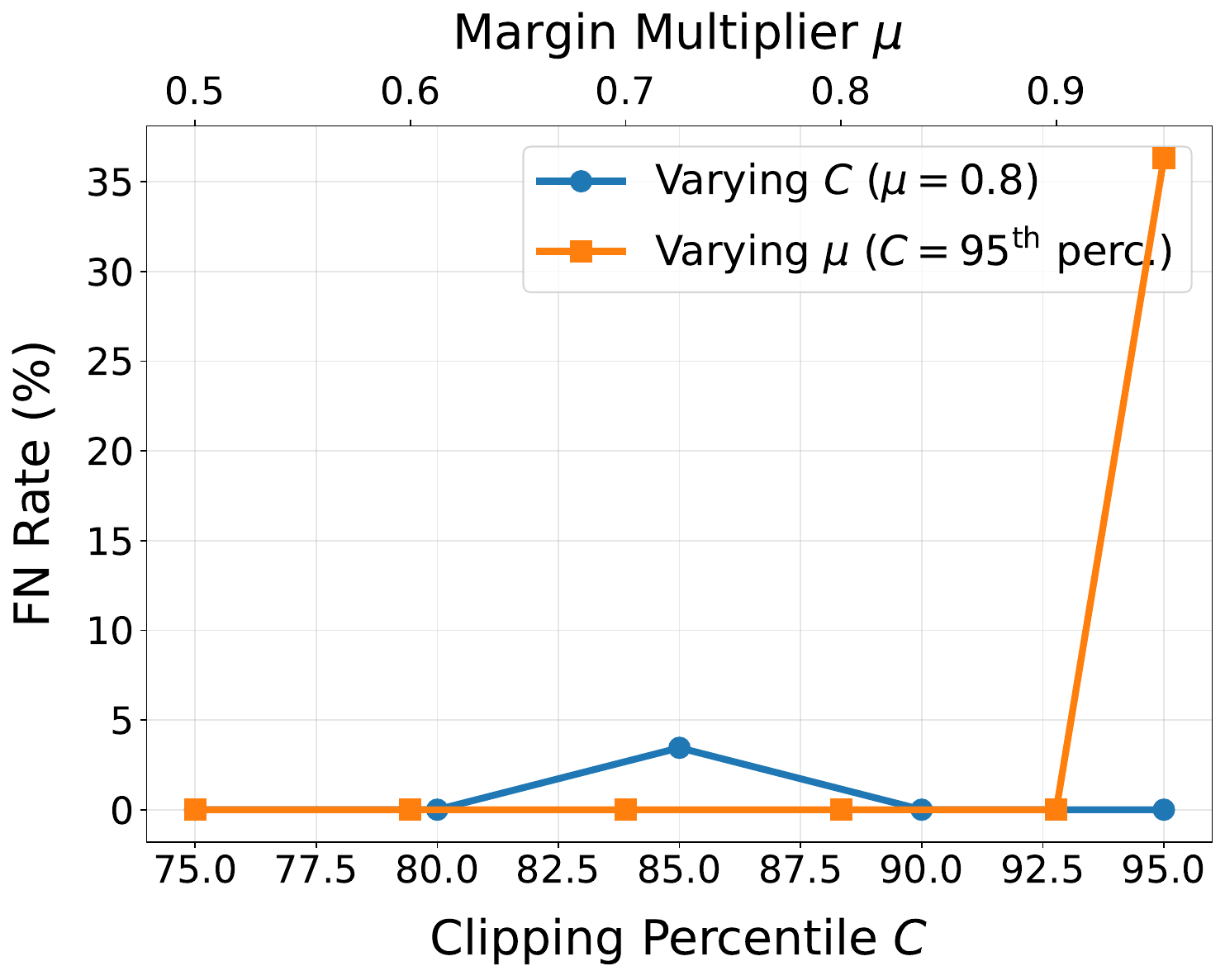}
        \caption{FN Rate}
    \end{subfigure}\hfill
    \begin{subfigure}{0.32\textwidth}
        \centering
        \includegraphics[width=\textwidth]{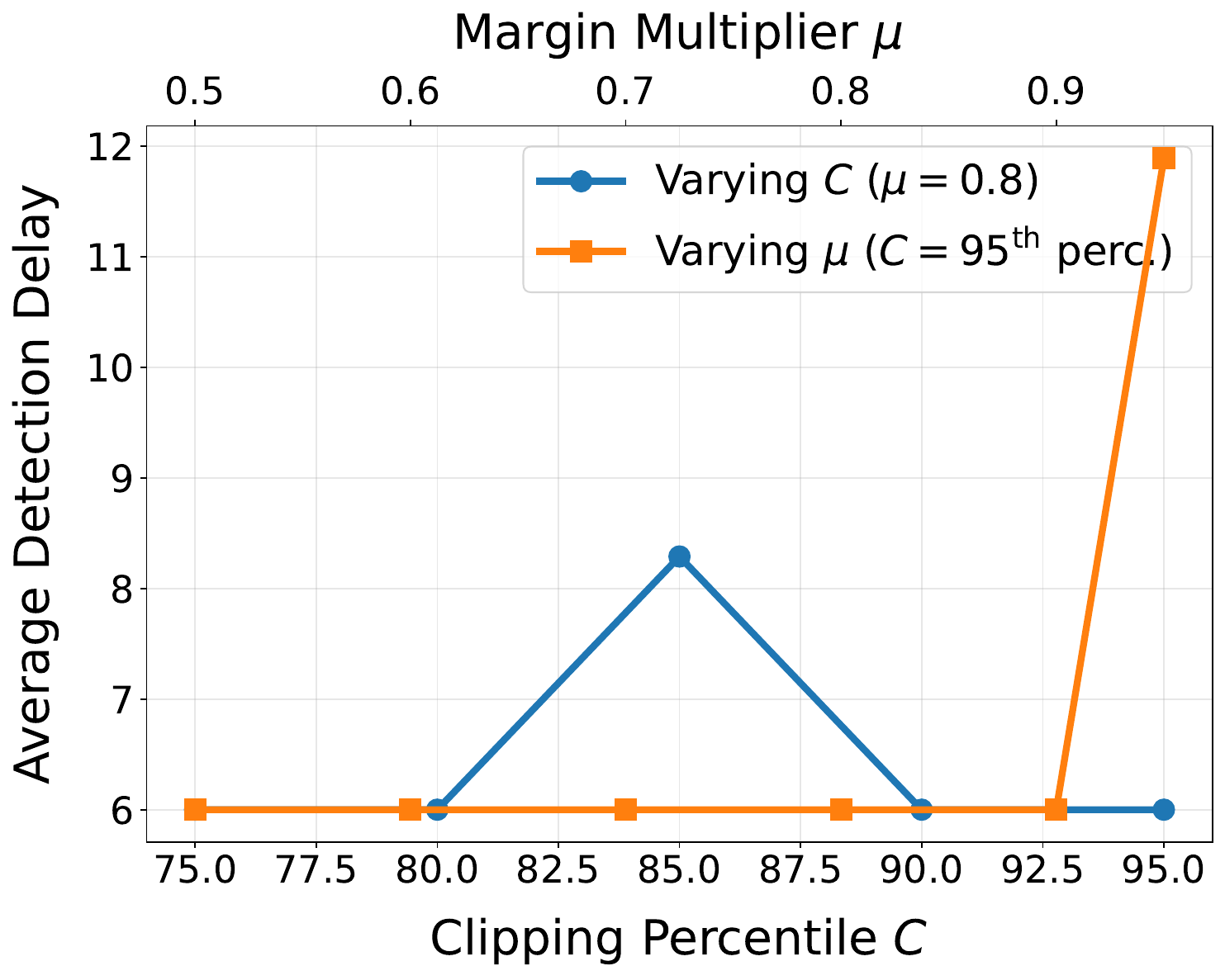}
        \caption{Detection Delay}
    \end{subfigure}
    \caption{Sensitivity of CUREB-SHT performance to hyperparameter variations on the CIC-IDS-2018 dataset. The clipping percentile $C$ is varied while fixing $\mu = 0.8$, and the margin multiplier $\mu$ is varied while fixing $C$ at the $95$th percentile.}
    \label{fig:sensitivity_sht}
\end{figure}

\begin{figure}[H]
    \centering
    \begin{subfigure}{0.32\textwidth}
        \centering
        \includegraphics[width=\textwidth]{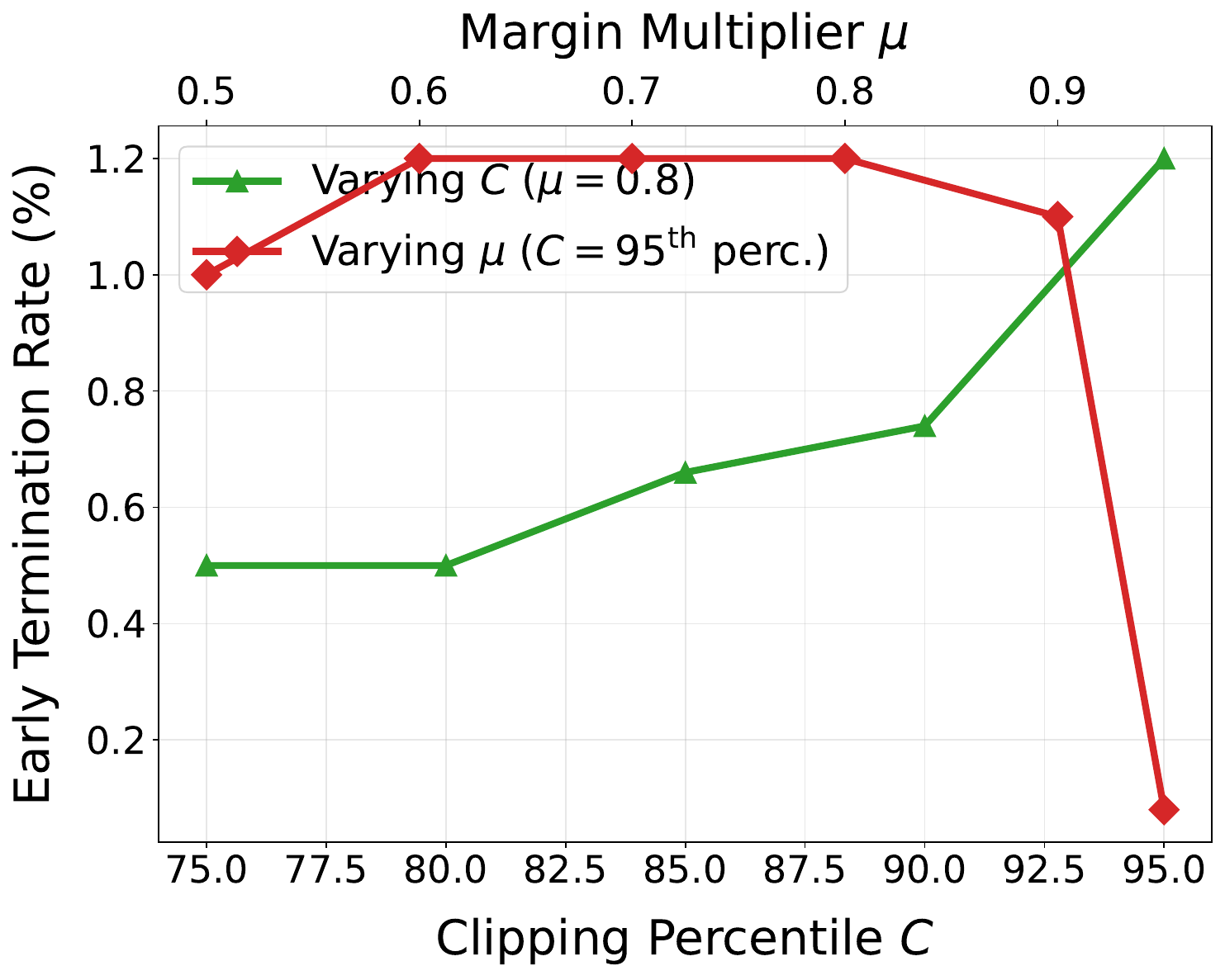}
        \caption{Early Termination Rate}
    \end{subfigure}\hfill
    \begin{subfigure}{0.32\textwidth}
        \centering
        \includegraphics[width=\textwidth]{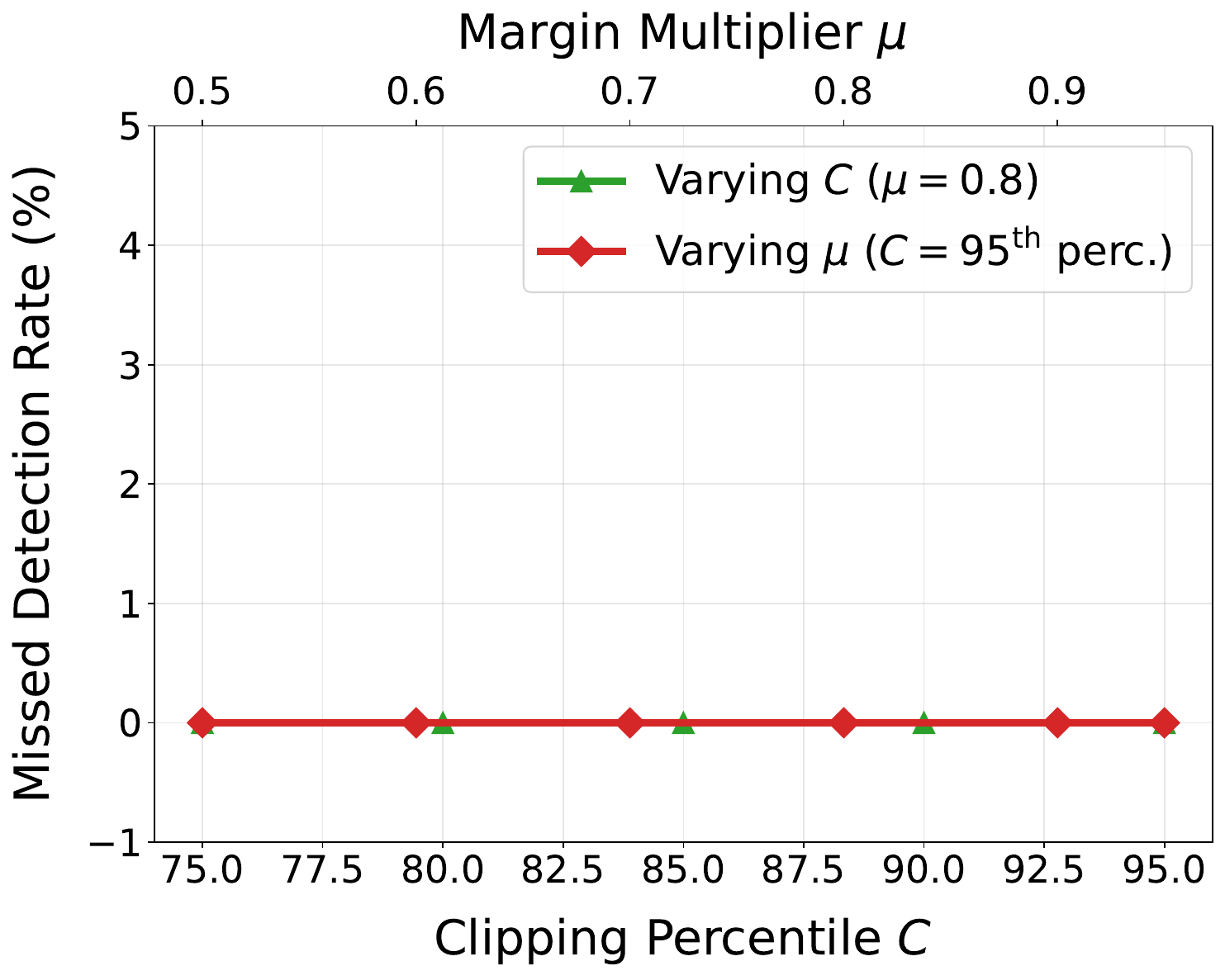}
        \caption{Missed Detection Rate}
    \end{subfigure}\hfill
    \begin{subfigure}{0.32\textwidth}
        \centering
        \includegraphics[width=\textwidth]{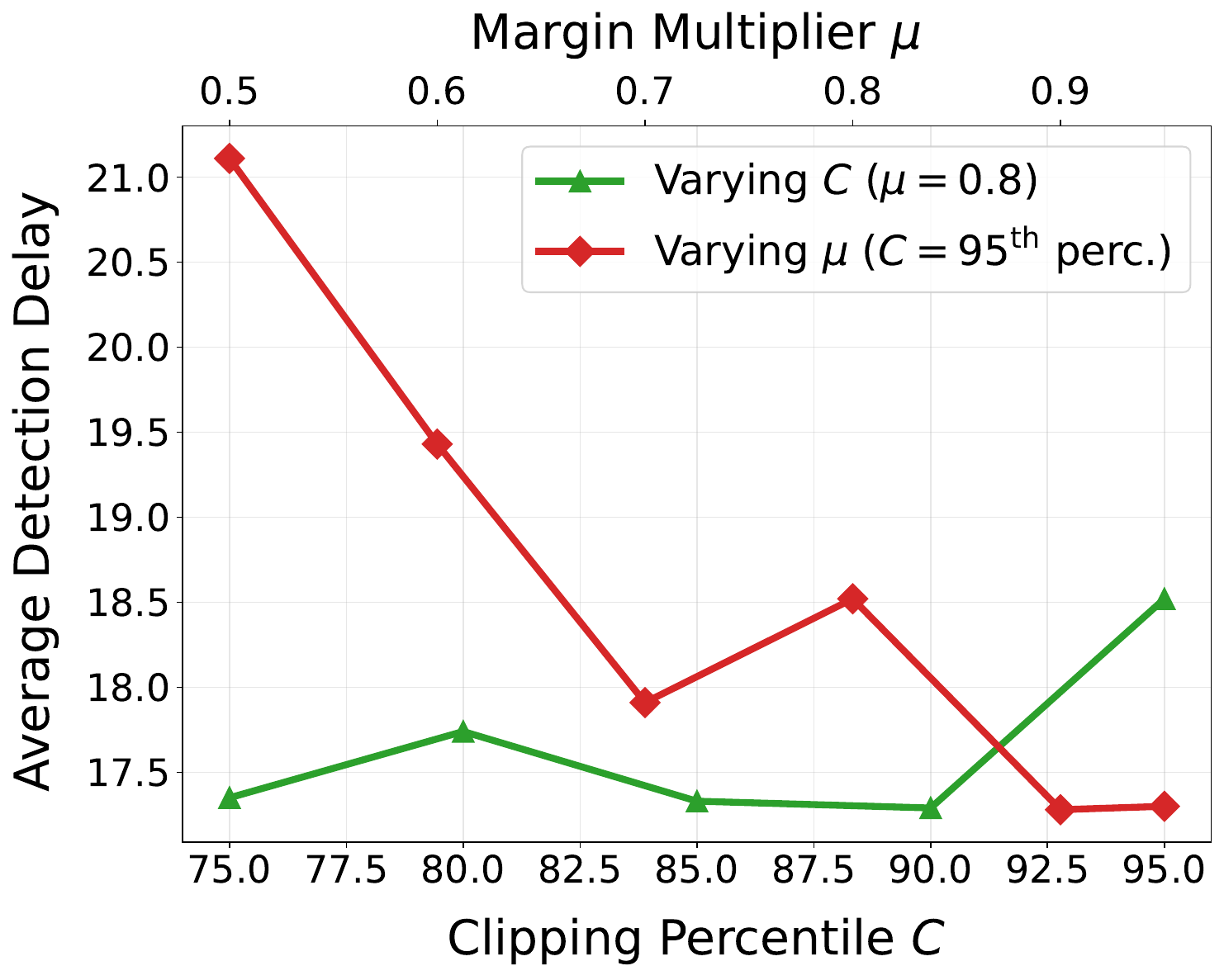}
        \caption{Detection Delay}
    \end{subfigure}
    \caption{Sensitivity of CUR-CUSUM performance to hyperparameter variations on the CIC-IDS-2018 dataset. The clipping percentile $C$ is varied while fixing $\mu = 0.8$, and the margin multiplier $\mu$ is varied while fixing $C$ at the $95$th percentile.}
    \label{fig:sensitivity_scpd}
\end{figure}

As shown in Fig. \ref{fig:sensitivity_sht}, CUREB-SHT maintains low error rates and keeps detection delays strictly between 6 and 8.5 steps, with FNs and delays only spiking when the margin multiplier reaches the extreme value of $\mu=0.95$. CUR-CUSUM proves to be even more stable (Fig. \ref{fig:sensitivity_scpd}), consistently recording zero misses and keeping early terminations below 1.2\% across all tested parameter combinations. Overall, these results confirm that both statistics are highly robust and do not require perfect hyperparameter tuning to operate reliably in the real world. Unless $\mu$ (and equivalently $\Delta_{\rm target}$) is set to a very large value, the detection errors remain small and the delays are slightly affected (less than $5$ steps average deviation).

\section{Reproducibility Details}\label{ap:reproducibility}
All neural models (LSTM/AE) are implemented using pytorch~\cite{pytorch}.  All experiments were conducted on a Linux (Ubuntu 20.04) desktop PC with an  NVIDIA GeForce RTX $3080$ GPU with $32$~GB of memory.
\subsection{Target Tracking Kinematics and Filter Specifications}
\label{app:tracking_kinematics}

\paragraph{2D Kinematic Model}
The state vector is $x_t = [p_{x,t}, p_{y,t}, v_{x,t}, v_{y,t}]^\top$ with sampling interval $\Delta t = 1\text{ sec}$.
Under $H_0$, the nominal state transition is linear with $x_{t+1} = F_{H_0} x_t + w_t$, where $w_t$ corresponds to AWGN with covariance matrix $Q_{2D}$ based on an acceleration noise model:
\begin{equation}
    F_{H_0} = \begin{bmatrix} 1 & 0 & \Delta t & 0 \\ 0 & 1 & 0 & \Delta t \\ 0 & 0 & 1 & 0 \\ 0 & 0 & 0 & 1 \end{bmatrix}, \quad Q_{2D} = \sigma_q^2 \begin{bmatrix} \frac{\Delta t^4}{4} & 0 & \frac{\Delta t^3}{2} & 0 \\ 0 & \frac{\Delta t^4}{4} & 0 & \frac{\Delta t^3}{2} \\ \frac{\Delta t^3}{2} & 0 & \Delta t^2 & 0 \\ 0 & \frac{\Delta t^3}{2} & 0 & \Delta t^2 \end{bmatrix},
\end{equation}
with $\sigma_q^2 = 0.01$.

Under $H_1$, the maneuver follows a Coordinated Turn (CT) with a dynamic turn rate $\omega_{t+1} = \omega_t + \delta_\omega$, where $\delta_\omega \sim \mathcal{N}(0, \sigma_\omega^2)$ acts as the angular acceleration noise with $\sigma_\omega^2 = 0.01$. This is governed by the nonlinear transition $x_{t+1} = f_{H_1}(x_t) + w_t$, where:
\begin{equation}
    f_{H_1}(x_t) = \begin{bmatrix}
        p_{x,t} + \frac{\sin(\omega_t \Delta t)}{\omega_t} v_{x,t} - \frac{1 - \cos(\omega_t \Delta t)}{\omega_t} v_{y,t} \\
        p_{y,t} + \frac{1 - \cos(\omega_t \Delta t)}{\omega_t} v_{x,t} + \frac{\sin(\omega_t \Delta t)}{\omega_t} v_{y,t} \\
        \cos(\omega_t \Delta t) v_{x,t} - \sin(\omega_t \Delta t) v_{y,t} \\
        \sin(\omega_t \Delta t) v_{x,t} + \cos(\omega_t \Delta t) v_{y,t}
    \end{bmatrix}.
\end{equation}
The sensor measures planar positions $y_t = H_{2D} x_t + v_t$ with observation matrix $H_{2D} = \begin{bmatrix} 1 & 0 & 0 & 0 \\ 0 & 1 & 0 & 0 \end{bmatrix}$ and measurement covariance $R_{2D} = \sigma_r^2 I_2$ ($\sigma_r^2 = 1.0$).

For both the null and alternative trajectories, the initial kinematic components are drawn from Uniform distributions, while the alternative maneuver initializes with a constant base turn rate:
\begin{align}
p_x(0), p_y(0) &\sim \mathcal{U}(-10, 10), \nonumber \\ 
v_x(0), v_y(0) &\sim \mathcal{U}(-2, 2), \nonumber \\
\omega_0 &= 0.15 \text{ rad/s}.
\end{align}

\paragraph{3D Nonlinear Kinematic Model}
The 6D state vector is $x_t = [p_{x,t}, p_{y,t}, p_{z,t}, v_{x,t}, v_{y,t}, v_{z,t}]^\top$ with sampling interval $\Delta t = 1\text{ s}$. For both hypotheses, the initial kinematics are drawn from Uniform distributions, and the base turn rate is initialized at $\omega_0 = 0.02 \text{ rad/s}$:
\begin{align}
    p_x(0), p_y(0) &\sim \mathcal{U}(-10, 10), \quad p_z(0) \sim \mathcal{U}(5, 15), \nonumber \\
    v_x(0), v_y(0) &\sim \mathcal{U}(-2, 2), \quad v_z(0) \sim \mathcal{U}(0, 0.5).
\end{align}

Under $H_0$, the target executes a nominal steady helical climb. The angular velocity acts as a hidden state subjected to process noise: $\omega_{t+1} = \omega_t + \delta_{\omega, t}$, where $\delta_{\omega, t} \sim \mathcal{N}(0, \sigma_\omega^2)$ with $\sigma_\omega = 0.01$. The nonlinear state transition $x_{t+1} = f_{H_0}(x_t, \omega_t)$ is explicitly defined as:
\begin{equation}
    f_{H_0}(x_t, \omega_t) = \begin{bmatrix}
        p_{x,t} + \frac{\sin(\omega_t \Delta t)}{\omega_t} v_{x,t} - \frac{1 - \cos(\omega_t \Delta t)}{\omega_t} v_{y,t} \\
        p_{y,t} + \frac{1 - \cos(\omega_t \Delta t)}{\omega_t} v_{x,t} + \frac{\sin(\omega_t \Delta t)}{\omega_t} v_{y,t} \\
        p_{z,t} + v_{z,t} \Delta t \\
        \cos(\omega_t \Delta t) v_{x,t} - \sin(\omega_t \Delta t) v_{y,t} \\
        \sin(\omega_t \Delta t) v_{x,t} + \cos(\omega_t \Delta t) v_{y,t} \\
        v_{z,t}
    \end{bmatrix}.
\end{equation}

Under $H_1$, the target  executes an aggressive, accelerating spiral. The turn rate experiences a severe deterministic acceleration combined with the existing process noise: $\omega_{t+1} = \omega_t + \alpha_\omega + \delta_{\omega, t}$, where $\alpha_\omega = 0.08 \text{ rad/s}^2$. The alternative state transition $f_{H_1}$ modifies the vertical altitude dynamics to reflect accelerating climb rates over time $t$:
\begin{align}
    p_{z,t+1} &= p_{z,t} + (v_{z,t} + a_z t)\Delta t, \\
    v_{z,t+1} &= v_{z,t} + \delta_z,
\end{align}
with the planar updates remaining identical to $f_{H_0}$ but evaluated using the rapidly accelerating $\omega_t$. The anomalous vertical acceleration parameters are set to $a_z = 0.1$ and $\delta_z = 0.05$.

For both scenarios, the surveillance sensor measures the 3D position coordinates $y_t = H_{3D} x_t + v_t$, with the observation matrix $H_{3D} = \begin{bmatrix} I_3 & 0_{3 \times 3} \end{bmatrix}$ and Gaussian measurement noise $v_t \sim \mathcal{N}(0, R_{3D})$, where $R_{3D} = \sigma_r^2 I_3$ ($\sigma_r^2 = 1.0$).
\subsection{Chaotic  Trajectory Generation}\label{ap:Chaotic_Generation}
\paragraph{H\'enon Map}
The data generation for the H\'enon map experiment follows its standard discrete-time difference equation. Because the H\'enon map is inherently discrete, its state evolution is computed exactly without requiring continuous-time numerical integration (such as Euler or Runge-Kutta) or Taylor polynomial approximations. The system's evolution is governed by the following second-order difference equation:
$$y_{t+1} = 1.0 - a y_t^2 + b y_{t-1}$$

Each trajectory is initialized with $y_0$ and $y_1$ drawn uniformly from the interval $[-0.5, 0.5]$. For the null regime, the system parameters are set to $a = 1.35$ and $b = 0.20$. For the alternative, the parameter $b$ is   shifted to $0.30$. 

\begin{figure}[H]
    \centering 
    \begin{subfigure}[b]{0.32\textwidth}
        \centering
\includegraphics[width=\textwidth]{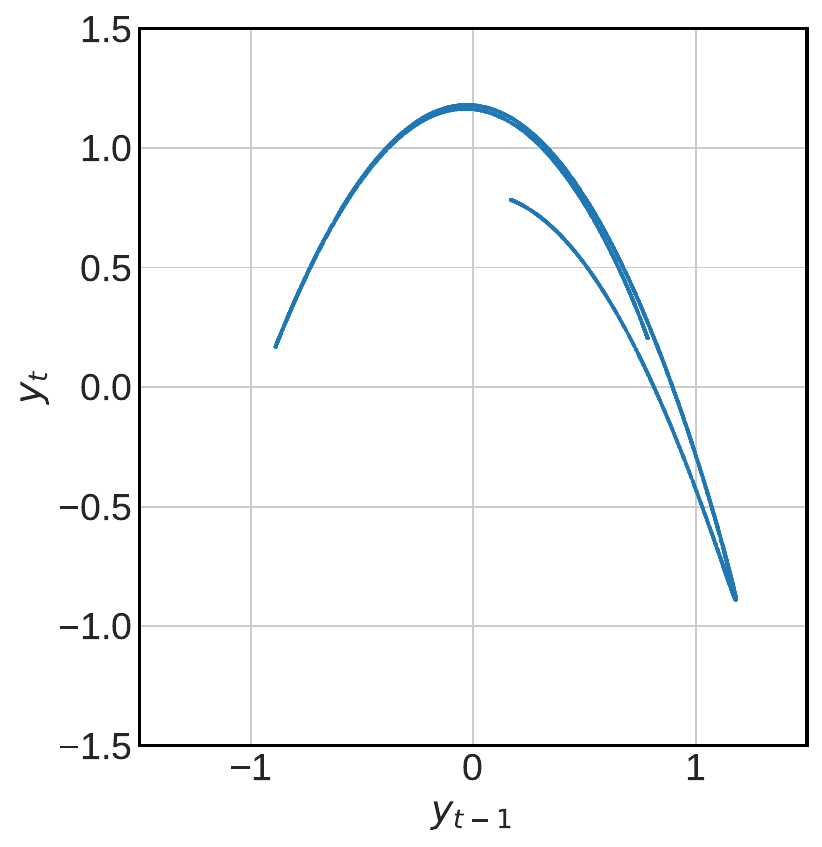}
        \caption{Null}
        \label{fig:henon_null}
    \end{subfigure}
    \hfill
    \begin{subfigure}[b]{0.32\textwidth}
        \centering       \includegraphics[width=\textwidth]{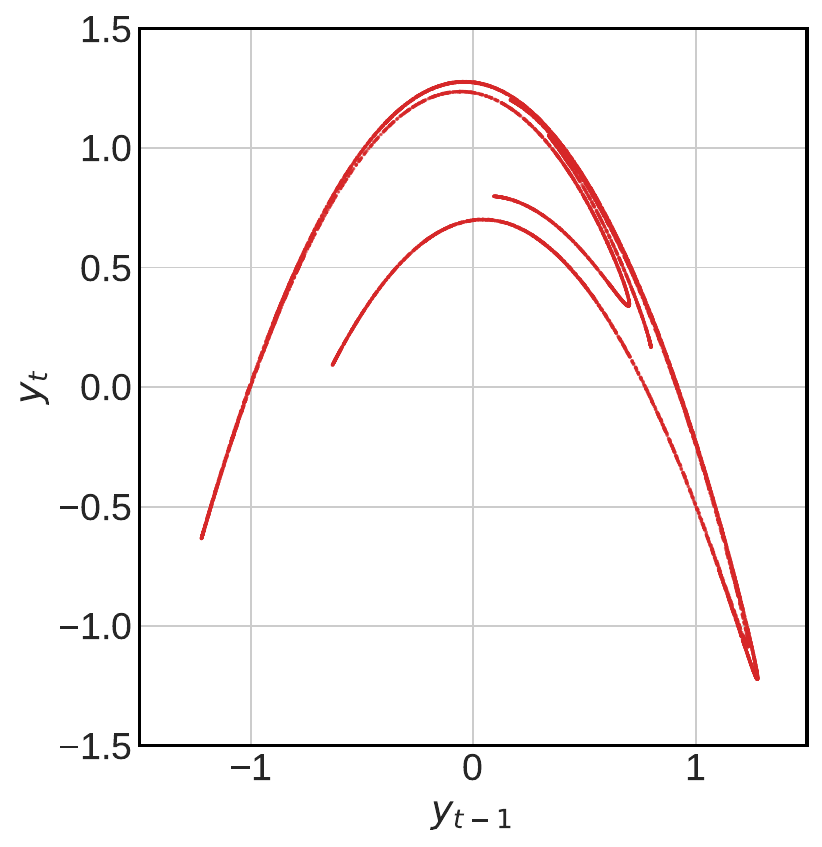}
        \caption{Alternative}
        \label{fig:henon_alt}
    \end{subfigure}
    \hfill
    \begin{subfigure}[b]{0.32\textwidth}
        \centering    \includegraphics[width=\textwidth]{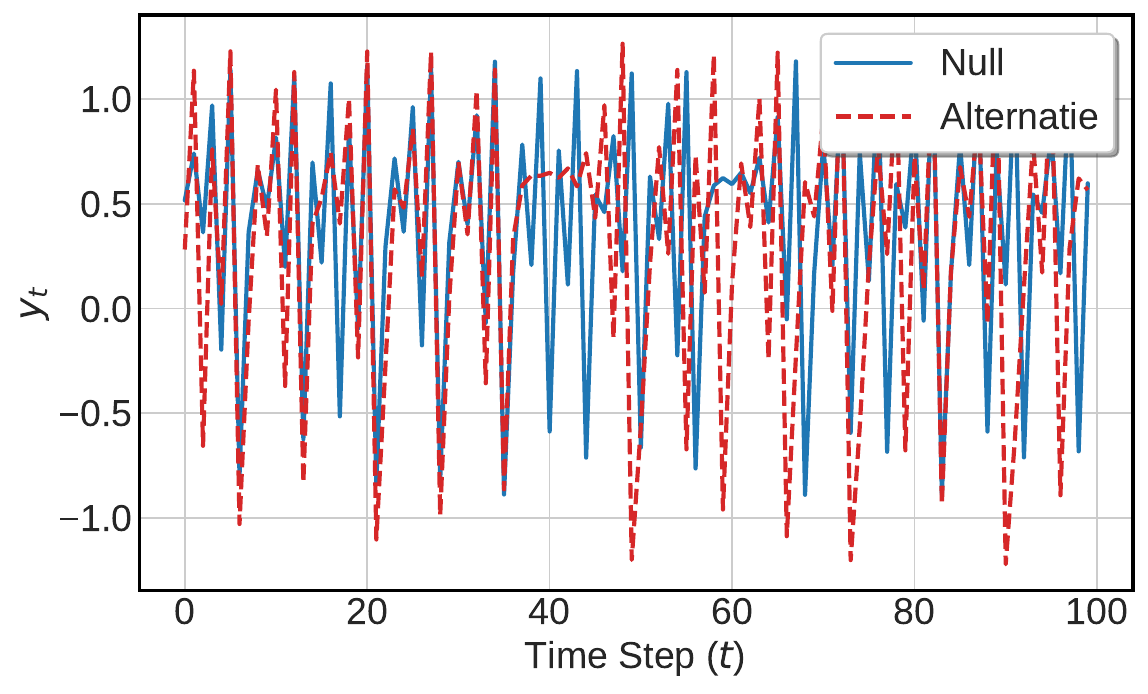}
        \caption{Time-Domain Comparison}
        \label{fig:henon_time}
    \end{subfigure}
    \caption{Visual comparison of the discrete Hénon map under null and alternative hypothesis. The phase portraits (a) and (b) demonstrate the highly intertwined, chaotic geometry of the attractors, while (c) highlights the subtle temporal divergence.}
    \label{fig:henon_unified}
\end{figure}

While no mathematical approximations are used to step the equations forward, two data processing transformations are applied to a generated sequence before it is passed to the LSTM:
\begin{itemize}
    \item \textbf{Burn-in Truncation:} The first 500 steps of the generated sequence are discarded to eliminate transient initialization effects and ensure the sequence has settled onto the chaotic attractor.
    \item \textbf{Subsampling:} The resulting sequence is subsampled at an interval of $W = 2$ (retaining every second observation) to break immediate temporal autocorrelation.
\end{itemize}
Training of the LSTM is performed on a null sequence of $T_{max} = 20,000$ observation steps. 
A visualization of two considered  null and alternative trajectories is provided in Fig. \ref{fig:henon_unified}. 
\paragraph{Lorenz-Chen Attractors}
The continuous-time multidimensional chaotic time series  are generated by numerically integrating the autonomous Ordinary Differential Equations (ODEs) of the Lorenz and Chen attractors. The null regime is modeled by the Lorenz system:
\begin{align}
\frac{dx}{dt} &= \sigma (y - x) \\
\frac{dy}{dt} &= x (\rho - z) - y \\
\frac{dz}{dt} &= xy - \beta z
\end{align}
with standard chaotic parameters $\sigma = 10.0$, $\rho = 28.0$, and $\beta = 8/3$. The alternative regime is modeled by the Chen system:
\begin{align}
\frac{dx}{dt} &= a (y - x) \\
\frac{dy}{dt} &= (c - a)x - xz + cy \\
\frac{dz}{dt} &= xy - bz
\end{align}
with parameters $a = 35.0$, $b = 3.0$, and $c = 28.0$.

\begin{figure}[H]
    \centering
    \begin{subfigure}[b]{0.48\textwidth}
        \centering
        \includegraphics[width=\textwidth]{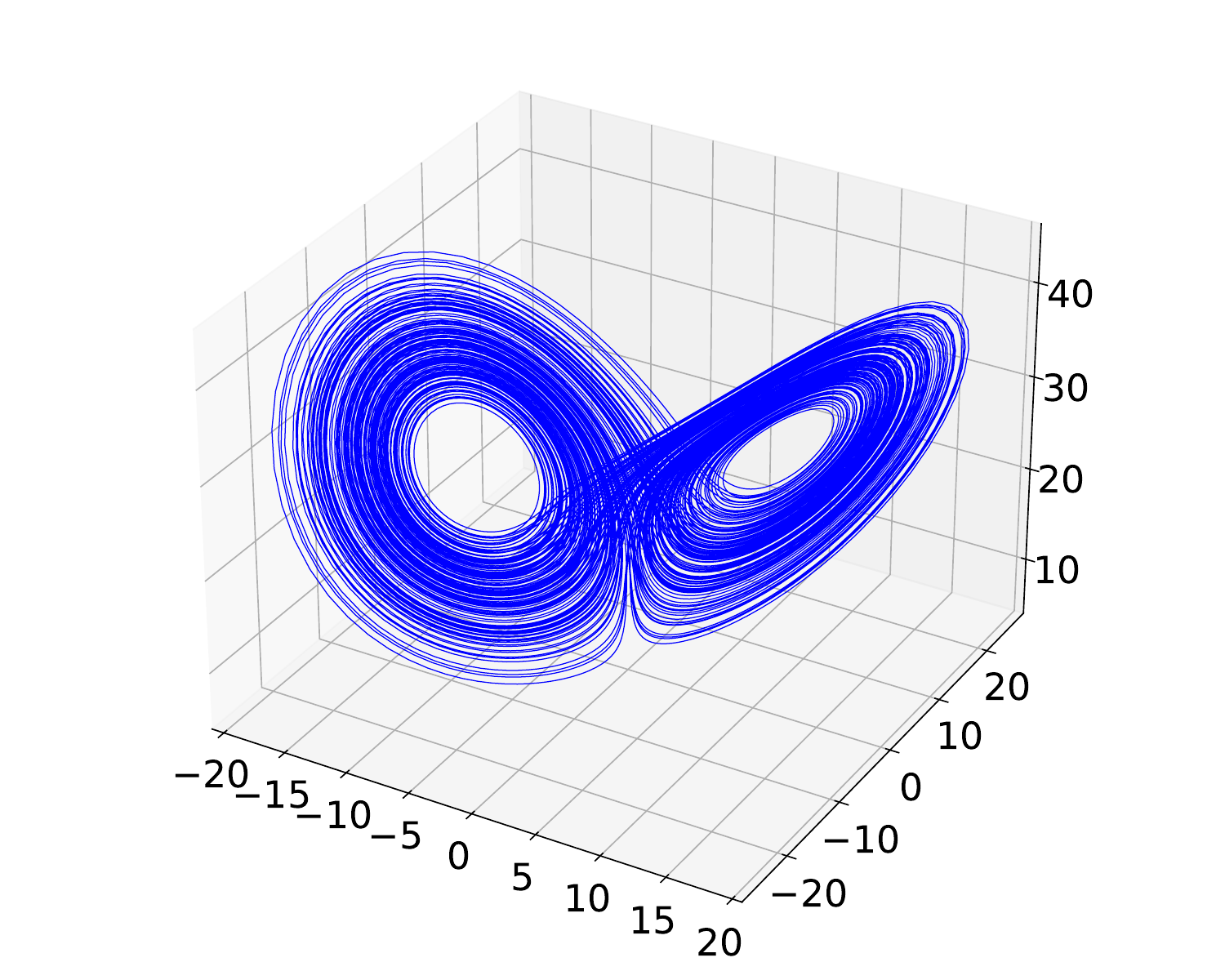}
        \caption{Lorenz Attractor ($H_0$)}
        \label{fig:lorenz_traj}
    \end{subfigure}
    \hfill
    \begin{subfigure}[b]{0.48\textwidth}
        \centering
        \includegraphics[width=\textwidth]{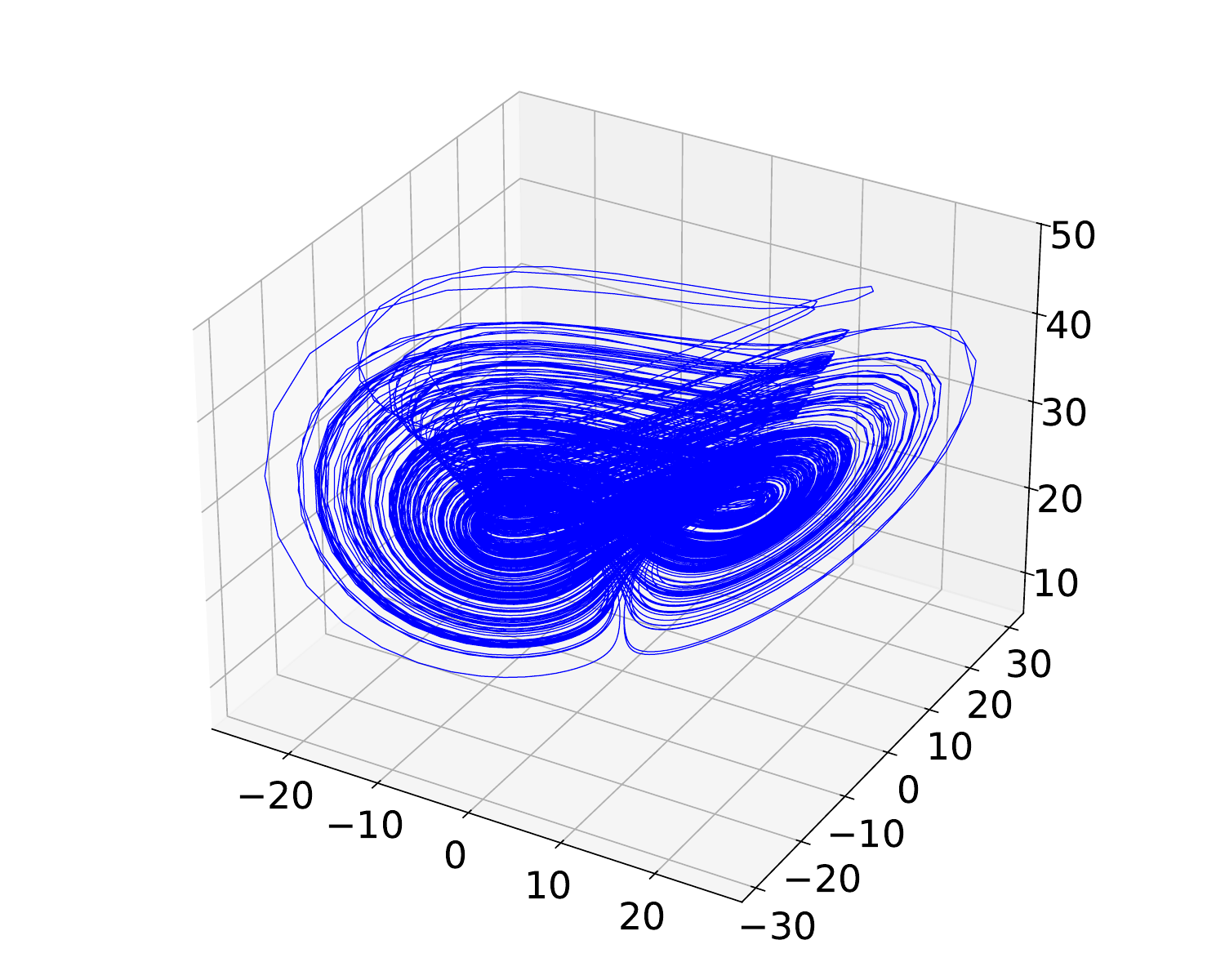}
        \caption{Chen Attractor ($H_1$)}
        \label{fig:chen_traj}
    \end{subfigure}
    \caption{Three-dimensional  trajectories (20,000 steps)   for the Lorenz and Chen dynamical systems.}
    \label{fig:attractors_side_by_side}
\end{figure}

The complete continuous trajectories are computed in a single batch operation using a robust numerical ODE solver (\texttt{scipy.integrate.odeint}) evaluated over a continuous time array with a fixed discretization step of $\Delta t = 0.01$. To synthesize the final observation sequences, the following data processing pipeline is applied:

\begin{itemize}
    \item \textbf{Initialization and Burn-in:} Continuous trajectories are initialized from spatial coordinates drawn uniformly from $[-1, 1]$. The first 2,000 time steps are discarded as a burn-in period to eliminate initial transient dynamics and guarantee the sequence strictly tracks the underlying  attractor.
    \item \textbf{Observational Noise:} Independent Gaussian noise $\mathcal{N}(0, 0.1^2)$ is injected into all three spatial coordinates of respective integrated trajectories to simulate  measurement uncertainty.
\end{itemize}
The LSTM model is  trained on a large  Lorenz sequence  of $T_{max} = 20,000$ observation steps, providing sufficient temporal coverage of the attractor's phase space for robust parameter estimation.
Trajectories from both attractors are visualized in Fig. \ref{fig:attractors_side_by_side}.
\subsection{AE for Network Traffic Tracking}
The AE models for both the CIC-IDS-2018 and CESNET24 datasets are implemented in PyTorch~\cite{pytorch} using two linear ReLU-activated layers in both the encoder and decoder. For CIC-IDS-2018, the network features a hidden dimension of 64 and a bottleneck of 32, trained on 100,000 data entries for 50 epochs with a batch size of 256. For the lower-dimensional CESNET24 dataset, a smaller AE model is employed, featuring a hidden dimension of 32 and a bottleneck of 8, trained on a 70\% split of the nominal data for 50 epochs with a batch size of 64. Both networks are optimized using the Adam optimizer~\cite{Adam} with a learning rate of $10^{-3}$ and a Mean Squared Error (MSE) loss function.
\subsection{Anomaly Detection Benchmarks for Fixed Horizon Network Traffic Tracking}
\begin{itemize}
    \item \textbf{One-Class SVM:} Configured with a linear kernel ($\texttt{kernel='linear'}$), automatic kernel coefficient scaling ($\texttt{gamma='scale'}$), and an outlier fraction parameter of $\nu = 0.05$. 
    The model is trained on 20,000 standardized benign flow records from the CIC-IDS-2018 benchmark, with sequence-level decision thresholds calibrated on 1,000 validation sequences.
    \item \textbf{Isolation Forest:} Constructed with 100 base trees ($\texttt{n\_estimators=100}$), and an expected contamination rate of $0.05$ ($\texttt{contamination=0.05}$).  Training and threshold calibration follow the identical 20,000-sample benign training set and $1,000$ sequences validation protocol.
\end{itemize}
The sklearn library is used~\cite{scikit-learn} and scaling is implemented via its standard scaler.
\subsection{LSTM for Chaotic Time Series}
The LSTM models for the chaotic time-series experiments (Hénon map, and Lorenz-to-Chen attractors) are implemented in PyTorch using a two-layer architecture with a hidden dimension of 24 for the Hénon model and 64 for the Lorenz-to-Chen model, mapping input features to predictions via a linear fully connected layer. For the 1D Hénon map experiments, a lookback window of 5 is used with a subsampling stride of $W = 2$, and the model is trained on baseline/null trajectories for 100 epochs using the Adam optimizer with a learning rate of $0.005$ and Mean Squared Error (MSE) loss. For the multi-dimensional Lorenz-to-Chen dynamical systems, a 3D input state with a lookback of 5 and a subsampling stride of $W = 2$ is processed, where the model is trained on null Lorenz trajectories for 100 epochs using the same optimizer and loss configuration.
\subsection{LSTM-based Fixed Horizon Benchmarks for Chaotic Time Series}
\begin{itemize}
    \item \textbf{Predictive LSTM:} A two-layer LSTM with a hidden dimension of 64 followed by a linear output layer, trained on the MSE loss via the Adam optimizer (learning rate $1\times 10^{-3}$, batch size 128) for $100$ epochs on $20,000$ nominal/Lorenz sequences to forecast the final step (step $T_{\rm fixed})$ of a trajectory.
    \item \textbf{Sequence-to-Sequence LSTM Autoencoder (LSTM-AE):} A two-layer encoder-decoder architecture featuring a hidden dimension of 64 and a compressed latent bottleneck dimension of 4 (with global temporal mean-pooling). It is trained using MSE loss via Adam (learning rate $1\times 10^{-3}$, batch size 128) for 100 epochs on 20,000 nominal sequences.
    \item \textbf{Probabilistic LSTM-VAE:} A two-layer LSTM encoder (with a hidden dimension of 64) mapping sequences to a 4-dimensional latent distribution via reparameterization, paired with a two-layer LSTM decoder (also with a hidden dimension of 64). It is optimized using the Evidence Lower Bound (ELBO) loss with a Kullback-Leibler divergence weight of $0.1$, trained using Adam (learning rate $1\times 10^{-3}$, batch size 128) for up to 1,000 epochs on 20,000 nominal samples.
    \item \textbf{Hybrid VAE-LSTM:} A two-stage architecture combining a local Window VAE (window size of $5$, 16 hidden units, 2-dimensional latent space) trained for 500 epochs, coupled with a single-layer LSTM (hidden dimension 32) trained for 500 epochs on the resulting latent embedding sequences using Adam ($1\times 10^{-3}$).
\end{itemize}
\subsection{Scan-B SCPD detector~\cite{ScanB}}
Because Scan-B requires a reference pool of pre-change data we generated pre-change only trajectories offline and constructed reference pools by randomly sampling disjoint blocks.  Each reference pool consists of $N=5$ blocks. 
 To calibrate the detection threshold for our target ARL of $5000$, we directly utilized the theoretical ARL approximation derived via the change-of-measure technique for the online Scan-B statistic. Furthermore, to ensure the threshold accurately maintains this target false alarm rate despite the non-vanishing skewness of the finite-sample statistic, we applied the paper's recommended skewness correction.

\subsection{kNN SCPD detector~\cite{kNNCPD}}
The method is provided with oracle knowledge that $\tau^* \geq 20$ in order to construct an initial neighbor graph required by the statistic.  Sequential evaluation is performed using the $T_3$ stopping rule from the paper.  The original framework introduces a local boundary buffer parameter, $n_0$, to maintain statistical stability by ensuring a minimum number of samples exist on both sides of any candidate split being evaluated. This constraint prevents the algorithm from dividing by near-zero variances. In our experiments, we set this parameter to $n_0 = 5$. Finally, to calibrate the decision threshold for the target ARL of $5000$, we utilize the skewness-corrected analytic approximation formula specifically derived for the $T_3$ rule.

\subsection{PCA-based SCPD detector~\cite{PCA_KDD_15}}
The implementation
initializes a nominal reference window directly from the first 30 samples of the incoming data stream (i.e, oracle knowledge that $\tau^* \geq 30)$. PCA is fitted to this window to capture the $99\%$  of the total data variance. Distributions between the reference and the sliding test windows are compared utilizing the maximum symmetric Kullback-Leibler (CD-MKL) divergence score. We follow the recommended hyperparameter configurations setting $\delta=0.005$ and $\xi=500$. 
\subsection{mSSA  SCPD detector~\cite{msSA}}
We followed closely the implementation guidelines of the supplementary material of the paper. The base matrix is constructed using an initial observation window of $T_0 = 0.6 \times \tau^*$, alongside a lag parameter configured as $L = \lfloor \sqrt{T_0} \rfloor$. The estimated time series order, $\hat{k}$, is dynamically determined by retaining the top singular vectors that account for 95\% of the base matrix's squared singular value sum. To ensure the CUSUM detection score maintains a negative expectation prior to a change, the shift parameter $c$ is calibrated using cross-validation on 90\% of the base matrix's lagged vectors, setting it slightly above the estimated noise floor and the maximum empirical subspace distance. Finally, the sequential decision threshold $h$ is established by scaling the maximum subspace distance observed during this same calibration process. Specifically, the threshold is set to 1.5 times the highest residual distance recorded when evaluating the remaining 10\% of the base matrix's lagged vectors against the estimated subspace.

\end{document}